\documentclass[11pt,twoside]{article}

\usepackage{fullpage}

\usepackage{comment}
\usepackage{epsf}
\usepackage{graphicx}

\usepackage{subfigure}

\usepackage{color}

\usepackage{mathtools}
\usepackage{amsfonts}
\usepackage{amsthm}
\usepackage{amsmath, bm}
\usepackage{amssymb}

\usepackage{pgfplots}
\pgfplotsset{width=5.35cm,compat=1.9}

\usepackage{textcomp}
\usepackage{xcolor}

\DeclareSymbolFont{extraup}{U}{zavm}{m}{n}
\DeclareMathSymbol{\vardiamond}{\mathalpha}{extraup}{87}

\usepackage[ruled, vlined, lined, commentsnumbered]{algorithm2e}

\newtheorem{theorem}{Theorem}

\newtheorem{assumption}{Assumption}
\newtheorem{proposition}{Proposition}
\newtheorem{lemma}{Lemma}
\newtheorem{corollary}{Corollary}
\newtheorem{definition}{Definition}

\long\def\comment#1{}

\newcommand{\red}[1]{\textcolor{red}{#1}}

\newcommand{\adityacomment}[1]{{\bf{{\red{{WISE-AG --- #1}}}}}}

\newcommand{\matsnorm}[2]{|\!|\!| #1 | \! | \!|_{{#2}}}

\newcommand{\defn}{:\,=}

\newcommand{\sgn}{\mathsf{sgn}}

\newcommand{\dist}{\mathsf{dist}}

\newcommand{\fronorm}[1]{\ensuremath{\matsnorm{#1}{\tiny{\mbox{F}}}}}
\newcommand{\opnorm}[1]{\ensuremath{\matsnorm{#1}{\tiny{\mbox{op}}}}}
\newcommand{\norms}[1]{\left\|#1 \right\|}

\newcommand{\paren}[1]{\left\{ #1 \right\} }

\newcommand{\eucnorm}[1]{\left\|#1 \right\|}

\newcommand{\abs}[1]{\ensuremath{ \left| #1 \right|}}

\newcommand{\sig}[1]{\ensuremath{\theta^{(#1)}}}
\newcommand{\appsig}[1]{\ensuremath{\beta^{(#1)}}}

\newcommand{\vstar}{\ensuremath{v^*}}
\newcommand{\ustar}{\ensuremath{u^*}}
\newcommand{\wstar}{\ensuremath{w^*}}
\newcommand{\vvec}[1]{\ensuremath{v^{(#1)}}}

\newcommand{\Deltamax}{\ensuremath{\mathsf{B}_{\mathsf{max}}}}

\newcommand{\Sup}[1]{\ensuremath{\mathsf{S}^{(#1)}}}
\newcommand{\sball}{\ensuremath{\zeta}}
\newcommand{\pimin}{\ensuremath{\pi_{\min}}}
\newcommand{\cs}{\ensuremath{c_{\mathsf{s}}}}
\newcommand{\Csig}{\ensuremath{C_{\eta, \sball, \cs}}}

\newcommand{\msdiff}{\ensuremath{\varrho}}

\newcommand{\Bvol}{\ensuremath{\mathtt{B}_{\mathsf{vol}}}}

\newcommand{\numdim}{\ensuremath{d}}
\newcommand{\numobs}{\ensuremath{n}}

\DeclareMathOperator{\tr}{{\sf tr}}
\newcommand{\ones}{\ensuremath{{\bf 1}}}

\newcommand{\inprod}[2]{\ensuremath{\langle #1 , \, #2 \rangle}}

\newcommand{\kull}[2]{\ensuremath{D_{\mathsf{KL}}(#1\; \| \; #2)}}

\newcommand{\E}{\ensuremath{{\mathbb{E}}}}
\newcommand{\EE}{\ensuremath{{\mathbb{E}}}}
\newcommand{\Prob}{\ensuremath{{\mathbb{P}}}}

\newcommand{\ind}[1]{\ensuremath{{\mathbf{1}\left\{ #1 \right\}}}}

\newcommand{\1}{\ensuremath{{\sf (i)}}}
\newcommand{\2}{\ensuremath{{\sf (ii)}}}
\newcommand{\3}{\ensuremath{{\sf (iii)}}}
\newcommand{\4}{\ensuremath{{\sf (iv)}}}
\newcommand{\5}{\ensuremath{{\sf (v)}}}

\newcommand{\eig}[1]{\ensuremath{\lambda_{#1}}}
\newcommand{\eigmax}{\ensuremath{\eig{\max}}}
\newcommand{\eigmin}{\ensuremath{\eig{\min}}}

\DeclareMathOperator{\diag}{diag}

\DeclareMathOperator{\cov}{cov}
\DeclareMathOperator*{\argmin}{argmin}
\DeclareMathOperator*{\argmax}{argmax}

\DeclareMathOperator{\vol}{\mathfrak{vol}}
\DeclareMathOperator{\rank}{{\sf rank}}

\newcommand{\NORMAL}{\ensuremath{\mathcal{N}}}

\newcommand{\Aspace}{\ensuremath{\mathcal{A}}}
\newcommand{\Xspace}{\ensuremath{\mathcal{X}}}

\newcommand{\Fspace}{\ensuremath{\mathcal{F}}}
\newcommand{\Espace}{\ensuremath{\mathcal{E}}}
\newcommand{\Ispace}{\ensuremath{\mathcal{I}}}

\newcommand{\Sspace}{\ensuremath{\mathcal{S}}}

\newcommand{\Wspace}{\ensuremath{\mathcal{W}}}
\newcommand{\Cspace}{\ensuremath{\mathcal{C}}}
\newcommand{\Bspace}{\ensuremath{\mathcal{B}}}
\newcommand{\Pspace}{\ensuremath{\mathcal{P}}}

\newcommand{\thetastar}{\ensuremath{\theta^*}}
\newcommand{\thetahat}{\ensuremath{\widehat{\theta}}}
\newcommand{\Mhat}{\ensuremath{\widehat{M}}}

\newcommand{\etatil}{\ensuremath{\widetilde{\eta}}}

\newcommand{\xitil}{\ensuremath{\widetilde{\xi}}}
\newcommand{\tautil}{\ensuremath{\widetilde{\tau}}}
\newcommand{\mutil}{\ensuremath{\mu_{\tau}}}
\newcommand{\Sigmatil}{\ensuremath{\Sigma_{\tau}}}

\newcommand{\Thetastar}{\ensuremath{\Theta^*}}

\newcommand{\order}{\ensuremath{\mathcal{O}}}

\newcommand{\Xmat}{\ensuremath{X}}
\newcommand{\AppXmat}{\ensuremath{\Xi}}
\newcommand{\sigstar}{\ensuremath{\theta^*}}
\newcommand{\bstar}{\ensuremath{b^*}}
\newcommand{\appsigstar}{\ensuremath{\beta^*}}

\newcommand{\meas}{\ensuremath{x}}
\newcommand{\appmeas}{\ensuremath{\xi}}

\newcommand{\real}{\ensuremath{\mathbb{R}}}
\newcommand{\Sd}{\ensuremath{\mathbb{S}^{d-1}}}
\newcommand{\Bd}{\ensuremath{\mathbb{B}^{d}}}

\newcommand{\Uhat}{\ensuremath{\widehat{U}}}
\newcommand{\Vhat}{\ensuremath{\widehat{V}}}

\newcommand{\Ustar}{\ensuremath{U^*}}
\newcommand{\Astar}{\ensuremath{A^*}}

\newcommand{\ellsharp}{\ensuremath{\ell^{\#}}}

\newcommand{\R}{\ensuremath{\mathbb{R}}}

\newcommand{\resp}{\ensuremath{\phi}}

\newcommand{\covarstar}{\ensuremath{\chi^*}}
\newcommand{\covarhat}{\ensuremath{\widetilde{\chi}}}
\newcommand{\sigstargen}{\ensuremath{\Upsilon^*}}
\newcommand{\siggen}{\ensuremath{\Upsilon}}
\newcommand{\sigstarint}{\ensuremath{\omega^*}}
\newcommand{\sigint}{\ensuremath{\omega}}
\newcommand{\sigstaraug}{\ensuremath{\nu^*}}

\newcommand\blfootnote[1]{%
  \begingroup
  \renewcommand\thefootnote{}\footnote{#1}%
  \addtocounter{footnote}{-1}%
  \endgroup
}

\def\qt#1{\qquad\text{#1}}

\begin{document}

\begin{center}

{\bf{\LARGE{Max-Affine Regression: Provable, Tractable, and Near-Optimal Statistical Estimation}}}

\vspace*{.2in}

{\large{
\begin{tabular}{cc}
Avishek Ghosh$^{\star, \dagger}$ & Ashwin Pananjady$^{\star, \dagger}$  \\
Adityanand Guntuboyina$^\ddagger$ & Kannan Ramchandran$^\dagger$
\end{tabular}
}}
\vspace*{.2in}

\begin{tabular}{c}
Department of Electrical Engineering and Computer Sciences, UC Berkeley$^\dagger$ \\
Department of Statistics, UC Berkeley$^\ddagger$
\end{tabular}

\vspace*{.2in}

\today

\end{center}
\vspace*{.11in}

\begin{abstract}
Max-affine regression\blfootnote{$^\star$Avishek Ghosh and Ashwin
  Pananjady contributed equally to this work.} refers to a model where the unknown 
regression function is modeled as a maximum of $k$ unknown affine
functions for a fixed $k \geq 1$. This generalizes linear regression
and (real) phase retrieval, and is closely related to convex regression. Working
within a non-asymptotic framework, we
study this problem in the high-dimensional setting 
assuming that $k$ is a fixed constant, and focus on estimation of the
unknown coefficients of the affine functions underlying the model. 
We analyze a natural alternating
minimization (AM) algorithm for the non-convex least squares
objective when the design is random. We show
that the AM algorithm, when initialized suitably, 
converges with high probability and at a geometric rate to a small
ball around the optimal coefficients. 
In order to initialize the algorithm, we propose and analyze a
combination of a spectral method and a random search scheme in a
low-dimensional space, which may be of independent interest. The final
rate that we obtain is near-parametric and minimax
optimal (up to a polylogarithmic factor) as a function of the dimension, sample size, and noise variance. In that sense, our
approach should be viewed as a \emph{direct} and implementable method
of enforcing regularization to alleviate the curse of dimensionality
in problems of the convex regression type. As a by-product of our
analysis, we also obtain guarantees on a classical algorithm for the
phase retrieval problem under considerably weaker assumptions on the
design distribution than was previously known. Numerical experiments
illustrate the sharpness of our bounds in the various problem
parameters.  
\end{abstract}


\section{Introduction}
Max-affine regression refers to the regression model
\begin{align}\label{mafd}
  Y = \max_{1 \leq j \leq k} \left(\inprod{X}{\sigstar_j} + \bstar_j \right) + \epsilon
\end{align}
where $Y$ is a univariate response, $X$ is a $d$-dimensional
vector of covariates and $\epsilon$ models zero-mean noise that is independent of $X$. We assume that $k \geq 1$ is a known integer and study the problem of estimating the unknown parameters $\theta_1^*, \dots, \theta_k^* \in \R^d$ and $b_1^*, \dots, b_k^* \in \R$ from independent observations $(x_1, y_1), \dots,
(x_n, y_n)$ drawn according to the model \eqref{mafd}. 

Let us provide some motivation for studying the
model~\eqref{mafd}. When $k = 1$, equation~\eqref{mafd} corresponds
to the classical linear regression model. When $k = 2$, the intercepts
$b^*_2 = b^*_1 = 0$, and $\sigstar_2 = -\sigstar_1 = \sigstar$, the
model \eqref{mafd} reduces to  
\begin{align}\label{prmd}
Y = |\inprod{X}{\sigstar}| + \epsilon.
\end{align}
The problem of recovering $\theta^*$ from observations drawn according to the above model is known as (real) phase retrieval---variants of which arise in a diverse array of science and engineering applications~\cite{harrison1993phase, fienup1987phase, chai2010array, fogel2016phase}---and has associated with it an extensive statistical and algorithmic literature.

To motivate the model~\eqref{mafd} for  general $k$, note that the function $x \mapsto \max_{1 \leq j \leq k} ( \inprod{x}{\sigstar_j}+b_j^* )$ is always a convex function and, thus, estimation under the model~\eqref{mafd} can be used to fit convex functions to the observed data. Indeed, the model~\eqref{mafd} serves as a parametric
approximation to the non-parametric convex regression model 
\begin{equation}\label{conre}
Y = \phi^*(X) + \epsilon,
\end{equation}
where $\phi^*: \R^d \rightarrow \R$ is an unknown convex function. It
is well-known that convex  regression suffers from the curse of
dimensionality unless $d$ is small, which is basically a consequence of
the fact that the metric entropy of natural totally bounded sub-classes
of convex functions grows exponentially in $d$ (see, e.g.,
\cite{bronshtein1976varepsilon, guntuboyina2013covering,
  gao2017entropy}). To overcome this curse of dimensionality, one
would need to work with more structured sub-classes of convex
functions. Since convex functions can be approximated to arbitrary
accuracy by maxima of affine functions, it is reasonable  to
regularize the problem by considering only those convex functions
that can be written as a maximum of a fixed number of affine
functions. Constraining the number of affine pieces in the function
therefore presents a simple method to enforce structure, and such function classes
have been introduced and studied in the convex regression literature
(see e.g., ~\cite{han2016multivariate}). This assumption directly leads to our model~\eqref{mafd},
and it has been argued by \cite{magnani_convex,
  hannah2013multivariate, balazs2016convex} that the parametric model \eqref{mafd}  is a tractable alternative to the full non-parametric convex regression model~\eqref{conre} in common applications of convex regression to data arising in economics, finance and operations research where $d$ is often moderate to large. 

Another motivation for the model \eqref{mafd} comes from the problem of estimating convex sets from support function measurements. The support function of a compact convex set $K \subseteq \real^d$ is defined by $h_K(x) \defn \sup_{u \in K} \inprod{x}{u}$ for $d$-dimensional unit vectors $x$. The problem of estimating an unknown compact, convex set $K^*$ from noisy measurements of $h_{K^*}(\cdot)$ arises in certain engineering applications such as robotic tactile sensing and projection magnetic resonance imaging (see, e.g., \cite{prince1990reconstructing, gregor2002three, gardner_2006}). Specifically, the model considered here is
\begin{equation*}
  Y = h_{K^*}(X) + \epsilon,
\end{equation*}
and the goal is to estimate the set $K^* \subseteq \R^d$. As in convex regression, this problem suffers from a curse of dimensionality unless $d$ is small, as is evident from known minimax lower bounds~\cite{guntuboyina2012optimal}. To alleviate this curse, it is natural to restrict $K^*$ to the class of all polytopes with at most $k$ extreme points for a fixed $k$; such a restriction has been studied as a special case of enforcing structure in these problems by Soh and Chandrasekharan~\cite{soh2019fitting}. Under this restriction, one is led to the model \eqref{mafd} with $b_1^* = \dots = b_k^* = 0$, since if $K^*$  is the polytope given by the convex hull of $\theta_1^*, \dots, \theta_k^* \in \R^d$, then its support function is equal to $x \mapsto \max_{1 \leq j \leq k} \inprod{x}{\theta_j^*}$.

Equipped with these motivating examples, our goal is to study a computationally efficient
estimation methodology for the unknown parameters of the model
\eqref{mafd} from i.i.d observations $(x_i, y_i)_{i = 1}^n$. Before presenting our contributions, let us first rewrite the
observation model~\eqref{mafd} by using more convenient notation, and use it to
describe existing estimation procedures for this model. Denote the unknown parameters by $\beta_j^*
\defn (\theta_j^*, b_j^*) \in \real^{d+1}$ for $j = 1, \dots, k$ and the
observations by $(\xi_i, y_i)$ for $i = 1, \dots, n$, where $\xi_i \defn (x_i,
1) \in \R^{d+1}$. In this notation, the observation model takes the form
\begin{align} \label{eq:model}
  y_i = \max_{1 \leq j \leq k}\; \inprod{\xi_i}{\beta_j^*} + \epsilon_i, 
  \qquad \text{ for } i = 1, 2, \ldots, \numobs. 
\end{align}
Throughout the paper, we assume that $x_1, \dots, x_n, \epsilon_1,
\dots, \epsilon_n$ are independent random variables with $x_1, \dots, x_n$
drawn from a common $d$-dimensional distribution $P_X$, and $\epsilon_1, \dots, \epsilon_n$ drawn from a (univariate) distribution that is zero-mean and sub-Gaussian, with unknown sub-Gaussian parameter~$\sigma$.  

Let us now describe existing estimation procedures for 
max-affine regression. The most obvious approach is the global least squares
estimator, defined as any minimizer of the least squares criterion
  \begin{align}
(\widehat{\beta}_1^{(\mathsf{ls})}, \dots, \widehat{\beta}_k^{(\mathsf{ls})}) \in \argmin_{\beta_1, \dots, \beta_k \in \R^{d+1}}  
  \sum_{i = 1}^n \left( y_i - \max_{1 \leq j \leq k} \inprod{\appmeas_i}{\beta_j} \right)^2. \label{intro:opt} 
\end{align}
It is easy to see (see Lemma~\ref{lse.exist} to follow) that a global minimizer of the least
squares criterion above always exists but it will not---at least in general---be unique,
since any relabeling of the indices of a minimizer will also be a
minimizer. While the least squares estimator has appealing statistical
properties (see, e.g.~\cite{vandegeerthesis, guntuboyina2012optimal,soh2019fitting}), the optimization problem \eqref{intro:opt} is non-convex
and, in general, NP-hard~\cite{fickus2014phase}. It is
interesting to compare~\eqref{intro:opt} to the optimization problem
used to compute the least squares estimator in the more general convex
regression model~\eqref{conre}, given by
\begin{align} \label{eq:clse}
  \widehat{\phi}^{(\mathsf{ls})} \in \argmin_{\phi} \sum_{i=1}^n \left(y_i -
    \phi(x_i) \right)^2,
\end{align}
where the minimization is over all convex functions $\phi$. In
sharp contrast to the problem~\eqref{intro:opt}, the optimization problem~\eqref{eq:clse} is
convex \cite{seijo2011nonparametric,lim2012consistency} and can be solved
efficiently for fairly large values of the pair $(d, n)$~\cite{mazumder2019computational}. Unfortunately however, the utility of
$\widehat{\phi}^{(\mathsf{ls})}$ in estimating the parameters of the
max-affine model is debatable, as it is unclear how one may obtain
estimates of the true parameters $\beta_1^*, \dots, \beta_k^*$  from
$\widehat{\phi}^{(\mathsf{ls})}$, which typically will \emph{not} be a maximum of only $k$ affine functions.

Three heuristic techniques for solving the non-convex
optimization problem~\eqref{intro:opt} were empirically evaluated by Bal{\'a}zs~\cite[Chapters 6 and 7]{balazs2016convex}, who compared running times and performance of these techniques
on a wide variety of real and synthetic datasets for convex regression.
The first technique is the alternating minimization
algorithm of Magnani and Boyd~\cite{magnani_convex}, the second technique is the convex adaptive partitioning (or CAP) algorithm of Hannah and Dunson~\cite{hannah2013multivariate}, and the third is the adaptive max-affine
partitioning algorithm proposed by Bal{\'a}zs himself~\cite{balazs2016convex}. 
The simplest and most intuitive of
these three methods is the first alternating minimization (AM) algorithm,
which is an iterative algorithm for estimating the parameters
$\beta_1^*, \dots, \beta_k^*$ and forms the focus of our study. In the $t$-th iteration of the algorithm, the
current estimates $\beta_1^{(t)}, \dots,
\beta_k^{(t)}$ are used to partition the observation indices $1,
\dots, n$ into $k$ sets $S_1^{(t)}, \dots, S_k^{(t)}$ such that $j \in
\argmax_{u \in [k]} \inprod{\xi_i}{\beta_u^{(t)}}$ for every $i
\in S_j^{(t)}$. For each $1 \leq j \leq k$, the next estimate $\beta_j^{(t+1)}$ is then  obtained by performing a least  squares
fit (or equivalently, linear regression) to the data $(\xi_i, y_i), i \in S_j^{(t)}$. More intuition and a formal description of the
algorithm are provided in Section~\ref{sec:setup}. 
Bal{\'a}zs found that when this algorithm was run on a variety of datasets
with multiple random initializations, it compared favorably with the 
state of the art in terms of its final predictive performance---see, for example,
Figures 7.4 and 7.5 in the thesis~\cite{balazs2016convex}, which show 
encouraging results when the algorithm is used to fit convex functions to datasets of average wages and aircraft profile drag data, respectively.
In the context of fitting convex sets to support function measurements, Soh and Chandrasekaran~\cite{soh2019fitting} recently proposed and empirically evaluated a similar algorithm.
 However, to the best of our knowledge, no theoretical results exist to support the performance
of such a technique.

In this paper, we present a theoretical analysis of the AM
algorithm for recovering the parameters
of the max-affine regression model under some assumptions on the
covariate distribution. Note that the AM algorithm described above can
be seen as a generalization of classical AM algorithms for (real) phase
retrieval~\cite{gerchberg1972practical, fienup1982phase}, which
have recently been theoretically analyzed in a series of
papers~\cite{netrapalli2013phase,walds, zhang2019phase} for Gaussian 
designs. The AM---and the closely related expectation maximization\footnote{Indeed, for many problems, the EM algorithm reduces to AM in the noiseless limit, and AM should thus be viewed as a variant of EM that uses hard-thresholding to determine values of the latent variables.}, or EM---methodology is
widely used for parameter estimation in missing data
problems~\cite{beale1975missing, hartley1958maximum} and mixture
models~\cite{xu1996convergence}, including those with covariates such
as mixtures-of-experts~\cite{jordan1994hierarchical} and
mixtures-of-regressions~\cite{chaganty2013spectral}
models. Theoretical guarantees for such algorithms have been established in multiple statistical contexts~\cite{wu1983convergence, tseng2004analysis, chretien2008algorithms, balakrishnan2017statistical}; in the case when the likelihood is not unimodal, these are typically of the local convergence type. In particular, algorithms of the EM type return, for many such latent variable models, minimax-optimal parameter estimates when initialized in a neighborhood of the optimal solution (e.g., ~\cite{chaganty2013spectral, zhang2014spectral, zhong2016mixed}); conversely, these algorithms can get stuck at spurious fixed points when initialized at random~\cite{jin2016local}. In some specific applications of EM to mixtures of two Gaussians~\cite{daskalakis2017ten, xu2016global} and mixtures of two regressions~\cite{kwon2019global}, however, it has been shown that randomly initializing the EM algorithm suffices in order to obtain consistent parameter estimates.
Here, we establish guarantees on the AM algorithm for
max-affine regression that are of the former type: we prove local
geometric convergence of the AM iterates when initialized in a
neighborhood of the optimal solution. 
We analyze the practical variant of the algorithm in which the steps
are performed without sample-splitting.
As in the case of mixture
models~\cite{hsu2013learning, chaganty2013spectral}, we use spectral
methods to obtain such an initialization. 

\paragraph{Contributions} Let us now describe our results in more detail. To simplify the exposition,
we state simplified corollaries of our theorems; for precise statements, see Section~\ref{sec:mainresults}. We begin by considering the case where the covariate distribution $P_X$ is the standard
$d$-dimensional Gaussian distribution. Such an assumption forms a natural starting point for the study of many iterative algorithms in related problems~\cite{netrapalli2013phase,walds,
  zhang2019phase, balakrishnan2017statistical}, and is also quite standard in theoretical
investigations of multidimensional regression problems. We also assume
that the true parameters are fixed. Under these assumptions, we prove in Theorem \ref{thm:mt}
that for each $\epsilon > 0$, the parameter estimates 
$\appsig{t}_1, \ldots, \appsig{t}_k$ returned by the AM algorithm at iteration
$t$ satisfy, with high probability, the inequality
\begin{equation}\label{intro:thm1}
  \sum_{j=1}^k \|\beta_j^{(t)} - \beta_j^*\|^2 \leq \epsilon +
  C(\appsigstar_1, \ldots, \appsigstar_k) \frac{\sigma^2 k d}{n} \log (kd) \log
  \left(\frac{n}{kd} \right)
\end{equation}
for every $t \geq \log_{4/3} \left( \frac{\sum_{j=1}^k \|\beta_j^{(0)} - \beta_j^*\|^2}{\epsilon} \right)$, provided that the sample size $n$ is sufficiently
large and that the initial estimates satisfy the condition
\begin{equation}\label{intro:init}
\min_{c > 0} \; \max_{1 \leq j \leq k} \; \| c \beta_j^{(0)} - \beta_j^*\|^2
\leq \frac{1}{k} c(\appsigstar_1, \ldots, \appsigstar_k). 
\end{equation}
Here $C(\appsigstar_1, \ldots, \appsigstar_k)$ and $c(\appsigstar_1, \ldots, \appsigstar_k)$ are
constants depending only on the true parameters $\appsigstar_1, \ldots, \appsigstar_k$, and their 
explicit values are given in Theorem~\ref{thm:mt}. The constant $c$ in equation~\eqref{intro:init} endows the initialization with a scale-invariance property: indeed, scaling all parameters $\appsig{0}_1, \ldots, \appsig{0}_k$ by the same positive constant $c$ produces the same initial partition of subsets $S_1^{(0)}, \dots, S_k^{(0)}$, from which the algorithm proceeds identically.

Treating $k$ as a fixed constant, inequality
\eqref{intro:thm1}  implies, under the initialization condition
\eqref{intro:init}, that 
the
  parameter estimates returned by AM converge geometrically to within a small ball
of the true parameters, and that this error term
%
is bounded by an error term that 
is nearly the parametric risk $\frac{\sigma^2 d}{n}$ up to a logarithmic factor. 
%
The initialization condition \eqref{intro:init} requires the distance
between the initial estimates and the true parameters to be at most a
specific ($k$-dependent) constant. It has been empirically observed that there exist bad
initializations under which the AM algorithm behaves poorly (see,
e.g.,~\cite{magnani_convex, balazs2016convex}) and the assumption
\eqref{intro:init} is one way to rule these out.

In Theorem~\ref{thm:sbgen}, we extend the conclusion of Theorem
\ref{thm:mt} in two directions (with some degradation in the constants $C(\appsigstar_1, \ldots, \appsigstar_k)$ and $c(\appsigstar_1, \ldots, \appsigstar_k)$). First, we analyze AM for more general covariate distributions
$P_X$ that are isotropic, sub-Gaussian and satisfy a
\textit{small-ball} condition (see Definition~\ref{def:sball} for the
precise formulation of this condition). This includes, for instance,
the uniform distribution on the set $[-c, c]^{d}$, which is commonly used
as a random design in investigations of non-parametric regression problems~\cite{wasserman2006all}, and
any isotropic log-concave
distribution with bounded support, studied, for instance, in the context
of convex regression and related problems~\cite{han2016multivariate}. 
The second strengthening afforded by Theorem \ref{thm:mt} is that equation~\eqref{intro:thm1}
is proved to hold uniformly over all true parameters lying in a large
space (unlike Theorem \ref{thm:mt} which holds for fixed
values of the true parameters). Such guarantees are common in the phase retrieval
literature; since the AM algorithm for max-affine
regression can be viewed as a generalization of the classical AM
algorithm for real phase retrieval, Theorem
\ref{thm:sbgen} essentially implies an accuracy result stated in Corollary~\ref{cor:PR} for real phase retrieval that holds for all isotropic, sub-Gaussian
distributions satisfying the small-ball condition. Notably, all previous results on the AM algorithm for phase retrieval~\cite{walds, zhang2019phase} only held
under the assumptions of Gaussian covariates and noiseless observations. 

A natural question based on our Theorems \ref{thm:mt} and~\ref{thm:sbgen} is whether it is possible to produce preliminary
estimates $\beta_1^{(0)}, \beta_k^{(0)}$ satisfying the initialization condition~\eqref{intro:init}. Indeed, one such method is to
repeatedly initialize parameters (uniformly) at random within the unit ball $\mathbb{B}^{d+1}$; Bal{\'a}zs empirically observed in a close relative of such a scheme (see Figure 6.6 in his thesis~\cite{balazs2016convex}) that increasing the number of random initializations is often sufficient to get the AM algorithm to succeed. However, reasoning heuristically,
the number of repetitions required to ensure that one such random initialization generates parameters that satisfy condition~\eqref{intro:init} increases exponentially in the ambient dimension $d$, and so it is reasonable to ask if, in large dimensions, there is some natural form of dimensionality reduction that allows us to perform this step in a lower-dimensional space.


When\footnote{If $k \geq d$, then this dimensionality reduction step can be done away with and one can implement the random search routine directly.} $k < d$ and the covariates are drawn from a Gaussian distribution,
we show that a natural spectral method (described formally in Algorithm~\ref{algo:pca}) is able to reduce the dimensionality of our problem from $d$ to $k$. In particular,
this method returns an
orthonormal basis of vectors $\widehat{U}_1, \dots, \widehat{U}_k$ such that the $k$-dimensional linear subspace spanned by these vectors accurately estimates the subspace spanned by the vectors $\sigstar_1, \dots, \sigstar_k$.
We form the matrix $\widehat{U} \defn [\widehat{U}_1 : \dots : \widehat{U}_k]$ by collecting these vectors as its columns, and in order to account for the intercepts, further append such a matrix to form the matrix
$
\Vhat \defn \begin{bmatrix} \Uhat & 0 \\ 0 & 1 \end{bmatrix}
 \in \real^{(d+1) \times (k+1)}$.
We then choose $M$ random initializations in $(k+1)$ dimensions---the $\ell$-th such initialization is given by a set of vectors $\nu^{\ell}_1, \ldots, \nu^{\ell}_k \in \real^{k+1}$ each chosen uniformly at random from the $(k+1)$-dimensional unit ball---so that the collection of $k$ vectors $\{ \Vhat \nu^\ell_j \}_{j = 1}^k$ serves as our $\ell$-th guess of the true parameters.
In order to decide which of these random points to choose, we evaluate (on an independent set of samples) the goodness-of-fit statistic $\min_{c \geq 0} \sum_i (y_i - c \max_{1 \leq j \leq k} \; \inprod{\appmeas_i}{ \Vhat \nu^{\ell}_j })^2$ for each $1 \leq \ell \leq M$, where the minimization over the constant $c$ accounts for the scale-invariance property alluded to above.
Letting $\ell^*$ denote the index with the smallest loss, we then return the initialization 
$\beta_j^{(0)} = \widehat{V} \nu^{\ell^*}_j$ for $j = 1, \dots, k$.

Our algorithm can thus be viewed as a variant of the repeated random initialization evaluated by Bal{\'a}zs~\cite{balazs2016convex}, but incurs significantly smaller computational cost, since we only run the full-blown iterative AM algorithm once.
Note that our algorithm treats the number of initializations $M$ as a tuning parameter to be chosen by the statistician, similar to Bal{\'a}zs~\cite{balazs2016convex}, but we show a concrete upper bound on $M$ that is sufficient to guarantee convergence. In particular, we show that in order to produce an initialization satisfying condition~\eqref{intro:init} with high probability, it suffices to choose $M$ as a function \emph{only} of the number of affine pieces $k$ and other geometric parameters of the problem (and independently of the sample size $n$ and ambient dimension $d$). 

%

To produce our overall guarantee for Gaussian covariates, we combine the initialization
with the AM algorithm in Corollary~\ref{cor:overall}, showing that provided the sample
size scales linearly in the dimension (with a multiplicative pre-factor that depends polynomially on $k$ and other problem-dependent parameters), we obtain estimates that are accurate up to the parametric risk. Our algorithm is also computationally efficient when $k$ is treated as a fixed constant. 


From a technical standpoint, our results for the AM algorithm are significantly more challenging
to establish than related results in the literature~\cite{balakrishnan2017statistical, mixture_many, walds, tan2017phase}.
First, it is technically very challenging to compute the \emph{population operator}~\cite{balakrishnan2017statistical}---corresponding to running the AM update in the infinite sample limit---in this setting even for Gaussian covariates, since the $\max$ function introduces intricate geometry in the problem that is difficult to reason about in closed form.
Second, we are interested in analyzing the AM update without sample-splitting, and so cannot assume that the iterates are independent of the covariates; the latter assumption has been used fruitfully in the literature to simplify analyses of such algorithms~\cite{mixture_many,zhang2019phase,netrapalli2013phase}. Third, and unlike algorithms for phase retrieval~\cite{walds, tan2017phase}, our algorithm performs least squares using sub-matrices of the covariate matrix $\AppXmat$ that are chosen \emph{depending} on our random iterates. Accordingly, a key technical difficulty of the proof, which may be of independent interest, is to control the spectrum of these random matrices, rows of which are drawn from (randomly) truncated variants of the covariate distribution. We explore two distinct techniques to obtain this control in Theorems~\ref{thm:mt} and~\ref{thm:sbgen}.  

Our spectral initialization algorithm is also a natural estimator based on the method-of-moments, and has been used in a variety of non-convex problems~\cite{chaganty2013spectral, zhang2014spectral, zhong2016mixed}. However, our guarantees for this step are once again non-trivial to establish. In particular, the eigengap of the population moment (on which the rates of the estimator depend) is difficult to compute in our case since the $\max$ function is not differentiable, and so it is not clear that higher order moments return reasonable estimates even in the infinite sample limit (see Section~\ref{sec:setup}).
However, since we operate exclusively with Gaussian covariates, we are able to use some classical moment calculations for truncated Gaussian distributions~\cite{tallis1961moment} in order to bound the eigengap. Translating these calculations into an eigengap is quite technical, and involves the isolation of many properties of the population moments that may be of independent interest.

Finally, it is important to note that owing to the scale invariance of our initialization condition~\eqref{intro:init} and goodness-of-fit statistic, our random search scheme does not require a bound on the size of the parameters; it suffices to initialize parameters uniformly within the \emph{unit} ball. This is in contrast to other search procedures employed for similar problems~\cite{mixture_two, shen2019iterative}, which are based on covering arguments and require a bound on the maximum norm of the unknown parameters.

\paragraph{Organization}
The rest of the paper is organized as follows. Section~\ref{sec:setup}
describes the problem setup and our methodology (including the AM
algorithm and initialization methods) in more detail. In
Section~\ref{sec:mainresults}, we present our main theoretical
results and their consequences, complementing our discussion with figures that verify that our results are borne out in simulation.
An overview of the main ideas behind our proofs is given in Section~\ref{sec:pfideas}. We
conclude the main paper with a discussion in Section~\ref{sec:discussion} of some related models and future directions.
Full proofs of our results are presented in the supplementary material in
Sections~\ref{sec:pf-thm1}-~\ref{sec:pf-thm4}, with further technical details relegated to the later sections of the appendix.

\paragraph{Notation}
For a positive integer $n$, let $[n] \defn \{1, 2, \ldots, n\}$. For a
finite set $S$, we use $|S|$ to denote its cardinality. All logarithms are to 
the natural base unless otherwise mentioned. For two
sequences $\{a_n\}_{n=1}^\infty$ and $\{b_n\}_{n=1}^\infty$, we write
$a_n \lesssim b_n$ if there is a universal constant $C$ such that $a_n
\leq C b_n$ for all $n \geq 1$. The relation $a_n \gtrsim b_n$ is
defined analogously, and we use $a_n \sim b_n$ to indicate that both $a_n \gtrsim b_n$ and $a_n \lesssim b_n$ hold simultaneously.  We use $c, C, c_1, c_2, \dots$ to denote
universal constants that may change from line to line. 
For a pair of vectors $(u, v)$, we let $u \otimes v \defn uv^\top$ denote their outer product. We use $\eucnorm{ \cdot }$ to denote the $\ell_2$ norm unless otherwise stated. Denote by $I_d$ the $d \times d$ identity matrix. 
We let $\ind{\Espace}$ denote the indicator of an event $\Espace$. Let $\sgn(t)$ denote the sign of a scalar $t$, with the convention that $\sgn(0) = 1$. 
Let $\lambda_i(\Gamma)$ denote the $i$-th largest eigenvalue of a symmetric matrix $\Gamma$.  Let $\Sd \defn \paren{ v \in \real^d: \| v \| = 1 }$ denote the unit shell in $d$-dimensions, and use $\Bd \defn \paren{ v \in \real^d: \| v \| \leq 1}$ to denote the $d$-dimensional unit ball.


\section{Background and problem formulation} \label{sec:setup}

In this section, we formally introduce the geometric parameters underlying the max-affine regression model, as well as the methodology we use to perform parameter estimation.

\subsection{Model and Geometric Parameters} \label{sec:model}

We work throughout with the observation model defined in equation~\eqref{eq:model}; recall our notation $P_X$ for the covariate distribution and that our noise is $\sigma$-sub-Gaussian. We let $\Xmat \in \real^{\numobs \times \numdim}$ denote the covariate matrix with row $i$ given by the vector $\meas_i$, and collect the responses in a vector $y \in \real^{\numobs}$.

Recall that $\appmeas_i = (\meas_i, \; 1) \in \real^{\numdim+1}$ for each $i \in [n]$; the matrix of appended covariates $\AppXmat \in \real^{\numobs \times (\numdim + 1)}$ is defined by appending a vector of ones to the right of the matrix $\Xmat$.
Our primary goal is to use the data $(\Xmat, y)$---or equivalently, the pair $(\AppXmat, y)$---to estimate the underlying parameters $\{ \appsigstar_j \}_{j = 1}^k$. 

An important consideration in achieving such a goal is the ``effective" sample size with which we observe the parameter $\appsigstar_j$. Toward that end, for $X \sim P_X$ independently of the parameters, let 
\begin{align} \label{eq:pi_def}
\pi_j(\appsigstar_1, \ldots, \appsigstar_k) \defn \Pr \left\{ \inprod{X}{\sigstar_j} + \bstar_j = \max_{j' \in [k]}\; \left( \inprod{X}{\sigstar_{j'}} + \bstar_{j'} \right) \right\}
\end{align} 
denote the probability with which the $j$-th parameter $\appsigstar_j = (\sigstar_j \; \; \bstar_j )$ attains the maximum. We primarily work with continuous distributions; in this case, the event on which more than one of the parameters attains the maximum has measure zero, except in the case where $\appsigstar_i = \appsigstar_j$ for some $i \neq j$. We explicitly disallow this case and assume that the parameters $\appsigstar_1, \ldots, \appsigstar_k$ are distinct.
Let 
\begin{align} \label{eq:pi_min_def}
\pi_{\min}(\appsigstar_1, \ldots, \appsigstar_k) \defn \min_{j \in [k]} \pi_j(\appsigstar_1, \ldots, \appsigstar_k),
\end{align} 
and assume that we have $\pi_{\min}(\appsigstar_1, \ldots, \appsigstar_k) > 0$; in other words, we ignore vacuous cases in which some parameter is never observed.
Roughly speaking, the sample size of the parameter that is observed most rarely is given by 
$
\min_{j \in [k]} \pi_j \numobs \sim \numobs \cdot \pi_{\min}(\appsigstar_1, \ldots, \appsigstar_k),
$
and so the error in estimating this parameter should naturally depend on $\pimin (\appsigstar_1, \ldots, \appsigstar_k)$. By definition, we always have 
$\pi_{\min}(\appsigstar_1, \ldots, \appsigstar_k) \leq 1/k$, and we say that the problem is ``well-balanced" if $\pi_{\min} (\appsigstar_1, \ldots, \appsigstar_k) \sim 1/k$.

Since we are interested in performing parameter estimation---as opposed to prediction---under the max-affine regression model, a few geometric quantities also appear in our bounds, and serve as natural notions of ``signal strength" and ``condition number" of the estimation problem. The signal strength is given by the minimum separation 
\[
\Delta (\appsigstar_1, \ldots, \appsigstar_k) = \min_{\substack{j, j' :  j \neq j'}} \eucnorm{ \sigstar_j - \sigstar_{j'} }^2;
\]
we also assume that $\Delta$ is strictly positive, since otherwise, a particular parameter is never observed. 
To denote a natural form of conditioning, define the quantities
\begin{align*}
\kappa_j (\appsigstar_1, \ldots, \appsigstar_k) = \frac{\max_{j' \neq j} \eucnorm{ \sigstar_j - \sigstar_{j'} }^2}{\min_{j' \neq j} \eucnorm{ \sigstar_j - \sigstar_{j'} }^2 }, \qquad \qquad \text{ with } \qquad \kappa (\appsigstar_1, \ldots, \appsigstar_k) = \max_{j \in [k] } \kappa_j (\appsigstar_1, \ldots, \appsigstar_k).
\end{align*}
Finally, let $\Deltamax(\appsigstar_1, \ldots, \appsigstar_k) \defn \max_{j \in [k]} \| \appsigstar_j \|$ denote the maximum norm of any unknown parameter.
We often use the shorthand 
\begin{align*}
\pimin &= \pi_{\min}(\appsigstar_1, \ldots, \appsigstar_k), \qquad \qquad \qquad \qquad \quad \; \; \; \Delta = \Delta (\appsigstar_1, \ldots, \appsigstar_k), \\
\kappa &= \kappa (\appsigstar_1, \ldots, \appsigstar_k), \qquad \qquad \text{ and} \qquad \qquad
\Deltamax = \Deltamax(\appsigstar_1, \ldots, \appsigstar_k)
\end{align*} 
when the true parameters $\appsigstar_1, \ldots, \appsigstar_k$ are clear from context.

\subsection{Methodology}

As discussed in the introduction, the most natural estimation procedure from i.i.d. samples $(\appmeas_i, y_i)_{i = 1}^n$ of the model~\eqref{eq:model} is the least
squares estimator~\eqref{intro:opt}. 
The following lemma (which does not seem to have been explicitly stated previously in the literature, except in the case $k = 2$~\cite{vandegeerthesis, lecue2013minimax}) proves that the least squares estimator $(\widehat{\beta}_1^{(\mathsf{ls})}, \dots, \widehat{\beta}_k^{(\mathsf{ls})})$ always exists. Note, however, that it will not be unique in general since any relabeling of a minimizer is also a minimizer.

\begin{lemma}\label{lse.exist}
The least squares estimator $\left(\widehat{\beta}_1^{(\mathsf{ls})}, \dots, \widehat{\beta}_k^{(\mathsf{ls})} \right)$ exists for every dataset $(\AppXmat, y)$.  
\end{lemma}
%
%
We postpone the proof of Lemma~\ref{lse.exist} to Appendix~\ref{pfs:lse}. In spite of the fact that the least squares estimator always exists, the problem~\eqref{intro:opt} is non-convex and NP-hard in general. The AM algorithm presents a tractable approach towards solving it in the statistical settings that we consider.


\subsubsection{Alternating Minimization}

We now formally describe the AM algorithm proposed by Magnani and Boyd~\cite{magnani_convex}. For each $\beta_1,\dots, \beta_k$, define the sets
\begin{equation} \label{eq:partition}
  S_j(\beta_1,\dots, \beta_k) := \left\{i \in [n] : j = \min \argmax_{1 \leq u \leq k} \left(\inprod{\appmeas_i}{\beta_u} \right) \right\} 
\end{equation}
for $j = 1, \dots, k$. In words, the set $S_j(\beta_1,\dots, \beta_k)$ contains the indices of samples on which parameter $\beta_j$ attains the maximum; in the case of a tie, samples having multiple parameters attaining the maximum are assigned to the set with the smallest corresponding index (i.e., ties are broken in the lexicographic order\footnote{In principle, it is sufficient to define the sets $S_j(\beta_1, \dots, \beta_k), j \in [k]$ as any partition of $[n]$ having the property that  $\inprod{\xi_i}{\beta_j} = \max_{u \in [k]} \inprod{\xi_i}{\beta_u}$ for every $j \in [k]$ and $i \in S_j(\beta_1, \dots, \beta_k)$; here ``any'' means that ties can be broken according to an arbitrary rule, and we have chosen this rule to be the lexicographic order in equation~\eqref{eq:partition}.}). Thus, the sets $\{ S_j(\beta_1,\dots, \beta_k) \}_{j = 1}^k$ define a  partition of $[n]$.  The AM algorithm employs an iterative scheme where one first constructs the partition $S_j \left(\beta_1^{(t)}, \dots, \beta_k^{(t)} \right)$ based on the current iterates $\beta_1^{(t)}, \dots, \beta_k^{(t)}$ and then calculates the next parameter estimate  $\beta_j^{(t+1)}$ by
a least squares fit to the dataset $\{(\appmeas_i, y_i), i \in S_j(\beta_1^{(t)}, \dots, \beta_k^{(t)})\}$. The algorithm (also described below as Algorithm~\ref{algo:AM}) is, clearly, quite intuitive and presents a natural approach to solving~\eqref{intro:opt}. 

\begin{algorithm}[ht] 
\KwIn{Data $\{\appmeas_i, y_i\}_{i=1}^n$; initial parameter estimates $\appsig{0}_1, \ldots, \appsig{0}_k$; number of iterations $T$.}

\KwOut{Final estimator of parameters $\widehat{\beta}_1, \ldots, \widehat{\beta}_k$.}

\nl Initialize $t \leftarrow 0$.


\Repeat{$t=T$}{
\nl Compute maximizing index sets
\begin{subequations}
\begin{align}
\label{eq:sets}
\Sup{t}_j = S_j(\appsig{t}_1,\dots, \appsig{t}_k),
\end{align}
for each $j \in [k]$, according to equation~\eqref{eq:partition}.

\nl Update 
\begin{align}
\label{eq:iteration}
\appsig{t+1}_j \in \argmin_{\beta \in \real^{\numdim + 1}} \sum_{i \in \Sup{t}_j} \left( y_i - \inprod{\appmeas_i}{\beta} \right)^2, 
\end{align}
\end{subequations}
for each $j \in [k]$. 
}

\nl Return $\widehat{\beta}_j = \appsig{T}_{j}$ for each $j \in [k]$.
\caption{Alternating minimization for estimating maximum of $k$ affine functions\label{algo:AM}}
\end{algorithm}
As a sanity check, Lemma~\ref{lse.fixed} (stated and proved in Appendix~\ref{pfs:lse}) shows
that the global least squares estimator~\eqref{intro:opt} is a fixed-point of this iterative scheme under a mild technical assumption. 

We also note that the AM algorithm was proposed by Soh~\cite{soh2019fitting-thesis} in the context of estimating structured convex sets from support function measurements. 
It should be viewed as a generalization of a classical algorithm for (real) phase retrieval due to Fienup~\cite{fienup1982phase}, which has been more recently analyzed in a series of
papers~\cite{netrapalli2013phase,walds} for Gaussian designs. While some analyses of AM algorithms assume sample-splitting across iterations (e.g.~\cite{netrapalli2013phase, mixture_many, zhang2019phase}), we consider the more practical variant of AM run without sample-splitting, since the update~\eqref{eq:sets}-\eqref{eq:iteration} is run on the full data $(\AppXmat, y)$ in every iteration.
The main contribution of this paper is the analysis of Algorithm~\ref{algo:AM} for max-affine regression under some assumptions on the covariate distribution.

\subsubsection{Initialization}
\label{subsec:init}

The alternating minimization algorithm described above requires an initialization. While the algorithm was proposed to be run from a random initialization with restarts~\cite{magnani_convex, soh2019fitting}, we propose to initialize the algorithm from parameter estimates that are sufficiently close to the optimal parameters. This is similar to multiple procedures to solve non-convex optimization problems in statistical settings (e.g.,~\cite{keshavan2010matrix, balakrishnan2017statistical}), that are based on iterative algorithms that exhibit \emph{local} convergence to the unknown parameters. Such algorithms are typically initialized by using a moment method, which (under various covariate assumptions) returns useful parameter estimates. %

\begin{algorithm}[h!] 
\KwIn{Data $\paren{ \appmeas_i, y_i }_{i=1}^{\numobs}$.}

\KwOut{Matrix $\Uhat \in \real^{d \times k}$ having orthonormal columns that (approximately) span the $k$ dimensional subspace spanned by the vectors $\sigstar_1, \ldots, \sigstar_k$.}

\nl Compute the quantities
\begin{align}
\Mhat_1 = \frac{2}{\numobs} \sum_{i = 1}^{\numobs/2} y_i \meas_i \qquad \text{ and } \qquad
\Mhat_2 = \frac{2}{\numobs} \sum_{i = 1}^{\numobs/2} y_i \left(\meas_i \meas_i^\top - I_d \right), \label{eq:momentcalc}
\end{align}
and let $\Mhat = \Mhat_1 \otimes \Mhat_1 + \Mhat_2$; here, $I_d$ denotes the $d \times d$ identity matrix and $\otimes$ denotes the outer product.


\nl Perform the eigendecomposition $\Mhat = \widehat{P} \widehat{\Lambda} \widehat{P}^\top$, and use the first $k$ columns of $\widehat{P}$ (corresponding to the $k$ largest eigenvalues) to form the matrix $\widehat{U} \in \real^{d \times k}$. Return $\Uhat$.
\caption{PCA for $k$-dimensional subspace initialization\label{algo:pca}}
\end{algorithm}

Our approach to the initialization problem is similar, in that we combine a moment method with random search in a lower-dimensional space. For convenience of analysis, we split the $\numobs$ samples into two equal parts---assume that $n$ is even without loss of generality---and perform each of the above steps on different samples so as to maintain independence between the two steps. 
The formal algorithm is presented in two parts as Algorithms~\ref{algo:pca} and~\ref{algo:rs}.

A few comments on the initialization are worth making. In related problems~\cite{sedghi2014provable, mixture_many, zhang2014spectral, chaganty2013spectral}, a combination of a second order and third order method (involving tensor decomposition) is employed to obtain parameter estimates in one shot. 
Take the problem of learning generalized linear models~\cite{sedghi2014provable} as an example; here, the analysis of the moment method relies on the link function being (at least) three times differentiable so that the population moment quantities can be explicitly computed. After showing that these expectations are closed form functions of the unknown parameters, matrix/tensor perturbation tools are then applied to show that the empirical moments concentrate about their population counterparts.
However, in our setting, the $\max$ function is not differentiable, and so it is not clear that higher order moments return reasonable estimates even in expectation since Stein's lemma (on which many of these results rely) is not applicable\footnote{A natural workaround is to use Stein's lemma on the infinitely differentiable ``softmax" surrogate function, but this approach also does not work for various technical reasons.} in this setting. Nevertheless, we show that the second order moment returns a $k$-dimensional subspace that is close to the true span of the parameters $\{ \sigstar_j \}_{j = 1}^k$; the degree of closeness depends only on the geometric properties of these parameters.

Second, note that while Algorithm~\ref{algo:pca} can in principle be implemented for any covariate distribution, the moment method in step 1 is sensible only for spherically symmetric distributions (see, e.g., Li~\cite{li1991sliced}). It is conceivable that replacing step 1 with score-function estimates for general covariate distributions---in the vein of related procedures by Babichev and Bach~\cite{babichev2018slice} and Sedghi et al.~\cite{sedghi2014provable}---still preserves the essence of our guarantees, but we do not pursue such an extension.

\begin{algorithm}[h] 
\KwIn{Data $\paren{ \appmeas_i, y_i }_{i=1}^{\numobs}$, subspace estimate $\Uhat \in \real^{d \times k}$ having orthonormal columns that (approximately) span the $k$ dimensional subspace spanned by the vectors $\sigstar_1, \ldots, \sigstar_k$, and number of random initializations $M \in \mathbb{N}$.}

\KwOut{Initial estimator of parameters $\appsig{0}_1, \ldots, \appsig{0}_k$.}

\nl Choose $M \cdot k$ random points $\nu^{\ell}_j$ i.i.d. for $\ell \in M$ and $j \in [k]$, each uniformly from the $(k+1)$-dimensional unit ball $\mathbb{B}^{k+1}$. Let 
\begin{align*}
\widehat{V} = 
\begin{bmatrix}
\widehat{U} & 0 \\
0 & 1
\end{bmatrix}
\end{align*}
be a matrix in $\real^{(d+1) \times (k+1)}$ having orthonormal columns.

\nl Compute the index
\begin{align*}
\ell^* \in \argmin_{\ell \in [M]} \; \frac{2}{n} \paren{ \min_{c \geq 0} \; \sum_{i = n/2 + 1}^n \left(y_i -  c \max_{j \in [k]}\; \inprod{\appmeas_i}{\Vhat \nu^{\ell}_j } \right)^2 }.
\end{align*}

\nl Return the $(d+1)$-dimensional parameters 
\begin{align*}
\appsig{0}_j = \widehat{V} \nu^{\ell^*}_j \qquad 
\text{ for each } j \in [k].
\end{align*}
\caption{Low-dimensional random search\label{algo:rs}}
\end{algorithm}

Let us also briefly discuss Algorithm~\ref{algo:rs}, which corresponds to performing random search in $(k+1) \cdot k$ dimensional space to obtain the final initialization. In addition to the random initialization employed in step 1 of this algorithm, we also use the mean squared error on a holdout set (corresponding to samples $n/2+1$ through $n$) to select the final parameter estimates. In particular, we evaluate the error in a scale-invariant fashion; the computation of the optimal constant $c$ in step~2 of the algorithm can be performed in closed form for each fixed index $\ell$, since for a pair of vectors $(u, v)$ having equal dimension, we have
\[
\argmin_{c \geq 0} \| u - c v \|^2 = \max\paren{ \frac{\inprod{u}{v} }{ \| v \|^2}, 0 }.
\]
A key parameter that governs the performance of our search procedure is the number of initializations $M$; we show in the sequel that it suffices to take $M$ to be a quantity that depends only on the number of affine pieces $k$, and on other geometric parameters in the problem.

Algorithm~\ref{algo:rs} can be viewed as a method to implement the least squares estimator in ambient dimension $k + 1$, since the random initializations are evaluated on the least squares criterion. Indeed, in Appendix~\ref{app:least-squares-init}, we 
analyze an alternative initialization procedure that combines the PCA subroutine of Algorithm~\ref{algo:pca} with such a least squares estimator to produce a similar guarantee, which involves analyzing the least squares algorithm with errors-in-variables. However, this initialization procedure is not computationally efficient, unlike the random search procedure presented above.


Our overall algorithm should be viewed as a slight variation of the AM algorithm with random restarts. It inherits similar empirical performance (see panel (b) of Figure~\ref{fig:overall} to follow), while significantly reducing the computational cost, since operations are now performed in ambient dimension $k + 1$, and the iterative AM algorithm is run only once overall. It also produces \emph{provable} parameter estimates, and as we show in the sequel, the number of random initializations $M$ can be set independently of the pair $(n, d)$.
Having stated the necessary background and described our methodology, we now proceed to statements and discussions of our main results.


\section{Main results} \label{sec:mainresults}

In this section, we present our main theoretical results for the methodology introduced in Section~\ref{sec:setup}.

\subsection{Local geometric convergence of alternating minimization}

We begin by establishing convergence results for the AM algorithm under two sets of covariate assumptions. 
Our first main result holds under a Gaussian assumption on the covariates.

Recall the definition of the parameters $(\pimin, \Delta, \kappa)$ introduced in Section~\ref{sec:setup}.

\begin{theorem} \label{thm:mt}
Suppose that the true parameters $\appsigstar_1, \ldots, \appsigstar_k$ are fixed, and that the covariates $\meas_i$ are drawn i.i.d. from the standard Gaussian distribution $\NORMAL(0, I_d)$. Then there exists a tuple of universal constants $(c_1, c_2)$ such that if the sample size satisfies the bound
\begin{align*} 
n \geq c_1 \max \paren{ d, 10 \log n } \max\paren{ \frac{k \kappa}{\pi_{\min}^3}, \; \frac{\log^2 (1 / \pi_{\min}) }{\pi_{\min}^3}, \log (n / d), \; \sigma^2 \frac{k^3}{\Delta \pi_{\min}^9} \log (k / \pi_{\min}^3) \log (n / d) },
\end{align*}
then for all initializations $\appsig{0}_1, \ldots, \appsig{0}_k$ satisfying the bound
\begin{subequations}
\begin{align}
\min_{c > 0} \; \max_{1 \leq j \neq j' \leq k} \; \frac{\left\| c \left( \appsig{0}_j - \appsig{0}_{j'} \right) - \left( \appsigstar_j -\appsigstar_{j'} \right) \right\|}{\| \sigstar_j -\sigstar_{j'} \| } \leq c_2 \frac{\pi_{\min}^3}{k \kappa} \log^{-3/2} \left( \frac{k \kappa}{\pimin^3} \right), \label{eq:init}
\end{align}
the estimation error at all iterations $t \geq 1$ is simultaneously bounded as
\begin{align}
\sum_{j = 1}^k \| \appsig{t}_j - \appsigstar_j \|^2 \leq \left ( \frac{3}{4} \right )^t \left( \sum_{j = 1}^k \| c^* \appsig{0}_j - \appsigstar_j \|^2 \right) + c_1 \sigma^2 \frac{kd}{\pi^3_{\min} n} \log(kd) \log(n/kd) \label{eq:thm1}
\end{align}
\end{subequations}
with probability exceeding $1 - c_2 \left(k \exp \left( - c_1 \numobs
    \frac{\pi_{\min}^4}{\log^2 (1 / \pi_{\min})} \right) +
  \frac{k^2}{n^7}\right)$. Here, the positive scalar $c^*$ minimizes the LHS
of inequality~\eqref{eq:init}. 
\end{theorem}
See Section~\ref{sec:pf-thm1} for a concise mathematical statement of the probability bound.

Let us interpret the various facets of Theorem~\ref{thm:mt}. As
mentioned before, it is a local convergence result, which requires the
initialization $\appsig{0}_1, \ldots, \appsig{0}_k$ to satisfy
condition~\eqref{eq:init}. In the well-balanced case (with $\pi_{\min}
\sim 1/k$) and treating $k$ as a fixed constant, the initialization condition~\eqref{eq:init} posits that
the parameters are a constant ``distance" from the true
parameters. 
Notably, closeness is measured in a relative sense, and
between pairwise differences of the parameter estimates as opposed to
the parameters themselves; the intuition for this is that the
initialization $\appsig{0}_1, \ldots, \appsig{0}_k$ induces the initial
partition of samples $S_1(\appsig{0}_1, \ldots,
  \appsig{0}_k), \ldots, S_1(\appsig{0}_1, \ldots,
  \appsig{0}_k)$, whose closeness to the true partition depends only
on the relative pairwise differences between parameters, and is also
invariant to a global scaling of the parameters.  
 It is also worth noting that local geometric convergence of the AM
 algorithm is guaranteed uniformly from \emph{all} initializations
 satisfying condition~\eqref{eq:init}. In particular, the
 initialization parameters are not additionally required to be
 independent of the covariates or noise, and this allows us to use the
 same $n$ samples for initialization of the parameters. 
 
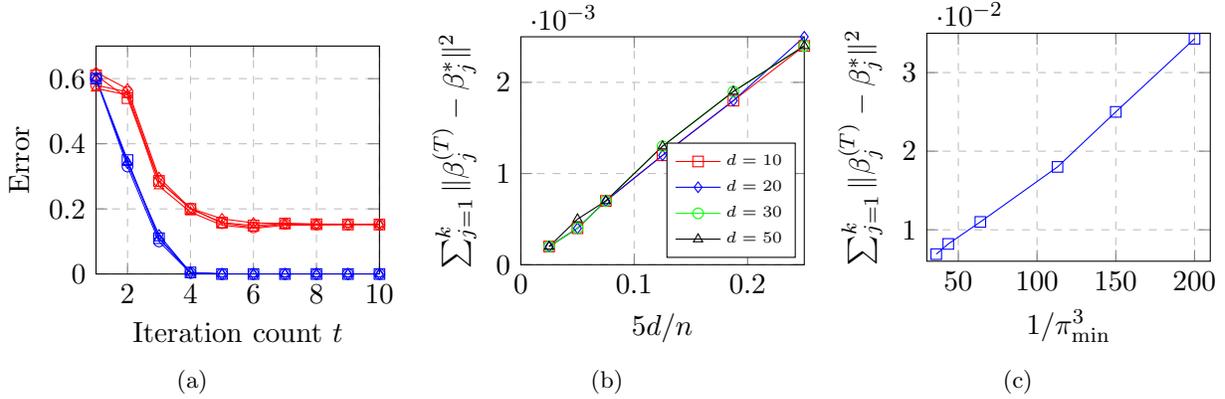
\begin{figure*}[t!]
\centering
\subfigure[]{
\begin{tikzpicture}
\begin{axis}[
    xlabel={Iteration count $t$},
    ylabel={Error},
    xmin=1, xmax=10,
    ymin=0, ymax=0.7,
    legend pos= north east,
    ymajorgrids=true,
    xmajorgrids=true,
    grid style=dashed,
]
 
\addplot[
    color=red,
    mark=square,
    ]
    coordinates {
    (1,0.61)(2,0.54 )(3,0.286)(4,0.20)(5,0.158)(6,0.149)(7,0.155)(8,0.1528)(9,0.152)(10,0.152)
    };

\addplot[
    color=red,
    mark=diamond,
    ]
    coordinates {
    (1,0.62)(2,0.5680 )(3, 0.2960)(4,0.2016)(5,0.1696)(6,0.1568)(7,0.1552)(8,0.1536)(9,0.153)(10,0.153)
    };
    
  \addplot[
    color=red,
    mark=o,
    ]
    coordinates {
    (1,0.58 )(2, 0.559 )(3,0.285)(4,0.201)(5, 0.156 )(6,0.143)(7,0.154 )(8,0.152)(9, 0.152)(10,0.152)
    };

    \addplot[
    color=red,
    mark=triangle,
    ]
    coordinates {
    
    (1,0.57 )(2,0.55)(3, 0.27 )(4, 0.19 )(5, 0.15 )(6, 0.14)(7,   0.15)(8, 0.15)(9,0.1525)(10, 0.1525)
    };

    \addplot[
    color=blue,
    mark=square,
    ]
    coordinates {
    (1,0.6)(2, 0.35)(3,0.11)(4,0.005)(5,0)(6, 0)(7, 0)(8, 0)(9, 0)(10, 0)
    };
    
\addplot[
    color=blue,
    mark=diamond,
    ]
    coordinates {
    (1,0.6)(2, 0.34)(3,0.12)(4,0.004)(5,0)(6, 0)(7, 0)(8, 0)(9, 0)(10, 0)
    };
    
    \addplot[
    color=blue,
    mark=o,
    ]
    coordinates {
    (1,0.6)(2, 0.33)(3,0.10)(4,0.003)(5,0)(6, 0)(7, 0)(8, 0)(9, 0)(10, 0)
    };

    \addplot[
    color=blue,
    mark=triangle,
    ]
    coordinates {
    (1,0.6)(2, 0.345)(3,0.11)(4,0.004)(5,0)(6, 0)(7, 0)(8, 0)(9, 0)(10, 0)
    };
\end{axis}
\end{tikzpicture}}
\subfigure[]{
\begin{tikzpicture}
\begin{axis}[
    legend style={font=\tiny},
    xlabel={$5d/n$},
    ylabel={$\sum_{j=1}^k  \| \appsig{T}_j - \appsigstar_j \|^2$ },
    xmin=0, xmax=0.25,
    ymin=0, ymax=0.0025,
    legend pos=south east,
    ymajorgrids=true,
    xmajorgrids=true,
    grid style=dashed,
]
 
\addplot[
    color=red,
    mark=square,
    ]
    coordinates {
    (0.0250,0.0002)    (0.0500,0.0004)  (0.0750,0.0007)  (0.1250,0.0012)(0.1875,0.0018)(0.2500,0.0024)
    };
    \addlegendentry{$d=10$}

\addplot[
    color=blue,
    mark=diamond,
    ]
    coordinates {
    (0.0250,0.0002)    (0.0500,0.0004)  (0.0750,0.0007)  (0.1250,0.0012)(0.1875,0.0018)(0.2500,0.0025)
    };
    \addlegendentry{$d=20$}

    \addplot[
    color=green,
    mark=o,
    ]
    coordinates {
    (0.0250,0.0002)    (0.0500,0.0004)  (0.0750,0.0007)  (0.1250,0.0013)(0.1875,0.0019)(0.2500,0.0024)
    };
    \addlegendentry{$d=30$}
    
    \addplot[
    color=black,
    mark=triangle,
    ]
    coordinates {
    (0.0250,0.0002)    (0.0500,0.0005)  (0.0750,0.0007)  (0.1250,0.0013)(0.1875,0.0019)(0.2500,0.0024)
    };
    \addlegendentry{$d=50$}

\end{axis}
\end{tikzpicture}}
\subfigure[]{
\begin{tikzpicture}
\begin{axis}[
    legend style={font=\tiny},
    xlabel={$1/\pi_{\min}^3 $},
    ylabel={$\sum_{j=1}^k  \| \appsig{T}_j - \appsigstar_j \|^2$ },
    xmin=30, xmax=210,
    ymin=0.006, ymax=0.035,
    legend pos=south east,
    ymajorgrids=true,
    xmajorgrids=true,
    grid style=dashed,
]
 
\addplot[
    color=blue,
    mark=square,
    ]
    coordinates {
    (36,0.0069) (43.62,0.0082)  (64,0.011)  (113,0.018) (150,0.025) (200,0.0343)
    };
    \end{axis}
\end{tikzpicture}}
\caption{Convergence of the AM with Gaussian covariates---in panel (a), we plot the optimization error (in blue) $\sum_{j=1}^k \| \appsig{t}_j-\appsig{T}_j \|^2$ and the deviation from the true parameters (in red) $\sum_{j=1}^k \norms{\appsig{t}_j -\appsigstar_j}^2/\sigma^2$ over iterations $t$ for different $\sigma$ ($0.15,0.25,0.4, 0.5$), with $k=5$, $d=100$, $T=50$ and $n=5d$, and averaged over $50$ trials. Panel (b) shows that the estimation error at $T=50$ scales at the parametric rate $d/n$, where we have chosen a fixed $k=5$ and $\sigma =0.25$. Panel (c) shows the variation of this error as a function of $\pi_{\min}$ where we fix $k=3, d=2,n=10^3, \sigma = 0.4$.}
\label{fig:conv_good}
\end{figure*}

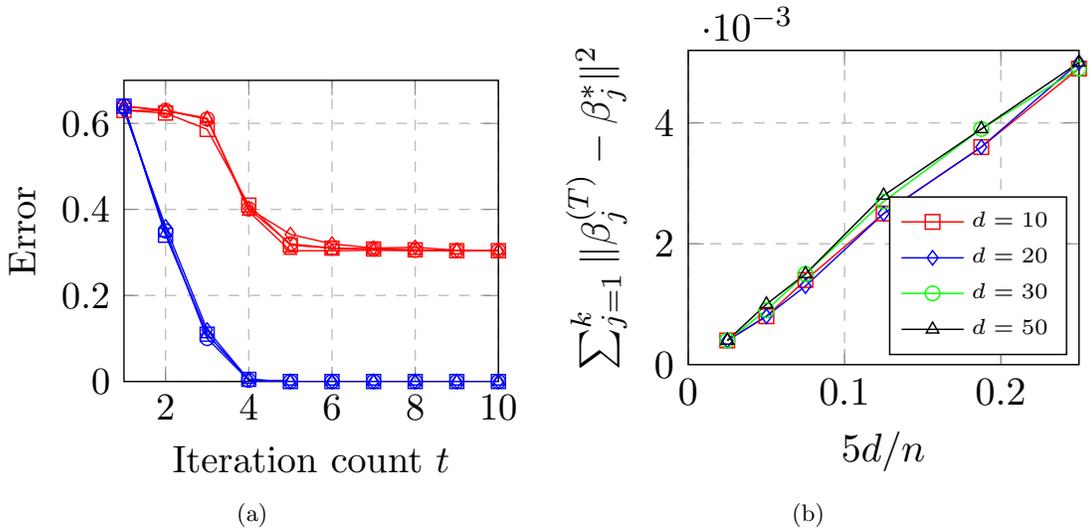
\begin{figure*}[t!]
\centering
\subfigure[]{\resizebox {0.44 \textwidth} {!}{
\begin{tikzpicture}
\begin{axis}[
    xlabel={Iteration count $t$},
    ylabel={Error},
    xmin=1, xmax=10,
    ymin=0, ymax=0.7,
    legend pos= north east,
    ymajorgrids=true,
    xmajorgrids=true,
    grid style=dashed,
]
 
\addplot[
    color=red,
    mark=square,
    ]
    coordinates {
    (1,0.63)(2,0.624 )(3,0.586)(4,0.41)(5,0.317)(6,0.310)(7,0.308)(8,0.306)(9,0.304)(10,0.304)
    };

\addplot[
    color=red,
    mark=diamond,
    ]
    coordinates {
    (1,0.63)(2,0.63 )(3, 0.61)(4,0.402)(5,0.342)(6,0.32)(7,0.310)(8,0.312)(9,0.305)(10,0.304)
    };
    
  \addplot[
    color=red,
    mark=o,
    ]
    coordinates {
    (1,0.64 )(2, 0.63 )(3,0.610)(4,0.401)(5, 0.320 )(6,0.310)(7,0.308 )(8,0.305)(9, 0.304)(10,0.304)
    };

    \addplot[
    color=red,
    mark=triangle,
    ]
    coordinates {
    
    (1,0.64 )(2,0.63)(3, 0.612 )(4, 0.395 )(5, 0.304 )(6, 0.304)(7,   0.305)(8, 0.304)(9,0.305)(10, 0.304)
    };

    \addplot[
    color=blue,
    mark=square,
    ]
    coordinates {
    (1,0.64)(2, 0.34)(3,0.11)(4,0.005)(5,0)(6, 0)(7, 0)(8, 0)(9, 0)(10, 0)
    };
    
\addplot[
    color=blue,
    mark=diamond,
    ]
    coordinates {
    (1,0.63)(2, 0.36)(3,0.12)(4,0.004)(5,0)(6, 0)(7, 0)(8, 0)(9, 0)(10, 0)
    };
    
    \addplot[
    color=blue,
    mark=o,
    ]
    coordinates {
    (1,0.64)(2, 0.35)(3,0.10)(4,0.003)(5,0)(6, 0)(7, 0)(8, 0)(9, 0)(10, 0)
    };

    \addplot[
    color=blue,
    mark=triangle,
    ]
    coordinates {
    (1,0.64)(2, 0.345)(3,0.11)(4,0.004)(5,0)(6, 0)(7, 0)(8, 0)(9, 0)(10, 0)
    };
\end{axis}
\end{tikzpicture}}}
\subfigure[]{\resizebox {0.44 \textwidth} {!}{
\begin{tikzpicture}
\begin{axis}[
    legend style={font=\tiny},
    xlabel={$5 d/n$},
    ylabel={$\sum_{j=1}^k  \| \appsig{T}_j - \appsigstar_j \|^2$ },
    xmin=0, xmax=0.25,
    ymin=0, ymax=0.0052,
    legend pos=south east,
    ymajorgrids=true,
    xmajorgrids=true,
    grid style=dashed,
]
 
\addplot[
    color=red,
    mark=square,
    ]
    coordinates {
    (0.0250,0.0004)    (0.0500,0.0008)  (0.0750,0.0014)  (0.1250,0.0025)(0.1875,0.0036)(0.2500,0.0049)
    };
    \addlegendentry{$d=10$}

\addplot[
    color=blue,
    mark=diamond,
    ]
    coordinates {
    (0.0250,0.0004)    (0.0500,0.0008)  (0.0750,0.0013)  (0.1250,0.0025)(0.1875,0.0036)(0.2500,0.0050)
    };
    \addlegendentry{$d=20$}

    \addplot[
    color=green,
    mark=o,
    ]
    coordinates {
    (0.0250,0.0004)    (0.0500,0.0009)  (0.0750,0.0015)  (0.1250,0.0027)(0.1875,0.0039)(0.2500,0.0049)
    };
    \addlegendentry{$d=30$}
    
    \addplot[
    color=black,
    mark=triangle,
    ]
    coordinates {
    (0.0250,0.0004)    (0.0500,0.0010)  (0.0750,0.0015)  (0.1250,0.0028)(0.1875,0.0039)(0.2500,0.0050)
    };
    \addlegendentry{$d=50$}

\end{axis}
\end{tikzpicture}}}
\caption{Convergence of AM when the covariates are drawn i.i.d. from the distribution $\mathsf{Unif}[-\sqrt{3},\sqrt{3}]^{\otimes d}$; parameter settings are identical to those used in Figure~\ref{fig:conv_good}.}
\label{fig:conv_good_gen}
\end{figure*}

Let us now turn our attention to the bound~\eqref{eq:thm1}, which consists of two terms. In the limit $t \to \infty$, the final parameters provide an estimate of the true parameters that is accurate to within the second term of the bound~\eqref{eq:thm1}. Up to a constant, this is the statistical error term 
\begin{align}
\delta_{n, \sigma}(d, k, \pi_{\min}) = \sigma^2 \frac{kd}{\pi^3_{\min} n} \log(kd) \log(n/kd) \label{eq:stat-err}
\end{align}
that converges to $0$ as $n \to \infty$, thereby providing a consistent estimate in the large sample limit. Notice that the dependence of $\delta_{n, \sigma}(d, k, \pi_{\min})$ on the tuple $(\sigma, d, n)$ is minimax-optimal up to the logarithmic factor $\log(n/d)$, since a matching lower bound can be proved for the linear regression problem when $k = 1$. In Proposition~\ref{prop:minimax}, (see Appendix~\ref{app:lbs}) we also show a parametric lower bound on the minimax estimation error for general $k$, of the order $\sigma^2 kd/n$. Panel (b) of Figure~\ref{fig:conv_good} verifies in a simulation that the statistical error depends linearly on $d/n$. The dependence of the statistical error on the pair $(k, \pi_{\min})$ is more involved, and we do not yet know if these are optimal. As discussed before, a linear dependence of $\pimin$ is immediate from a sample-size argument; the cubic dependence arises because the sub-matrices of $\Xi$ chosen over the course of the algorithm are not always well-conditioned, and their condition number scales (at most) as $\pimin^2$. In Appendix~\ref{sec:covariance}, we show a low-dimensional example (with $d = 2$ and $k = 3$) in which the least squares estimator incurs a parameter estimation error of the order $\frac{1}{\pimin^3 n}$ even when provided with the \emph{true} partition of covariates $\{ S_j (\appsigstar_1, \ldots, \appsigstar_k ) \}_{j = 1}^3$. While this does not constitute an information theoretic lower bound, it provides strong evidence to suggest that our dependence on $\pimin$ is optimal at least when viewed in isolation. We verify this intuition via simulation: in panel (c) of Figure~\ref{fig:conv_good_gen}, we observe that on this example, the  error of the final AM iterate varies linearly with the quantity $1/\pi_{\min}^3$.

%

The first term of the bound~\eqref{eq:thm1} is an optimization error that is best interpreted in the noiseless case $\sigma = 0$, wherein the parameters $\appsig{t}_1, \ldots, \appsig{t}_k$ converge at a geometric rate to the true parameters $\appsigstar_1, \ldots, \appsigstar_k$, as verified in panel (a) of Figure~\ref{fig:conv_good_gen}. In particular, in the noiseless case, we obtain exact recovery of the parameters provided $n \geq C \frac{kd}{\pi_{\min}^3} \log (n / d)$. Thus, the ``sample complexity" of parameter recovery is linear in the dimension $d$, which is optimal. In the well-balanced case, the dependence on $k$ is quartic, but lower bounds based on parameter counting suggest that the true dependence ought to be linear. Again, we are not aware of whether the dependence on $\pi_{\min}$ in the noiseless case is optimal; our simulations in Appendix~\ref{app:sim} suggest that the sample complexity depends inversely on $\pimin$, and so closing this gap is an interesting open problem\footnote{If we are interested solely in the noiseless case, then the sample complexity bound can be sharpened by using a technique similar to the proof of Theorem~\ref{thm:sbgen}; we avoid this since it leads to a strictly worse final error when $\sigma > 0$.}.
When $\sigma > 0$, we have an overall sample size requirement
\begin{align} \label{eq:nam}
n \geq c \max \paren{ \frac{kd \kappa}{\pi_{\min}^3}, \; \frac{d \log^2 (1 / \pi_{\min}) }{\pi_{\min}^3}, d \log (n / d), \; \sigma^2 \frac{k^3 d}{\Delta \pi_{\min}^9} \log (k / \pi_{\min}^3) \log (n / d) } \defn n_{\mathsf{AM}}(c).
\end{align}

As a final remark, note that Theorem~\ref{thm:mt} holds under Gaussian covariates and when the true parameters 
$\appsigstar_1, \ldots, \appsigstar_k$ are fixed independently of the covariates. We now
show that both of these features of the result can be relaxed, i.e., AM converges geometrically even
under a milder covariate assumption, and this convergence occurs \emph{for all} true parameters that are geometrically similar.
Our covariate assumption relies on the following technical definition.
\begin{definition}(Small-ball) \label{def:sball}
A distribution $P_X$ satisfies a $(\sball, \cs)$-small-ball property
if we have
\begin{align} \label{eq:sb}
\sup_{u \in \Sd, \; w \in \real } \; \Pr \paren{ (\inprod{X}{u}
  + w )^2 \leq \delta } &\leq (\cs \delta)^{\sball} \; \; \text{ for } X \sim P_X \text{ and each }
                          \delta > 0. 
\end{align}
\end{definition}

The small-ball properties of various classes of distributions have
been studied extensively in the probability
literature~\cite{paouris2012small, rudelson2014small}; for instance, a straightforward calculation yields that provided the density of the random variable $\inprod{X}{u}$ is bounded by $\sqrt{c}$ for each $u \in \Sd$, the distribution $P_X$ satisfies the $(1/2, c)$-small ball property. We now present
our assumption on the covariate distribution; recall that a random vector $X \in \real^d$ is said to be $\eta$-sub-Gaussian if
$
\sup_{u \in \Sd} \EE [ \exp ( \lambda \inprod{X}{u} ) ] \leq \exp \left( \frac{ \lambda^2 \eta^2}{2} \right)$ for each $\lambda \in \real$.

\begin{assumption} \label{assptn:gen}
The distribution $P_X$ is isotropic, $\eta$-sub-Gaussian, and satisfies a $(\sball, \cs)$ small-ball condition.
\end{assumption}


Before stating our theorem, let us briefly state a few immediate examples of random variables whose distributions
satisfy Assumption~\ref{assptn:gen} with particular values of the tuple $(\eta, \sball, \cs)$. 

\paragraph{Compactly supported log-concave random vectors:} 
As we verify in Appendix~\ref{sec:sblc} by applying the Carbery-Wright
inequality for thresholds of polynomial
functions~\cite{carbery2001distributional}, log-concave random vectors
satisfy the the small ball conditions with $(\sball, \cs) = (1/2, C)$
for an absolute constant $C$. Boundedness of the RV further implies
sub-Gaussianity. As a specific example, consider a random vector $X$
with each entry drawn i.i.d. according to the distribution
$\mathsf{Unif}[-\sqrt{3}, \sqrt{3}]$. The associated distribution
$P_X$ is isotropic by definition, and has $(\eta, \sball, \cs) = (12,
1/2, C)$. 
Similarly, any
other uniform distribution on a bounded, isotropic convex set would also satisfy
Assumption~\ref{assptn:gen}. 
\hfill 
$\clubsuit$  

\paragraph{Standard Gaussian random vector:}
This is the canonical example of a sub-Gaussian RV whose distribution
satisfies a small-ball condition. As we verify in
Appendix~\ref{sec:sbg} with $\chi^2$ tail bounds, the standard
Gaussian satisfies $(\eta, \sball, \cs) = (1, 1/2,
e)$. 
\hfill 
$\clubsuit$  
\vspace{3mm}


Thus, Assumption~\ref{assptn:gen} is strictly more general than the
Gaussian covariate assumption underlying Theorem~\ref{thm:mt}. It is
also important to note that Assumption~\ref{assptn:gen} allows a
larger class of distributions than even log-concave distributions;
heuristically speaking, the small-ball condition only disallows
distributions that are significantly ``peakier" than the Gaussian
distribution, and the sub-Gaussian condition disallows heavy-tailed
distributions. 
 
Our second goal was to prove a result that holds uniformly for all
true parameters $\appsigstar_j, j = 1, \dots, k$ once the covariates
have been drawn. However, this is 
clearly impossible in a general sense, since we cannot hope to obtain
consistent estimates if some parameters are never
observed in the sample. A workaround is 
to hold certain geometric quantities fixed while sweeping over all
possible allowable parameters $\appsigstar_j, j = 1, \dots,
k$. Accordingly, for each triple of positive scalars $(\pi, \Delta,
\kappa)$, we define the set of ``admissible" true parameters 
as 
\[
\Bvol(\pi, \Delta, \kappa) = \paren{ \beta_1, \ldots, \beta_k: \pimin(\beta_1, \ldots, \beta_k) \geq \pi \; , \; \Delta(\beta_1, \ldots, \beta_k) \geq \Delta \;, \; \kappa(\beta_1, \ldots, \beta_k) \leq \kappa }.
\]
For each pair $1 \leq i \neq j \leq k$ and $t \geq 0$, we also use the shorthand 
$
\vstar_{i, j} = \appsigstar_{i} - \appsigstar_{j}
$ and  
$\vvec{t}_{i, j} = \appsig{t}_i - \appsig{t}_j$ to denote the pairwise differences between parameters.


\begin{theorem} \label{thm:sbgen}
Suppose that Assumption~\ref{assptn:gen} holds. Then there exists a pair of universal constants $(c_1, c_2)$ and constants $(\Csig^{(1)}, \Csig^{(2)})$ depending only on the triple $(\eta, \sball, \cs)$ such that if the sample size satisfies the bound
\begin{align*} 
n \geq \Csig^{(1)} \max \paren{ d, 10 \log n } \cdot \frac{k \kappa}{\pi_{\min}^{1 + 2 \sball^{-1}}} \log (n / d)
\end{align*}
then for
all true parameters $\appsigstar_1, \ldots, \appsigstar_k \in
\Bvol(\pimin, \Delta, \kappa)$ and all initializations satisfying  
\begin{subequations}
\begin{align}
\min_{c > 0} \; \max_{1 \leq j \neq j' \leq k} \; \frac{\left\| c \vvec{0}_{j, j'} - \vstar_{j, j'} \right\|}{\| \sigstar_j -\sigstar_{j'} \| } &\leq \Csig^{(2)} \left(\frac{\pi_{\min}^{1 + 2 \sball^{-1}}}{k \kappa}\right)^{\sball^{-1}}  \log^{1 + \sball^{-1}} \left( \frac{k \kappa}{ \pi_{\min}^{1 + 2 \sball^{-1} } } \right), \label{eq:init2gen}
\end{align}
the estimation error at all iterations $t \geq 1$ is simultaneously bounded as
\begin{align}
\sum_{j = 1}^k \| \appsig{t}_j - \appsigstar_j \|^2 \leq \left ( \frac{3}{4} \right )^t \left( \sum_{j = 1}^k \| c^* \appsig{0}_j - \appsigstar_j \|^2 \right) + \Csig^{(1)} \cdot \sigma^2 \frac{kd}{n \pimin^{1 + 2\sball^{-1} }} \log(kd) \log(n/kd) \label{eq:thm1gen}
\end{align}
\end{subequations}
with probability exceeding $1 - c_1 \paren{ \frac{k^2}{n^7} + \exp \left(- c_2 n \pi_{\min}^2 \right) }$. Here, the positive scalar $c^*$ minimizes the LHS of inequality~\eqref{eq:init2gen}.
\end{theorem}
Once again, a concise mathematical statement of the probability bound
given by the above theorem is postponed to Section~\ref{sec:pf-thm2}.
The structure and statement of Theorem~\ref{thm:sbgen} exactly parallels that of Theorem~\ref{thm:mt}, so we restrict our discussion of it to the notable differences between the theorems.

The most direct comparison between the theorems is obtained by
specializing Theorem~\ref{thm:sbgen} to the Gaussian setting, in which
$\eta = 1$ and $\sball = 1/2$. In this case, all terms of the form
$\pimin^3$ in Theorem~\ref{thm:mt} are replaced by terms of the form
$\pimin^5$. In particular, we see that the initialization
condition~\eqref{eq:init2gen} is more stringent than the corresponding
condition~\eqref{eq:init}. The final statistical rate of the estimate
(corresponding to the limit $t \to \infty$) now attains an estimation
error that is a factor $\pimin^{-2}$ higher than the corresponding
rate of Theorem~\ref{thm:mt}. The sample size requirement is similarly
affected. 
On the other hand, the geometric convergence result~\eqref{eq:thm1gen}
now holds uniformly for all true parameters $\appsigstar_1, \ldots, \appsigstar_k \in \Bvol(\pimin, \Delta, \kappa)$, as opposed to the
result~\eqref{eq:thm1}, which holds when the true parameters are held
fixed. The more stringent initialization condition and sample size
requirements can therefore be viewed as the price to pay for the
convergence of AM in this universal sense. Notably, the dependence on
all other parameters of the problem remains unchanged. 

In Figure~\ref{fig:conv_good}, we verify that for independent, isotropic
covariates chosen uniformly from a symmetric interval---such a distribution is
compactly supported and log-concave, and therefore satisfies Assumption~\ref{assptn:gen}---the AM algorithm exhibits the properties predicted by Theorem~\ref{thm:sbgen}. In a complementary direction, we perform some numerical experiments Appendix~\ref{app:sim} to test the necessity of covariate Assumption~\ref{assptn:gen}
for convergence of the AM algorithm. In particular, we find that convergence does occur
when the covariates are drawn from a Rademacher or (centered) binomial distribution,
but only when the number of samples scales quadratically with dimension $d$. Establishing necessary conditions for the convergence of the AM algorithm is an interesting direction for future work.

A notable consequence of Theorem~\ref{thm:sbgen} is that it can be
applied to the phase retrieval problem---in which results are usually
proved uniformly over all true parameters~\cite{candes2015phase, chen2015solving}---to yield a convergence 
result under general distributional assumptions on the covariates. In
particular, setting $\pi_{\min} = 1/2$ and $k = 2$ yields a local
linear convergence result for the AM algorithm of the
Gershberg-Saxton-Fienup type (presented for completeness as
Algorithm~\ref{algo:PR}) uniformly \emph{for all} $\sigstar$
provided the covariates satisfy a small-ball condition. Other
algorithms for phase retrieval have also been shown to succeed under
such small-ball assumptions~\cite{eldar2014phase, duchi2017solving}. The
  following result holds under the observation model~\eqref{prmd}.  

\begin{algorithm}[ht] 
\KwIn{Data $\{\meas_i, y_i\}_{i=1}^n$; initial parameter estimate $\sig{0} \in \real^d$; number of iterations $T$.}

\KwOut{Final estimator $\widehat{\theta}$.}

\nl Initialize $t \leftarrow 0$.


\Repeat{$t=T$}{
\nl Compute vector of signs $s^{(t)}$ by computing its $i$-th entry as
\begin{subequations}
\begin{align}
\label{eq:signs}
s^{(t)}_i = \sgn( \inprod{\meas_i}{\sig{t}} ) \; \; \text{ for each } i \in [n].
\end{align}

\nl Update 
\begin{align}
\label{eq:PRnext}
\sig{t+1} = \arg \min_{\theta \in \real^{\numdim}} \sum_{i = 1}^n \left( y_i - s^{(t)}_i \inprod{\meas_i}{\theta} \right)^2. 
\end{align}
\end{subequations}
}

\nl Return $\widehat{\theta} = \sig{T}$.
\caption{Alternating minimization for real phase retrieval\label{algo:PR}}
\end{algorithm}

\begin{corollary} \label{cor:PR}
Suppose that Assumption~\ref{assptn:gen} holds. Then there exists a universal constant $c_1$ and a pair of constants $(\Csig^{(1)}, \Csig^{(2)})$ depending only on the triple $(\eta, \sball, \cs)$ such that if
\begin{align} \label{eq:n-req2}
n \geq \Csig^{(1)} \max \paren{ d, 10 \log n } \log (n / d)
\end{align}
then for all true parameters $\sigstar \in \real^d$ and all initializations $\sig{0}$ satisfying 
\begin{subequations}
\begin{align}
\min_{c > 0} \min_{s \in \{ -1, 1\} } \frac{\left\| c \sig{0} - s \sigstar \right\|}{\| \sigstar \| } &\leq \Csig^{(2)},\label{eq:init2genPR}
\end{align}
the estimation error at all iterations $t \geq 1$ is simultaneously bounded as
\begin{align}
\min_{s \in \{ -1, 1\} } \; \| \sig{t} - s \sigstar_j \|^2 \leq \left ( \frac{3}{4} \right )^t  \min_{s \in \{ -1, 1\} } \| \sig{0} - s \sigstar \|^2  + c_1 \sigma^2 \frac{d}{n} \log(n/d) \label{eq:thm1genPR}
\end{align}
\end{subequations}
with probability exceeding $1 -  c_1 n^{-7}$.
\end{corollary}

The proof of Corollary~\ref{cor:PR} is provided in Appendix~\ref{sec:pf-cor1}. Let us conclude with a brief comparison with the sharpest existing local convergence result for AM in the phase retrieval problem due to Waldspurger~\cite{walds}, which holds for Gaussian covariates and in the noiseless setting\footnote{To be fair, Waldspurger~\cite{walds} deals with the complex phase retrieval problem, whose analysis is significantly more complicated than the real phase retrieval problem considered here.}.
Specializing Corollary~\ref{cor:PR} 
to the noiseless setting, we
observe that
provided the ratio $n/d$ is larger than a fixed constant (that depends
only on the triple $(\eta, \sball, \cs)$), we obtain exact recovery of
the underlying parameter, up to a global sign, with high probability
provided the covariates (or ``measurement vectors" as they are called
in the signal processing literature) are sub-Gaussian and satisfy a
small-ball condition.
To the best of our knowledge, prior work on the AM algorithm had
not established provable guarantees for non-Gaussian covariates even in the noiseless setting.
In the noisy case, Corollary~\ref{cor:PR} guarantees convergence of the iterates to a small
neighborhood around either $\thetastar$ or $-\thetastar$, and the size
of this neighborhood is within a logarithmic factor of being minimax
optimal~\cite{chen2015solving,lecue2013minimax}. Once again, to the best of our knowledge, guarantees for AM as applied to noisy phase retrieval did not exist in the literature.


Having established guarantees on the AM algorithm, we now turn to establishing guarantees for our initialization methodology. 

\subsection{Initialization}

In this section, we provide guarantees on the initialization method
described in Algorithms~\ref{algo:pca} and~\ref{algo:rs} in
Theorems~\ref{thm:pca} and~\ref{thm:rs}, respectively. As discussed
before, we do not pursue a generalization to arbitrary covariate
distributions in this case, and restrict ourselves to the setting of
Gaussian covariates with $x_i \stackrel{i.i.d.}{\sim} N(0, 
  I_d)$. 

Consider the matrices $\widehat{U}$ and $\Mhat$ defined in Algorithm
\ref{algo:pca}. Algorithm~\ref{algo:pca} is a moment method: we
extract the top $k$ 
principal components of a carefully chosen moment statistic of the
data to obtain a subspace estimate $\widehat{U}$. Spectral algorithms
such as these have been used to obtain initializations in a wide
variety of non-convex problems~\cite{mixture_many, zhong2017recovery,
  chen2015solving} to obtain an accurate estimate of the subspace
spanned by the unknown parameters.  It is well-known that the
performance of the algorithm in recovering a $k$-dimensional subspace
depends on $\lambda_k (\EE[\Mhat])$, which is the $k$-th largest eigenvalue of the population moment $\EE[\Mhat]$. 
We show in the proof (see the discussion following Lemma~\ref{lem:expmom}) that there is a strictly positive scalar $\gamma$ such that
\begin{align} \label{eq:gammadef}
\lambda_k (\EE[\Mhat]) \geq \gamma.
\end{align}  
It should be stressed that we obtain an explicit expression for
$\gamma$ as a function of the various problem parameters (in
equation~\eqref{eq:gammadef-1} of the proof) that is, a priori\footnote{While this 
may seem surprising---after all, the unknown parameters $\sigstar_1, \ldots, \sigstar_k$ live in dimension $d$---all the interesting action is confined to the $k$ dimensional subspace spanned by these parameters and $\gamma$ is a function of the geometry induced by the parameters on this subspace.},
independent of the ambient dimension~$\numdim$.


This characterization
is the main novelty of our contribution, and allows us to establish
the following guarantee on the PCA algorithm. We let $\Ustar \in
\R^{\numdim \times k}$ denote a matrix whose orthonormal columns span
the linear subspace spanned by the vectors $\sigstar_1, \ldots,
\sigstar_k$, and 
define the quantity
\begin{align} \label{eq:psi}
\varsigma \defn \max_{j \in [k]} \paren{ \| \thetastar_j \|_1 + \bstar_j }.
\end{align}

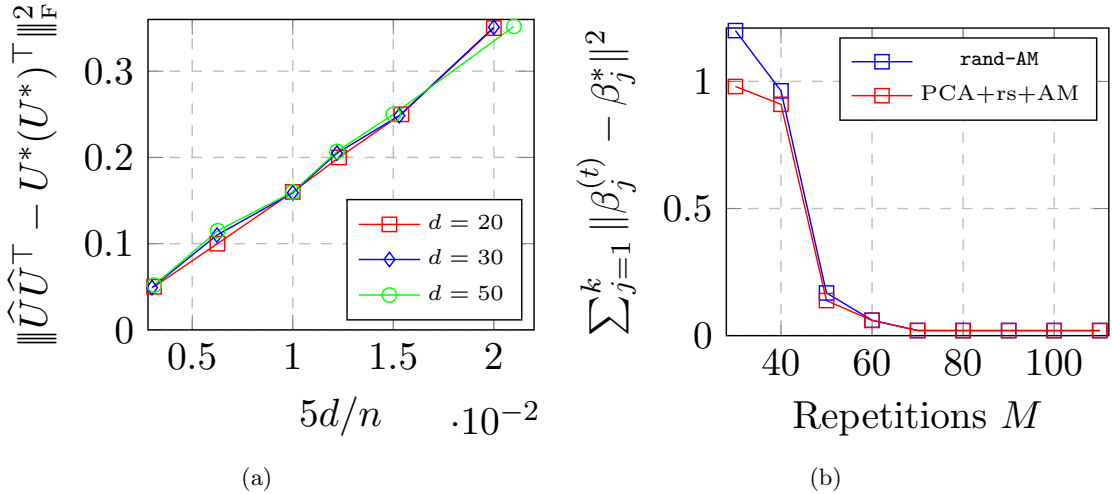
\begin{figure*}[t!]
\centering
\subfigure[]{\resizebox {0.45 \textwidth} {!}{
\begin{tikzpicture}
\begin{axis}[
    legend style={font=\tiny},
    xlabel={$5d/n$},
    ylabel={$ \fronorm{\widehat{U}\widehat{U}^\top - U^* (U^*)^\top}^2$ },
    xmin=0.0028, xmax=0.022,
    ymin=0, ymax=0.36,
    legend pos=south east,
    ymajorgrids=true,
    xmajorgrids=true,
    grid style=dashed,
]
 
\addplot[
    color=red,
    mark=square,
    ]
    coordinates {
    (0.0031,0.05)    (0.00625,0.1)  (0.01,0.16)  (0.0123,0.20)(0.0154,0.25)(0.02,0.35)
    };
    \addlegendentry{$d=20$}

\addplot[
    color=blue,
    mark=diamond,
    ]
    coordinates {
    (0.003,0.049)    (0.00623,0.11)  (0.01,0.159)  (0.0122,0.205)(0.0153,0.2485)(0.02,0.351)
    };
    \addlegendentry{$d=30$}

    \addplot[
    color=green,
    mark=o,
    ]
    coordinates {
    (0.00311,0.052)    (0.00627,0.115)  (0.01,0.16)  (0.0122,0.207)(0.015,0.25)(0.021,0.352)
    };
    \addlegendentry{$d=50$}
\end{axis}
\end{tikzpicture}}}
\subfigure[]{\resizebox {0.45 \textwidth} {!}{
\begin{tikzpicture}
\begin{axis}[
    legend style={font=\tiny},
    xlabel={Repetitions $M$},
    ylabel={$\sum_{j=1}^k  \| \appsig{t}_j - \appsigstar_j \|^2$},
    xmin=28, xmax=112,
    ymin=0, ymax=1.21,
    legend pos= north east,
    ymajorgrids=true,
    xmajorgrids=true,
    grid style=dashed,
]
 
\addplot[
    color=blue,
    mark=square,
    ]
    coordinates {
    (30,1.2)(40,0.963)(50,0.168)(60,0.0608)(70,0.0204)(80,0.0201)(90,0.0201)(100,0.02)(110,0.0201)
    };
    \addlegendentry{\texttt{rand-AM}}

    \addplot[
    color=red,
    mark=square,
    ]
    coordinates {
    (30,0.98)(40,0.910)(50,0.138)(60,0.0601)(70,0.0201)(80,0.02)(90,0.02)(100,0.02)(110,0.02)
    };
    \addlegendentry{PCA+rs+AM}
\end{axis}
\end{tikzpicture}}}
\caption{Simulation of the PCA and overall guarantees for Gaussian covariates. We assume that the true parameter matrix $\Thetastar=A^*(U^*)^\top$ for a $\real^{d \times k}$ matrix $U^*$ and an invertible $A^* \in \real^{k \times k}$, and that Algorithm~\ref{algo:pca} returns a subspace estimate $\widehat{U}$. Panel (a) reveals the subspace estimation error as a function of $d/n$, which is corroborated by Theorem~\ref{thm:pca}. In panel (b), we compare the performance of our overall algorithm (in red) with that of AM with repeated random initialization~\cite{balazs2016convex} (in blue) averaged over $50$ trials. We fix $k=3$, $d=50$, $n=35kd$ and $\sigma = 0.1$. For a sufficiently large $M$, both schemes perform in a similar fashion.}
\label{fig:overall}
\end{figure*}

\begin{theorem} \label{thm:pca}
Suppose that the covariates $\meas_i$ are drawn i.i.d from the standard Gaussian distribution $\NORMAL(0,I_d)$, and that the true parameters $\appsigstar_1, \ldots, \appsigstar_k$ are fixed. Then there is a universal constant $C$ such that $\Uhat$ satisfies the bound
\begin{align*}
\fronorm{\Uhat \Uhat^\top - \Ustar (\Ustar)^\top}^2 \leq C \left (\frac{\sigma^2 + \varsigma^2 }{\gamma^2} \right ) \frac{k d \log^3 (nk) }{n}
\end{align*}
with probability greater than $1 - C n^{-10}$.
\end{theorem}
The proof of Theorem~\ref{thm:pca} is provided in Appendix~\ref{sec:pf-thm3}. We have thus shown that the projection matrix $U^* (U^*)^\top$ onto the true subspace spanned by the vectors $\thetastar_1, \ldots, \thetastar_k$ can be estimated
at the parametric rate via our PCA procedure. This guarantee is illustrated via simulation in panel (a) of Figure~\ref{fig:overall}. 

Let us now turn to establishing a guarantee on Algorithm~\ref{algo:rs} when it is given a (generic) subspace estimate $\Uhat$ as input.
Since the model~\eqref{mafd} is only identifiable up to a relabeling of the individual parameters, we can only hope to show that a suitably permuted set of the initial parameters is close to the true parameters. 
Toward that end, let $\mathcal{P}_k$ denote the set of all permutations from $[k] \rightarrow [k]$, and let 
\begin{align} \label{eq:distfunc}
\dist \left( \paren{\appsig{0}_j }_{j=1}^{k}, \paren{\appsigstar_j }_{j=1}^{k} \right) \defn \min_{P \in \mathcal{P}_k} \sum_{j = 1}^k \|  \appsig{0}_{P(j)} - \appsigstar_j \|^2 
\end{align}
denote the minimum distance attainable via a relabeling of the parameters.
With this notation in place, we are now ready to state our result for parameter initialization. In it, we assume that the input matrix $\Uhat$ is fixed independently of the samples used to carry out the random search.

\begin{theorem} \label{thm:rs}
Let $\delta \in (0, 1)$ and $0 \leq r \leq \frac{\Delta \pimin^{5/2}\log^{-1/2}(k/\pi_{min})}{8 k^3}$ denote two positive scalars. 
Suppose that the covariates $\meas_i$ are drawn i.i.d from the standard Gaussian distribution $\NORMAL(0,I_d)$, and that the true parameters $\appsigstar_1, \ldots, \appsigstar_k$ are fixed. 
Further suppose that 
\begin{align*}
\opnorm{ \Uhat \Uhat^\top - U^* (U^*)^\top } \leq \frac{\Delta \pimin^{5/2}\log^{-1/2}(k/\pi_{min})}{8 \Deltamax k^3}, \quad \quad \text{ and that } \quad \quad
M \geq \left( 1 + \frac{\Deltamax}{r} \right)^{k^2} \log(1 / \delta).
\end{align*}
Then there is a tuple of universal constants $(c_1, c_2)$ such that if \\
 $n \geq c_1 \max \paren{ d \frac{k}{\pimin}, \sigma^2 \frac{k^5}{\pimin^5 \Delta^2} \log(k/\pimin) \log M }$, then
\begin{align*}
& \min_{c > 0} \; \dist\left( \paren{ c \appsig{0}_j }_{j = 1}^k, \paren{ \appsigstar_j}_{j = 1}^k \right) \\
 & \leq  c_1 \log(k/\pimin) \left( \frac{k}{\pimin} \right)^5 \bigg \lbrace 4k \left( r^2 + \Deltamax^2 \opnorm{ \Uhat \Uhat^\top - U^* (U^*)^\top }^2 \right)  + \frac{\sigma^2 \log M }{n} \bigg \rbrace
\end{align*}
with probability exceeding $1 - \delta - 2 e^{-c_2 n}$.
\end{theorem}

We prove Theorem~\ref{thm:rs} in Appendix~\ref{sec:pf-thm4}. Combining Theorems~\ref{thm:pca} and~\ref{thm:rs} with some algebra then allows us to prove a guarantee for the initialization procedure that combines Algorithms~\ref{algo:pca} and~\ref{algo:rs} in sequence. In particular, fix a pair of positive scalars $\epsilon \leq \Delta$ and $\delta \in (0, 1)$. Then under the Gaussian covariate assumption, and provided the true parameters are fixed, combining the theorems shows that if (for an appropriately large universal constant $c$), we have~\mbox{$M \geq \left( 1 + c \frac{\Deltamax k^3 \log^{1/2}(k/\pimin)}{\epsilon \pimin^{5/2}} \right)^{k^2} \log(1 / \delta)$}, and the sample size $n$ is greater than
\begin{align} \label{eq:ninit}
n_{\mathsf{init}}(\epsilon, M, c) &\defn c \max \paren{ d \frac{k}{\pimin}, \sigma^2 \frac{k^5}{\pimin^5 \epsilon^2} \log(k/\pimin) \log(M / \delta), d \log^3(nk) \log(k/\pimin) \frac{k^7 \Deltamax^2}{\gamma^2 \pimin^5 \epsilon^2} (\sigma^2 + \varsigma^2 ) },
\end{align}
then $
\min_{c > 0} \; \dist\left( \paren{ c \appsig{0}_j }_{j = 1}^k, \paren{ \appsigstar_j}_{j = 1}^k \right) \leq \epsilon^2 $ with probability greater than $1 - \delta - c n^{-10}$.
Equipped with this guarantee on our initialization step, we are now in a position to state an end-to-end guarantee on our overall methodology in the next section.

\subsection{Overall algorithmic guarantee} \label{sec:overall}

Assume without loss of generality that the identity permutation minimizes the distance measure $\dist$, so that $\appsig{0}_j$ is the estimate of the parameter $\appsigstar_j$ for each $j \in [k]$. 
%
%
Recall the statistical error $\delta_{n, \sigma}(d, k, \pi_{\min})$ defined in equation~\eqref{eq:stat-err}, which is, up to a constant factor, the final (squared) radius of the ball to which the AM update converges when initialized suitably, and the notation $n_{\mathsf{AM}}(c)$ and $n_{\mathsf{init}}(\epsilon, M, c)$ from equations~\eqref{eq:nam} and~\eqref{eq:ninit}, respectively. We now state a guarantee for our overall procedure that runs Algorithms~\ref{algo:pca},~\ref{algo:rs}, and~\ref{algo:AM} in that sequence; we omit the proof since it follows by simply putting together the pieces from Theorem~\ref{thm:mt} and the discussion above.

\begin{corollary} \label{cor:overall}
Suppose that the covariates $\meas_i$ are drawn i.i.d. from a standard Gaussian distribution $\NORMAL(0, I_d)$, and that the unknown parameters $\appsigstar_1, \ldots, \appsigstar_k$ are fixed. Then there exist universal constants $c_1$ and $c_2$ such that for each $\delta \in (0, 1)$, if
\begin{align*}
M &\geq \left( 1 + c_1 \frac{\Deltamax k^4 \log^{1/2}(k/\pimin)}{\pimin^{11/2}} \right)^{k^2} \log(1 / \delta), \qquad \numobs \geq \max \paren{n_{\mathsf{init}} \left( c_{2} \frac{\pimin^3}{k}, c_1, M \right), n_{\mathsf{AM}} (c_1) }  \\
&\qquad \qquad \qquad \text{ and } \qquad 
T_0 = c_1 \log \left( \frac{1}{\delta_{n, \sigma}(d, k, \pi_{\min})} \right),
\end{align*}
then the combined algorithm satisfies, simultaneously for all $T \geq T_0$, the bound
\begin{align*}
\Pr \paren{ \sum_{j = 1}^k \| \appsig{T}_j - \appsigstar_j \|^2 \geq c_1 \delta_{n, \sigma}(d, k, \pi_{\min}) } \leq \delta + c_1 \left( n^{-10} + k \exp \left( - c_2 \numobs \frac{\pi_{\min}^4}{\log^2 (1 / \pi_{\min})} \right) + \frac{k^2}{n^7}\right).
\end{align*}
\end{corollary}

We thus obtain, for Gaussian covariates, an algorithm that when given a number of samples that is near-linear in the ambient dimension, achieves the rate $\delta_{n, \sigma}(d, k, \pi_{\min}) = \frac{\sigma^2 k d}{ \pi_{\min}^3 n} \log (kd) \log (n / kd)$
of estimation of all $kd$ parameters in squared $\ell_2$ norm. This convergence is illustrated in simulation in Figure~\ref{fig:overall}, in which we choose $k=3$, $d=50$ and $n=35kd$. Interestingly, panel (b) of this figure shows that our provable multi-step algorithm has performance similar to the algorithm that runs AM with repeated random initializations.

The computational complexity of our overall algorithm (with exact matrix inversions) is given by $\order \left( k n d^2 \log \left( \frac{1}{\delta_{n, \sigma}(d, k, \pi_{\min}} \right) + M nd \right)$, where we also assume that the $k$ top eigenvectors of the matrix $\Mhat$ are computed exactly in Algorithm~\ref{algo:pca}. This guarantee can also be extended to the case where the linear system is solved up to some numerical precision by (say) a conjugate gradient method and the eigenvectors of $\Mhat$ are computed using the power method, thereby reducing the computational complexity. Such an extension is standard and we do not detail it here.


\subsection{Proof ideas and technical challenges} \label{sec:pfideas}

Let us first sketch, at a high level, the ideas required to establish guarantees on the AM algorithm. We need to control the iterates of the AM algorithm without sample-splitting across iterations, and so the iterates themselves are random and depend on the sequence of random variables $(\appmeas_i, \epsilon_i)_{i = 1}^n$. A popular and recent approach to handling this issue in related iterative algorithms (e.g.,~\cite{balakrishnan2017statistical}) goes through two steps: first, the population update, corresponding to running~\eqref{eq:sets}-\eqref{eq:iteration} in the case $n \to \infty$, is analyzed, after which the random iterates in the finite-sample case are shown to be close to their (non-random) population counterparts by using concentration bounds for the associated empirical process. The main challenge in our setting is that the population update is quite non-trivial to write down even for Gaussian covariates, since it involves a delicate understanding of the geometry of the covariate distribution induced by the maxima of affine functions. 
We thus resort to handling the random iterates directly, thereby sidestepping the calculation of the population operator entirely.

Broadly speaking, we analyze the update~\eqref{eq:sets}-\eqref{eq:iteration} by relating the error of the parameters generated by this update to the error of the parameters from which the update is run. This involves three distinct technical steps; these steps are spiritually similar for the proofs of both Theorems~\ref{thm:mt} and~\ref{thm:sbgen}. The first step (handled by Lemma~\ref{lem:noise}) is to control the behavior of the noise in the problem. In order to do so, we apply standard concentration bounds for quadratic forms of sub-Gaussian random variables, in conjunction with bounds on the \emph{growth functions} of multi-class classifiers~\cite{daniely2012multiclass}. Crucially, this affords a uniform bound on the noise irrespective of which iterate the alternating minimization update is run from, and handles \emph{any} covariate matrix. The second step corresponds (roughly) to 
controlling the prediction error in the noiseless problem, for which we show quantitative results (in Lemmas~\ref{lem:lin-affine} and~\ref{lem:lin-affine-c}) that strictly generalize a result of Waldspurger~\cite{walds}. This step crucially uses the small-ball condition satisfied by the covariate distribution (corresponding, in Theorem~\ref{thm:mt}, to bounds on the lower tail of non-central $\chi^2$ variates.)
Finally, in order to translate a prediction error guarantee into a guarantee on the estimation error, we invert specifically chosen sub-matrices of the covariate matrix $\AppXmat$ over the course of the algorithm, and our bounds naturally depend on how these sub-matrices are conditioned. A key technical difficulty of the proof is therefore to control the spectrum of these random matrices, rows of which are drawn from (randomly) truncated variants of the covariate distribution. 
Our techniques for controlling the spectrum differ in the proofs of the two theorems. The first technique is specific to the Gaussian covariate distribution: the expectation of such a random matrix can be characterized by appealing to tail bounds on the non-central $\chi^2$ distribution, and the Gaussian covariate assumption additionally allows us to show that an analogous result holds for the random matrix with high probability (see Lemma~\ref{lem:sing}).
The second technique---applied in Theorem~\ref{thm:sbgen}---is more broadly applicable, and provides a simultaneous lower bound on the minimum singular value of \emph{all} sub-matrices of $\Xi$ of a certain size, provided that the distribution satisfies a small-ball condition. We expect this result (stated and proved in Lemma~\ref{lem:minsingsb}) to be of broader interest.
Here, our initialization condition is crucial: the aforementioned singular value control suffices for the sub-matrices formed by the \emph{true} parameters, and we translate these bounds to the sub-matrices generated by random parameters by appealing to the fact that the initialization is sufficiently close to the truth.

Let us now turn to discussing the techniques used to prove Theorems~\ref{thm:pca} and~\ref{thm:rs}. As mentioned before, our proof of Theorem~\ref{thm:pca} relies on a lower bound on the eigengap of the population moment. Since we operate exclusively with Gaussian covariates, we obtain such a lower bound by appealing to classical moment calculations for suitably truncated Gaussian distributions~\cite{tallis1961moment}. Translating these calculations into an eigengap is quite technical, and involves the isolation of many properties of the population moments that may be of independent interest. As briefly alluded to in Section~\ref{sec:setup}, the heart of the technical difficulty is due to the fact that that $\max$ function is not differentiable, and so moments cannot be calculated by repeated applications of Stein's lemma like in related problems~\cite{babichev2018slice, mixture_many, zhong2017recovery}.

In order to establish Theorem~\ref{thm:rs}, we crucially use the scale-invariance property of the initialization along with some arguments involving empirical process theory to show that the goodness-of-fit statistic employed in the algorithm is able to isolate a good initialization. Establishing these bounds requires us to relate the prediction and estimation errors in the problem (in Lemma~\ref{lem:pred_est_gen}), which may be of independent interest.

\section{Discussion} \label{sec:discussion}

We conclude this portion of the paper with short discussions of related models 
and future directions.

\subsection{Related models} \label{sec:related-models}

Models closely related to \eqref{mafd} also appear in second price
auctions, where an item having $d$ features is bid on and sold to the
highest bidder at the second highest bid~\cite{medina2014learning,
  morgenstern2016learning}. Assuming that each of $k$ user groups bids
on an item and that each bid is a linear function of the features, one
can use a variant of the model \eqref{mafd} with the $\max$ function
replaced by the second order statistic to estimate the individual bids
of the user groups based on historical data. Another related problem
is that of multi-class classification~\cite{daniely2012multiclass}, in
which one of $k$ labels is assigned to each sample based on the
$\argmax$ function, i.e., for a class of functions $\mathcal{F}$, we
have the model $Y = \argmax_{1 \leq j \leq k} f_j (X)$ for $j$
distinct functions $f_1, \ldots, f_k \in \mathcal{F}$. When $\mathcal{F}$ is
the class of linear functions based on $d$ features, this can be
viewed as the ``classification" variant of our regression problem.

The model \eqref{mafd} can also be seen as a special case of
multi-index models \cite{li1991sliced, horowitz2009semiparametric} as
well as mixture-of-experts models \cite{jacobs1991adaptive,
  yuksel2012twenty}. Multi-index models are of the form $Y =
g( \inprod{\sigstar_1}{X}, \ldots, \inprod{\sigstar_k}{X}) + \epsilon$
for an \textit{unknown} function $g$ and this function $g$ is taken to
be the $\max(\cdot)$ function in the model~\eqref{mafd}. In the
mixture-of-experts model, the covariate space is partitioned into $k$
regions via certain  \emph{gating functions}, and the observation
model is given by $k$ distinct regression functions: one on each
region. The model~\eqref{mafd} is clearly a member of this class,
since the $\max(\cdot)$ function implicitly defines a partition of
$\real^d$ depending on which of the $k$ linear functions of $X$
attains the maximum, and on each of these partitions, the regression
function is linear in $X$.

\subsection{Future directions} \label{sec:future}

In this paper, we analyzed a natural alternating minimization algorithm for estimating the maximum of unknown affine functions, and established that it enjoys local linear convergence to a ball around the optimal parameters. We also proposed an initialization based on PCA followed by random search in a lower-dimensional space. The random search step is just one way to mimic the least squares procedure in low dimensions; the latter is computationally inefficient, but we analyze it for completeness in Appendix~\ref{app:least-squares-init} to follow. An interesting open question is if there are other efficient methods besides random search that work just as well post dimensionality reduction. Another interesting question has to do with the necessity of dimensionality reduction: in simulations (see, e.g., Figure~\ref{fig:overall}), we have observed that if the AM algorithm is repeatedly initialized in $(d+1)$-dimensional space without dimensionality reduction, then the number of repetitions required to obtain an initialization from which it succeeds (with high probability) is similar to the number of repetitions required after dimensionality reduction. This suggests that our (sufficient) initialization conditions~\eqref{eq:init} and~\eqref{eq:init2gen} may be too stringent, and that the necessary conditions on the initialization to ensure convergence of the AM algorithm are actually much weaker. We leave such a characterization for future work, but note that some such conditions must exist: the AM algorithm when run from a single random initialization, for instance, fails with constant probability when $k \geq 3$. 
%
Understanding the behavior of the randomly initialized AM algorithm is also an open problem in the context of phase retrieval~\cite{walds, zhang2018phase}.

In the broader context of max-affine estimation, it is also interesting to analyze other non-convex procedures (e.g. gradient descent) to obtain conditions under which they obtain accurate parameter estimates. The CAP estimator of Hannah and Dunson~\cite{hannah2013multivariate} and the adaptive max-affine
partitioning algorithm of Bal{\'a}zs~\cite{balazs2016convex} are also interesting procedures for estimation under these models, and it would be interesting to analyze their performance when the number of affine pieces $k$ is fixed and known.
For applications in which the dimension $d$ is very large, it is also interesting to study the model with additional restrictions of sparsity on the unknown parameters---such problems are known to exhibit interesting statistical-computational gaps even in the special case of sparse phase retrieval (see, e.g., Cai et al.~\cite{cai2016optimal}).

\subsection*{Acknowledgements}

Avishek Ghosh and Kannan Ramchandran were supported in part by NSF grant  
NSF CCF-1527767.
Ashwin Pananjady was supported in part by Office of Naval Research Grant
ONR-N00014-18-1-2640 and National Science Foundation Grants
NSF DMS-1612948 and NSF CCF-1704967. Adityanand Guntuboyina was supported in part by NSF CAREER Grant DMS-16-54589.
We thank Bodhi Sen for helpful discussions.

\bibliographystyle{alpha}
\bibliography{max-aff-annals}

\appendix

\vspace{5mm}

\begin{center}
{\bf{\LARGE{Appendix}}}
\end{center}

\vspace{5mm}

We now present proofs of our main results. We assume throughout that the sample size 
$\numobs$ is larger than some universal constant; in the complementary case, the constant factors in our bounds can be modified appropriately. Values of constants $c, c_1, c', \ldots$ may change from line to line. Statements of our theorems, for instance, minimize the number of constants by typically using one of these to denote a large enough constant, and another to denote a small enough constant.


\section{Proof of Theorem~\ref{thm:mt}} \label{sec:pf-thm1}


Let us begin by introducing some shorthand notation, and providing a formal statement of the probability bound guaranteed by the theorem.
For a scalar $\wstar$, vectors $\ustar \in \real^{d}$ and $\vstar = (\ustar, \; \wstar) \in \real^{d + 1}$, and a positive scalar $r$, let
$
\mathcal{B}_{\vstar} (r) = \paren{ v \in \real^{d+1}: \frac{\eucnorm{ v - \vstar } }{\eucnorm{\ustar} } \leq r },
$
and let 
\[
\Ispace \left(r; \paren{ \appsigstar_j }_{j = 1}^k \right) = \paren{ \beta_1, \ldots \beta_k \in \real^{d+1}: \exists c>0: c (\beta_i - \beta_j) \in \mathcal{B}_{ \appsigstar_i - \appsigstar_j } ( r ) \text{ for all } 1 \leq i \neq j \leq k }.
\]
Also, use the shorthand
\begin{align*}
\vartheta_t \left(r; \paren{ \appsigstar_j }_{j = 1}^k \right) \defn \; \sup_{ \appsig{0}_1, \ldots, \appsig{0}_k \in \Ispace(r)  } \; &\sum_{j = 1}^k \| \appsig{t}_j - \appsigstar_j \|^2 - \left ( \frac{3}{4} \right )^t \left( \sum_{j = 1}^k \| c^* \appsig{0}_j - \appsigstar_j \|^2 \right), \text{ and } \\
\delta^{\NORMAL}_{n, \sigma}(d, k, \pi_{\min}) &\defn \sigma^2 \frac{kd}{\pi^3_{\min} n} \log(kd) \log(n/kd)
\end{align*}
to denote the error tracked over iterations (with $c^*$ denoting the smallest $c > 0$ such that $c (\beta_i - \beta_j) \in \mathcal{B}_{ \appsigstar_i - \appsigstar_j } ( r ) \text{ for all } 1 \leq i \neq j \leq k$), and a proxy for the final statistical rate, respectively.

Theorem~\ref{thm:mt} states that provided the true parameters $\paren{ \appsigstar_j }_{j = 1}^k$ are fixed, there are universal constants $c_1$ and $c_2$ such that if the sample size obeys the condition $n \geq n_{\mathsf{AM}}(c_1)$, 
then we have
\begin{align} \label{eq:aditya-bound1}
\Pr \paren{ \max_{t \geq 1} \; \vartheta_t \left( c_2 \frac{\pimin^3}{k \kappa} ; \paren{ \appsigstar_j }_{j = 1}^k \right) \geq c_1 \delta^{\NORMAL}_{n, \sigma}(d, k, \pi_{\min}) } \leq c_2 \left(k \exp \left( - c_1 \numobs \frac{\pi_{\min}^4}{\log^2 (1 / \pi_{\min})} \right) + \frac{k^2}{n^7}\right).
\end{align}

Let us now proceed to a proof of the theorem, assuming without loss of generality that the scalar $c^*$ above is equal to $1$.
It is convenient to state and prove another result that guarantees a one-step contraction, from which Theorem~\ref{thm:mt} follows as a corollary. In order to state this result, we assume that one step of the alternating minimization update~\eqref{eq:sets}-\eqref{eq:iteration} is run starting from the parameters $\paren{\beta_j}_{j = 1}^k$ to produce the next iterate $\paren{\beta^+_j}_{j = 1}^k$. In the statement of the proposition, we use the shorthand
\begin{align*}
\vstar_{i, j} &= \appsigstar_{i} - \appsigstar_{j}, \\
v_{i, j} &= \beta_i - \beta_j, \text{ and } \\ 
v^+_{i, j} &= \beta^+_i - \beta^+_j.
\end{align*}
Also recall the definitions of the geometric quantities $(\Delta, \kappa)$. The following proposition guarantees the one step contraction bound. 

\begin{proposition} \label{thm:onestep}
Suppose that the covariates are drawn i.i.d. from the standard Gaussian distribution, and that the true parameters $\{ \appsigstar_j \}_{j = 1}^k$ are fixed. Then there exist universal constants $c_1$ and $c_2$ such that

(a) If the sample size satisfies the bound $n \geq c_1 \max \paren{ d, 10 \log n } \max \paren{ \frac{k}{\pi_{\min}^3}, \frac{\log^2 (1 / \pi_{\min}) }{\pi_{\min}^3}, \log (n / d) }$, then for all parameters $\paren{ \beta_j }_{j = 1}^k$ satisfying 
\begin{subequations}
\begin{align}
\max_{1 \leq j \neq j' \leq k} \; \frac{\left\| v_{j, j'} - \vstar_{j, j'} \right\|}{\| \sigstar_j -\sigstar_{j'} \| } \log^{3/2} \left( \frac{\| \sigstar_j -\sigstar_{j'} \| }{\left\| v_{j, j'} - \vstar_{j, j'} \right\|} \right) &\leq c_2 \frac{\pi_{\min}^3}{k \kappa},\label{eq:init-thm1-i}
\end{align}
we have, simultaneously for all pairs $1 \leq j \neq \ell \leq k$, the bound
\begin{align}
\frac{\left\| v^+_{j, \ell} - \vstar_{j, \ell} \right\|^2}{\| \sigstar_j -\sigstar_{\ell} \|^2} &\leq \max \paren{ \frac{d \kappa}{\pi_{\min}^3 n}, \frac{1}{4k} } \left( \sum_{j' = 1}^k \frac{ \eucnorm{ v_{j, j'} - \vstar_{j, j'} }^2}{ \| \sigstar_j -\sigstar_{j'} \|^2 } + \frac{ \eucnorm{ v_{\ell, j'} - \vstar_{\ell, j'} }^2}{ \| \sigstar_\ell -\sigstar_{j'} \|^2 } \right) + c_1 \frac{\sigma^2}{\Delta} \frac{kd}{\pi^3_{\min} n} \log(n/d) \label{eq:1step-thm1-i}
\end{align}
with probability exceeding $ 1 - c_1 \left(k \exp \left( - c_2 \numobs \frac{\pi_{\min}^4}{\log^2 (1 / \pi_{\min})} \right) + \frac{k^2}{n^7}\right)$.
\end{subequations}

(b) If the sample size satisfies the bound $n \geq c_1 \max\paren{ \max \paren{ d, 10 \log n } \max \paren{ \frac{k}{\pi_{\min}^3}, \frac{\log^2 (1 / \pi_{\min}) }{\pi_{\min}^3}, \log (n / d) }, \frac{kd}{\pi_{\min}^3}}$, then for all parameters $\paren{ \beta_j }_{j = 1}^k$ satisfying 
\begin{subequations}
\begin{align}
\max_{1 \leq j \neq j' \leq k} \; \frac{\left\| v_{j, j'} - \vstar_{j, j'} \right\|}{\| \sigstar_j -\sigstar_{j'} \| } \log^{3/2} \left( \frac{\| \sigstar_j -\sigstar_{j'} \| }{\left\| v_{j, j'} - \vstar_{j, j'} \right\|} \right) &\leq c_2 \frac{\pi_{\min}^3}{k}, \label{eq:init-thm1-ii}
\end{align}
we have the overall estimation error bound
\begin{align}
\sum_{i = 1}^k \| \beta^+_j - \appsigstar_j \|^2 &\leq \frac{3}{4} \cdot \left( \sum_{i = 1}^k \| \beta_j - \appsigstar_j \|^2 \right) + c_1 \sigma^2 \frac{kd}{\pi^3_{\min} n} \log(k) \log(n/dk) \label{eq:1step-thm1-ii}
\end{align}
\end{subequations}
with probability exceeding $ 1 - c_1 \left(k \exp \left( - c_2 \numobs \frac{\pi_{\min}^4}{\log^2 (1 / \pi_{\min})} \right) + \frac{k^2}{n^7}\right)$.
\end{proposition}

Let us briefly comment on why Proposition~\ref{thm:onestep} implies Theorem~\ref{thm:mt} as a corollary. Clearly, equations~\eqref{eq:init-thm1-ii} and~\eqref{eq:1step-thm1-ii} in conjunction show that the estimation error decays geometrically after running one step of the algorithm. The only remaining detail to be verified is that the next iterates $\paren{ \beta^+_j }_{j = 1}^k$ also satisfy condition~\eqref{eq:init-thm1-i} provided the sample size is large enough; in that case, the one step estimation bound~\eqref{eq:1step-thm1-ii} can be applied recursively to obtain the final bound~\eqref{eq:thm1}.

For the constant $c_2$ in the proposition, let $r_b$ be the largest value in the interval $[0, e^{-3/2}]$ such that $r_b \log^{3/2} (1/ r_b) \leq c_2 \frac{\pi_{\min}^3}{k}$. Similarly, let $r_a$ be the largest value in the interval $[0, e^{3/2}]$ such that $r_a \log^{3/2} (1/ r_a) \leq c_2 \frac{\pi_{\min}^3}{k \kappa}$.

Assume that the current parameters satisfy the bound~\eqref{eq:init-thm1-i}. Choosing $n \geq 4 \kappa d / \pi_{\min}^3$ and applying inequality~\eqref{eq:1step-thm1-i}, we have, for each pair $1 \leq j \neq \ell \leq k$, the bound
\begin{align*}
\frac{\left\| v^+_{j, \ell} - \vstar_{j, \ell} \right\|^2}{\| \sigstar_j -\sigstar_{\ell} \|^2} &\leq \frac{1}{4k} \left( \sum_{j' = 1}^k \frac{ \eucnorm{ v_{j, j'} - \vstar_{j, j'} }^2}{ \| \sigstar_j -\sigstar_{j'} \|^2 } + \frac{ \eucnorm{ v_{\ell, j'} - \vstar_{\ell, j'} }^2}{ \| \sigstar_\ell -\sigstar_{j'} \|^2 } \right) + c_1 \frac{1}{\| \sigstar_j -\sigstar_{\ell} \|^2} \sigma^2 \frac{kd}{\pi^3_{\min} n} \log(n/d) \\
&\leq \frac{1}{2} r_a^2 + c_1 \frac{\sigma^2}{\Delta} \frac{kd}{\pi^3_{\min} n} \log(n/d).
\end{align*}
Further, if $n \geq C \sigma^2 \frac{k^3 \kappa^2 d}{\pi_{\min}^9 \Delta r_0^2} \log (k \kappa /\pimin^3) \log(n/d)$ for a sufficiently large constant $C$, we have
\begin{align*}
\frac{\left\| v^+_{j, \ell} - \vstar_{j, \ell} \right\|^2}{\| \sigstar_j -\sigstar_{\ell} \|^2} &\leq r_a^2.
\end{align*}
Thus, the parameters $\paren{\beta_j^+ }_{j = 1}^k$ satisfy inequality~\eqref{eq:init-thm1-i} for the sample size choice required by Theorem~\ref{thm:mt}. 
Finally, noting, for a pair of small enough scalars $(a, b)$, the implication
\begin{align*}
a \leq \frac{b}{2} \log^{-3/2} (1/b) \implies a \log^{3/2} (1/a) \leq b,
\end{align*}
and adjusting the constants appropriately to simplify the probability statement completes the proof of the theorem. 

\subsection{Proof of Proposition~\ref{thm:onestep}}

We use the shorthand notation
$
S_j \defn S_j(\beta_1,\dots, \beta_k),
$
and let $P_{S_j}$ denote the projection matrix onto the range of the matrix $\AppXmat_{S_j}$.
Recall our notation for the difference vectors.

Let $y^*$ denote the vector with entry $i$ given by $\max_{\ell \in [k]} \; \inprod{\appmeas_i}{\appsigstar_{\ell}}$. We have
\begin{align}
\| \AppXmat_{S_j} (\beta^+_j - \beta^*_j ) \|^2 &= \| P_{S_j} y_{S_j} - \AppXmat_{S_j} \appsigstar_j \|^2 \notag \\
&= \| P_{S_j} y^*_{S_j} + P_{S_j} \epsilon_{S_j} - \AppXmat_{S_j} \appsigstar_j \|^2 \notag \\
&\leq 2 \| P_{S_j} ( y^*_{S_j} - \AppXmat_{S_j} \appsigstar_j) \|^2 + 2 \| P_{S_j} \epsilon_{S_j} \|^2 \notag \\
&\leq 2 \| y^*_{S_j} - \AppXmat_{S_j} \appsigstar_j \|^2  + 2 \| P_{S_j} \epsilon_{S_j} \|^2, \label{eq:key}
\end{align}
where we have used the fact that the projection operator is non-expansive on a convex set.

Let
\[
\paren{ \inprod{\appmeas_i}{\beta_{\ell}} = \max } \defn \paren{ \inprod{\appmeas_i}{\beta_{\ell}} = \max_{ u \in [k] } \inprod{\appmeas_i}{\beta_{u}} }, \text{ for each } i \in [n], \ell \in [k]
\]
denote a convenient shorthand for these events.
The first term on the RHS of inequality~\eqref{eq:key} can be written as
\begin{align*}
\sum_{i \in S_j} ( y^*_i - \inprod{\appmeas_i}{\appsigstar_j})^2 &\leq \sum_{i = 1}^n \sum_{j': j' \neq j } \ind{ \inprod{\appmeas_i}{\beta_{j}} = \max \text{ and } \inprod{\appmeas_i}{\appsigstar_{j'}} = \max  } \inprod{\appmeas_i}{\appsigstar_{j'} - \appsigstar_{j}}^2,
\end{align*}
where the inequality accounts for ties. Each indicator random variable is bounded, in turn, as
\begin{align*}
\ind{ \inprod{\appmeas_i}{\beta_{j}} = \max \text{ and } \inprod{\appmeas_i}{\appsigstar_{j'}} = \max  } &\leq \ind{ \inprod{\appmeas_i}{\beta_{j}} \geq \inprod{\appmeas_i}{\beta_{j'}} \text{ and } \inprod{\appmeas_i}{\appsigstar_{j'}} \geq \inprod{\appmeas_i}{\appsigstar_{j}}  } \\
&= \ind{ \inprod{\appmeas_i}{v_{j, j'}} \cdot \inprod{\appmeas_i}{\vstar_{j, j'}} \leq 0 }.
\end{align*}
Switching the order of summation yields the bound
\begin{align*}
\sum_{i \in S_j} ( y^*_i - \inprod{\appmeas_i}{\appsigstar_j})^2 &\leq \sum_{j': j' \neq j} \sum_{i = 1}^n  \ind{ \inprod{\appmeas_i}{v_{j, j'}} \cdot \inprod{\appmeas_i}{\vstar_{j, j'}} \leq 0 } \inprod{\appmeas_i}{\vstar_{j, j'}}^2.
\end{align*}

Recalling our notation for the minimum eigenvalue of a symmetric matrix, the LHS of inequality~\eqref{eq:key} can be bounded as
\begin{align*}
\| \AppXmat_{S_j} (\beta^+_j - \appsigstar_j ) \|^2 &\geq \lambda_{\min}\left( \AppXmat^\top_{S_j} \AppXmat_{S_j} \right) \cdot \| \beta^+_j - \appsigstar_j \|^2.
\end{align*}

Putting together the pieces yields, for each $j \in [k]$, the pointwise bound
\begin{align}
\frac{1}{2} \lambda_{\min} \left( \AppXmat^\top_{S_j} \AppXmat_{S_j} \right) \cdot \| \beta^+_j - \appsigstar_j \|^2 \leq \sum_{j':j' \neq j} \sum_{i = 1}^n  \ind{ \inprod{\appmeas_i}{v_{j, j'}} \cdot \inprod{\appmeas_i}{\vstar_{j, j'}} \leq 0 } \inprod{\appmeas_i}{\vstar_{j, j'}}^2 + \| P_{S_j} \epsilon_{S_j} \|^2. \label{eq:key3}
\end{align}
Up to this point, note that all steps of the proof were deterministic. In order to complete the proof, it suffices to show high probability bounds on the various quantities appearing in the bound~\eqref{eq:key3}. Since the set $S_j$ is in itself random and could depend on the pair $(\Xi, \epsilon)$, bounding individual terms is especially challenging. Our approach is to show bounds that hold uniformly over all parameters $\paren{ \beta_j }_{j = 1}^k$ that are close to the true parameters. 

Recall the notation
\begin{align*}
\mathcal{B}_{\vstar} (r) = \paren{ v \in \real^{d+1}: \frac{\eucnorm{ v - \vstar } }{\eucnorm{\ustar} } \leq r }
\end{align*}
introduced before, and the definitions of the pair of scalars $(r_a, r_b)$.

To be agnostic to the scale invariance of the problem, we set $c^*= 1$ and  define the set of parameters
\begin{align*}
\Ispace(r) = \paren{ \beta_1, \ldots, \beta_k: v_{i, j} \in \mathcal{B}_{\vstar_{i, j}} (r) \text{ for all } 1 \leq i \neq j \leq k },
\end{align*}
and use the shorthand $\Ispace_a \defn \Ispace(r_a)$ and $\Ispace_b \defn \Ispace(r_b)$, to denote the set of parameters satisfying conditions~\eqref{eq:init-thm1-i} and~\eqref{eq:init-thm1-ii}, respectively,

Recall that we denote by
\begin{equation*}
  S_j(\beta_1,\dots, \beta_k) := \left\{1 \leq i \leq n : \inprod{\appmeas_i}{\beta_j} = \max_{1 \leq u \leq k} \left(\inprod{\appmeas_i}{\beta_u} \right) \right\},
\end{equation*}
the indices of the rows for which $\beta_j$ attains the maximum, and we additionally keep this sets disjoint by breaking ties lexicographically. To lighten notation, we use the shorthand
\begin{align*}
\Xi^j(\beta_1,\dots, \beta_k) \defn \Xi_{S_j(\beta_1,\dots, \beta_k) }.
\end{align*}

Having defined this notation, we are now ready to return to the proof of Proposition~\ref{thm:onestep}. We make the following claims to handle the three terms in the bound~\eqref{eq:key3}. First, we claim that the noise terms are uniformly bounded as
\begin{subequations}
\begin{subequations}
\renewcommand{\theequation}{\theparentequation.\Roman{equation}}
\begin{align}
&\Pr \left\{ \sup_{\beta_1,\dots, \beta_k \in \real^{d + 1} } \; \sum_{j = 1}^k \| P_{\Xi^j(\beta_1,\dots, \beta_k)} \epsilon_{S_j(\beta_1,\dots, \beta_k)} \|^2 \geq 2 \sigma^2 k (d+1) \log (kd) \log (n/kd) \right\} \leq \binom{n}{kd}^{-1}, \text{ and } \label{clm:one-i} \\
&\Pr \left\{ \sup_{\beta_1,\dots, \beta_k \in \real^{d + 1} } \; \| P_{\Xi^j(\beta_1,\dots, \beta_k)} \epsilon_{S_j(\beta_1,\dots, \beta_k)} \|^2 \geq 2 \sigma^2 k (d+1) \log (n/d) \right\} \leq \binom{n}{d}^{-1} \text{ for each } j \in [k]. \label{clm:one-ii}
\end{align}
\end{subequations}
Second, we show that the indicator quantities are simultaneously bounded for all $j, j'$ pairs. In particular, we claim that there exists a tuple of universal constants $(C, c_1, c_2, c')$ such that for each positive scalar $r \leq 1/24$, we have
\begin{align}
&\Pr \Big\{ \exists 1 \leq j \neq j' \leq k, \; v_{j, j'} \in \Bspace_{\vstar_{j, j'}} (r): \sum_{j': j' \neq j} \sum_{i = 1}^n  \ind{ \inprod{\appmeas_i}{v_{j, j'}} \cdot \inprod{\appmeas_i}{\vstar_{j, j'}} \leq 0 } \inprod{\appmeas_i}{\vstar_{j, j'}}^2 \notag \\
&\qquad \qquad \qquad \qquad \geq C \max\{ \numdim, nr \log^{3/2} (1/r) \}  \sum_{j': j' \neq j} \| v_{j, j'} - \vstar_{j, j'} \|^2 \Big\} 
\leq c_1 \binom{k}{2} \paren{ ne^{-c_2n} + e^{-c'\max \paren{ d, 10 \log n } }}. \label{clm:two}
\end{align}
Finally, we show a bound on the LHS of the bound~\eqref{eq:key3} by handling the singular values of (random) sub-matrices of $\AppXmat$ with a uniform bound. In particular, we claim that there are universal constants $(C, c, c')$ such that if $n \geq C d \max \paren{ \frac{k}{\pi_{\min}^3}, \frac{\log^2 (1 / \pi_{\min}) }{\pi_{\min}^3}, \log (n / d) }$, then for each $j \in [k]$, we have
\begin{align}
&\Pr \left\{\inf_{\beta_1, \ldots, \beta_k \in \Ispace_b} \; \; \lambda_{\min} \left( \Xi^j(\beta_1,\dots, \beta_k)^\top \cdot \Xi^j(\beta_1,\dots, \beta_k) \right) \leq C \pimin^3 n \right\} \notag \\
&\qquad \qquad \qquad \leq c \exp \left( -c \numobs \frac{\pi_{\min}^4}{\log^2 (1 / \pi_{\min})} \right) + c' \exp( - c' n \cdot \pi_{\min}). \label{clm:three}
\end{align}
\end{subequations}
\setcounter{equation}{31}

Notice that claim~\eqref{clm:one-i} implicitly defines a high probability event $\Espace^{(a.I)}$, claim~\eqref{clm:one-ii} defines high probability events $\Espace_j^{(a.II)}$, claim~\eqref{clm:two} defines a high probability event $\Espace^{(b)}(r)$, and claim~\eqref{clm:three} defines high probability events $\Espace^{(c)}_j$. Define the intersection of these events as 
\begin{align*}
\Espace(r) \defn \Espace^{(a.I)} \bigcap \left( \bigcap_{j \in [k]} \Espace^{(a.II)}_j \right) \bigcap \Espace^{(b)}(r) \bigcap \left( \bigcap_{j \in [k]} \Espace^{(c)}_j \right),
\end{align*}
and note that the claims in conjunction with the union bound guarantee that if the condition on the sample size~\mbox{$n \geq c_1 d \max \paren{ \frac{k}{\pi_{\min}^3}, \frac{\log^2 (1 / \pi_{\min}) }{\pi_{\min}^3}, \log (n / d) }$} holds, then for all $r \leq r_b$, we have
\begin{align*}
\Pr \paren{ \Espace(r) } \geq 1 - c_1 \left(k \exp \left( - c_2 \numobs \frac{\pi_{\min}^4}{\log^2 (1 / \pi_{\min})} \right) + \frac{k^2}{n^7}\right),
\end{align*}
where we have adjusted constants appropriately in stating the bound.
We are now ready to prove the two parts of the proposition.

\paragraph{Proof of part (a):} Work on the event $\Espace(r_a)$.
Normalizing inequality~\eqref{eq:key3} by $n$ and using claims~\eqref{clm:one-ii}.~\eqref{clm:two}, and~\eqref{clm:three} with $r = r_a$ then yields, simultaneously for all $j \in [k]$, the bound
\begin{align*}
\| \beta^+_j - \appsigstar_j \|^2 &\leq C \max\paren{ \frac{d}{\pi_{\min}^3 n}, \frac{r_a}{\pi_{\min}^3} \log^{3/2} (1/r_a) } \sum_{j': j' \neq j} \| v_{j, j'} - \vstar_{j, j'} \|^2 + C' \sigma^2 \frac{kd}{\pi_{\min}^3 n} \log(n/d) \\
&\stackrel{\1}{\leq} \max\paren{ \frac{C d}{\pi_{\min}^3 n}, \frac{1}{4k \kappa} } \sum_{j': j' \neq j} \| v_{j, j'} - \vstar_{j, j'} \|^2 + C' \sigma^2 \frac{kd}{\pi_{\min}^3 n} \log(n/d),
\end{align*}
where in step $\1$, we have used the definition of the quantity $r_a$.
Using this bound for the indices $j, \ell$ in conjunction with the definition of the quantity $\kappa$ proves inequality~\eqref{eq:1step-thm1-i}. 
\qed

\paragraph{Proof of part (b):} We now work on the event $\Espace(r_b)$ and proceed again from the bound
\begin{align*}
\| \beta^+_j - \appsigstar_j \|^2 &\leq C \max\paren{ \frac{d}{\pi_{\min}^3 n}, \frac{r_b}{\pi_{\min}^3} \log^{3/2} (1/r_b) } \sum_{j': j' \neq j} \| v_{j, j'} - \vstar_{j, j'} \|^2 + \frac{C}{\pimin^3 n} \| P_{S_j} \epsilon_{S_j} \|^2.
\end{align*}
Summing over $j \in [k]$ and using the Cauchy-Schwarz inequality, we obtain
\begin{align*}
\sum_{j = 1}^k \| \beta^+_j - \appsigstar_j \|^2 &\leq C \max\paren{ \frac{kd}{\pi_{\min}^3 n}, \frac{kr_b}{\pi_{\min}^3} \log^{3/2} (1/r_b) } \left( \sum_{j = 1}^k \| \beta_j - \appsigstar_j \|^2 \right) + \frac{C}{\pimin^3 n} \sum_{j \in [k]} \| P_{S_j} \epsilon_{S_j} \|^2 \\
&\stackrel{\2}{\leq} \frac{3}{4} \left( \sum_{j = 1}^k \| \beta_j - \appsigstar_j \|^2 \right) + C' \sigma^2 \frac{kd}{\pi_{\min}^3 n} \log(k) \log(n/kd),
\end{align*}
where in step $\2$, we have used the definition of the quantity $r_b$, the bound $n \geq Ckd / \pi_{\min}^3$, and claim~\eqref{clm:one-i}. This completes the proof.
\qed

\noindent We now prove each of the claims in turn. This constitutes the technical meat of our proof, and involves multiple technical lemmas whose proofs are postponed to the end of the section.

\paragraph{Proof of claims~\eqref{clm:one-i} and~\eqref{clm:one-ii}:} We begin by stating a general lemma about concentration properties of the noise.
\begin{lemma} \label{lem:noise}
Consider a random variable $z \in \real^n$ with i.i.d. $\sigma$-sub-Gaussian entries, and a fixed matrix $\AppXmat \in \real^{\numobs \times (\numdim + 1)}$. Then, we have
\begin{subequations}
\begin{align} \label{eq:noise-1}
\sup_{\beta_1,\dots, \beta_k \in \real^{d+1}} \sum_{j = 1}^k \; \| P_{\Xi^j(\beta_1,\dots, \beta_k)} z \|^2 \leq 2 \sigma^2 k (d+1) \log (kd) \log(n/kd)
\end{align}
with probability greater than $1 - \binom{n}{kd}^{-1}$ and
\begin{align} \label{eq:noise-2}
\sup_{\beta_1,\dots, \beta_k \in \real^{d+1}} \; \max_{j \in [k]} \; \| P_{\Xi^j(\beta_1,\dots, \beta_k)} z_{S_j(\beta_1,\dots, \beta_k)} \|^2 \leq 2 \sigma^2 k (d+1) \log(n/d)
\end{align}
\end{subequations}
with probability greater than $1 - \binom{n}{d}^{-1}$.
\end{lemma}
The proof of the claims follows directly from Lemma~\ref{lem:noise}, since the noise vector $\epsilon$ is independent of the matrix $\AppXmat$, and $\Ispace_b \subseteq \left(\real^{d+1}\right)^{\otimes k}$.
\qed

\paragraph{Proof of claim~\eqref{clm:two}:} We now state a lemma that directly handles indicator functions as they appear in the claim.

\begin{lemma} \label{lem:lin-affine}
Suppose that the covariates are drawn i.i.d. from a standard Gaussian distribution. Also, let $\ustar \in \real^d$ and $\wstar \in \real$, and consider a fixed parameter $\vstar = (\ustar, \; \wstar ) \in \real^{\numdim+1}$. Then there are universal constants $(c_1, c_2, c_3, c_4)$ such that for all positive scalars $r \leq 1/24$, we have
\begin{align*}
\sup_{v \in \Bspace_{\vstar} (r) } \left( \frac{1}{\numobs} \sum_{i = 1}^n \ind{ \inprod{\xi_i}{v} \cdot \inprod{\xi_i}{\vstar} \leq 0 } \inprod{\xi_i}{\vstar}^2 \right) / \| v - \vstar \|^2 \leq c_1  \cdot \max \left\{ \frac{d}{n}, r \log^{3/2} \left( \frac{1}{r} \right) \right\}
\end{align*}
with probability exceeding $1 - c_1e^{-c_2 \max \paren{ d, 10 \log n }} - c_3 ne^{-c_4 n}$. Here, we adopt the convention that $0/0 = 0$.
\end{lemma}
Applying Lemma~\ref{lem:lin-affine} with $v = v_{j, j'}$ and $\vstar = \vstar_{j, j'}$ for all pairs $(j, j')$ and using a union bound
directly yields the claim.
\qed

\paragraph{Proof of claim~\eqref{clm:three}:} For this claim, we state three technical lemmas pertaining to the singular values of random matrices whose rows are formed by truncated Gaussian random vectors.
We let $\vol(K)$ denote the volume of a set $K \subseteq \real^\numdim$ with respect to $d$-dimensional standard Gaussian measure, i.e., with $\vol(K) = \Pr\{ Z \in K \}$ for $Z \sim \NORMAL(0, I_d)$. 
\begin{lemma} \label{lem:sing}
Suppose $n$ vectors $\{\meas_i \}_{i = 1}^n$ are drawn i.i.d. from $\NORMAL(0, I_d)$, and $K \subseteq \real^d$ is a fixed convex set. Then there exists a tuple of universal constants $(c_1, c_2)$ such that if $\vol^3(K) n \geq c_1 \numdim \log^2 \left(1 / \vol(K) \right)$, then
\begin{align*}
\lambda_{\min} \left( \sum_{i: \meas_i \in K} \appmeas_i \appmeas_i^\top \right) \geq c_2 \vol^3(K) \cdot n
\end{align*}
with probability greater than $1 - c_1 \exp \left( -c_2 \numobs \frac{\vol^4(K)}{\log^2 (1 / \vol(K))} \right) - c_1 \exp( - c_2 n \cdot \vol(K))$.
\end{lemma}

For a pair of scalars $(w, w')$ and $d$-dimensional vectors $(u, u')$, define the \emph{wedge} formed by the $d+1$-dimensional vectors $v = (u, \; w)$ and  $v' = (u', \; w')$ as the region
\begin{align*}
W(v, v') = \{ x \in \real^d: (\inprod{x}{u} + w) \cdot (\inprod{x}{u'} + w') \leq 0\},
\end{align*}
and let $\Wspace_{\delta} = \{W = W(v, v'): \vol(W) \leq \delta \}$ denote the set of all wedges with Gaussian volume less than $\delta$. 
%
%
%
The next lemma bounds the maximum singular value of a sub-matrix formed by any such wedge.
\begin{lemma} \label{lem:wedge}
Suppose that $n$ vectors $\{\meas_i \}_{i = 1}^n$ drawn i.i.d. from $\NORMAL(0, I_d)$. Then there is a tuple of universal constants $(c_1, c_2)$ such that if $\numobs \geq c_1 \numdim \log (\numobs/\numdim)$, then
\begin{align*}
\sup_{W \in \Wspace_{\delta}} \lambda_{\max} \left( \sum_{i: \meas_i \in W} \appmeas_i \appmeas_i^\top \right) \leq c_1 \left(\delta \numobs + \numdim +  \numobs  \delta \log(1/ \delta) \right )
\end{align*}
with probability greater than $1 - 2\exp(-c_2 \delta \numobs) - \binom{n}{c_2 \delta n}^{-1}$.
\end{lemma}

We are now ready to proceed to a proof of claim~\eqref{clm:three}.
For convenience, introduce the shorthand notation
\begin{align*}
S^*_j \defn S_j \left(\appsigstar_1, \ldots, \appsigstar_k \right)
\end{align*}
to denote the set of indices corresponding to observations generated by the true parameter $\appsigstar_j$. Letting $A \Delta B \defn \left( A \setminus B \right) \bigcup \left( B \setminus A \right)$ denote the symmetric difference between two sets $A$ and $B$, we have
\begin{align*}
\lambda_{\min}\left( \AppXmat^\top_{S_j} \AppXmat_{S_j} \right) &\geq \lambda_{\min}\left( \AppXmat^\top_{S^*_j} \AppXmat_{S^*_j} \right) - \lambda_{\max}\left( \AppXmat^\top_{S^*_j \Delta S_j} \AppXmat_{S^*_j \Delta S_j} \right).
\end{align*}
Recall that by definition, we have 
\begin{align}
S^*_j \Delta S_j &= \{i: \inprod{\appmeas_i}{\appsigstar_j} \} = \max \text{ and } \inprod{\appmeas_i}{\beta_j} \neq \max \} \bigcup \{i: \inprod{\appmeas_i}{\appsigstar_j} \neq \max \text{ and } \inprod{\appmeas_i}{\beta_j} = \max \} \notag \\
&\subseteq \bigcup_{j' \in [k] \setminus j} \{i: \inprod{\appmeas_i}{\vstar_{j, j'}} \cdot \inprod{\appmeas_i}{v_{j, j'}} < 0 \} \notag \\
&= \bigcup_{j' \in [k] \setminus j} \{i: x_i \in W \left(\vstar_{j, j'}, v_{j,j'} \right) \}. \label{eq:symmdiff}
\end{align}
Putting together the pieces, we have
\begin{align}
\lambda_{\min}\left( \AppXmat^\top_{S_j} \AppXmat_{S_j} \right) \geq \lambda_{\min}\left( \AppXmat^\top_{S^*_j} \AppXmat_{S^*_j} \right) - \sum_{j' \neq j} \lambda_{\max} \left( \sum_{i: x_i \in W \left(\vstar_{j, j'}, v_{j,j'} \right) } \appmeas_i \appmeas_i^\top \right). \label{eq:singfinal}
\end{align}
Conditioned on the event guaranteed by Lemma~\ref{lem:wedge} with $\delta = \vol \left( W \left(\vstar_{j, j'}, v_{j,j'} \right) \right)$ and for a universal constant $C_1$, we have the bound
\begin{align*}
&\sup_{v_{j, j'} \in \Bspace_{\vstar_{j, j'}}(r_0) } \lambda_{\max} \left( \sum_{i: x_i \in W \left(\vstar_{j, j'}, v_{j,j'} \right) } \appmeas_i \appmeas_i^\top \right) \\
&\qquad \qquad \leq \sup_{v_{j, j'} \in \Bspace_{\vstar_{j, j'}}(r_0) } C_1 ( n \vol(W \left(\vstar_{j, j'}, v_{j,j'} \right)) \log (1 / \vol(W \left(\vstar_{j, j'}, v_{j,j'} \right)) ) + d )\\
&\qquad \qquad \stackrel{\1}{\leq} \sup_{v_{j, j'} \in \Bspace_{\vstar_{j, j'}}(r_0) } C_1 \left( n \frac{\eucnorm{ v_{j, j'} - \vstar_{j, j'} } }{ \eucnorm{ \ustar_{j, j'}} } \log^{3/2} \frac{\eucnorm{ \ustar_{j, j'}}}{\eucnorm{ v_{j, j'} - \vstar_{j, j'} }} + d \right) \\
&\qquad \qquad \stackrel{\2}{\leq} n r_0 \log^{3/2} (1/r_0) + d \\
&\qquad \qquad \stackrel{\3}{\leq} n C \frac{\pi_{\min}^3}{k},
\end{align*}  
where in step $\1$, we have used Lemma~\ref{lem:vol_gauss}, and in step $\2$, we have used the definition of the set $\Bspace$. Step $\3$ uses the assumption $n \geq c_1 kd / \pi_{\min}^3$.

Moreover, Lemma~\ref{lem:sing} guarantees the bound $\lambda_{\min}\left( \AppXmat^\top_{S^*_j} \AppXmat_{S^*_j} \right) \geq c_2 n \cdot \pimin^3$, so that putting together the pieces, we have
\begin{align}
\inf_{\beta_1, \ldots, \beta_k \in \Ispace_b } \; \lambda_{\min}\left( \AppXmat^\top_{S_j} \AppXmat_{S_j} \right) &\geq c_2 n \pimin^3 - C n k \frac{\pi^3_{\min}}{k} \notag \\
&\geq C \pimin^3 n, \label{eq:sing}
\end{align}
with probability greater than $1 - c \exp \left( -c \numobs \frac{\pi_{\min}^4}{\log^2 (1 / \pi_{\min})} \right) - c' \exp( - c' n \cdot \pi_{\min})$.
These assertions hold provided
\mbox{$n \geq C d \max \paren{ \frac{k}{\pi_{\min}^3}, \frac{\log^2 (1 / \pi_{\min}) }{\pi_{\min}^3}, \log (n / d) }$}, and this completes the proof.
\qed

\noindent Having proved the claims, we turn to proofs of our technical lemmas.

\subsubsection{Proof of Lemma~\ref{lem:noise}}

In this proof, we assume that $\sigma = 1$; our bounds can finally be scaled by $\sigma^2$.

It is natural to prove the bound~\eqref{eq:noise-2} first followed by bound~\eqref{eq:noise-1}.
First, consider a fixed set of parameters $\{ \beta_1, \ldots, \beta_k \}$. Then, we have
\begin{align*}
\eucnorm{ P_{\Xi^j(\beta_1,\dots, \beta_k)} z_{S^j} }^2 = \eucnorm{ UU^\top z_{S^j} }^2,
\end{align*}
where $U \in \real^{|\Xi^j| \times (d+1)}$ denotes a matrix with orthonormal columns that span the range of $\Xi^j(\beta_1,\dots, \beta_k)$.

Applying the Hanson-Wright inequality for independent sub-Gaussians (see \cite[Theorem 2.1]{rudelson2013hanson}) and noting that $\fronorm{UU^\top} \leq \sqrt{d+1}$ we obtain
\begin{align*}
\Pr \paren{ \eucnorm{ UU^\top z_{S^j} }^2 \geq (d + 1) + t } \leq e^{-ct / (d+1)},
\end{align*}
for each $t \geq 0$. In particular, this implies that the random variable $\eucnorm{ UU^\top z_{S^j} }^2$ is sub-exponential. 

This tail bound holds for a fixed partition of the rows of $\Xi$; we now take a union bound over all possible partitions. Toward that end, define the sets
\begin{align*}
\Sspace^j &= \paren{ S_j(\beta_1, \ldots, \beta_k) : \beta_1, \ldots, \beta_k \in \real^{d+1} }, \text{ for each } j \in [k].
\end{align*}

From Lemma~\ref{lem:sauer-shelah}, we have the bound $|\Sspace^j| \leq 2^{c k d \log (en/d)}$.
Thus, applying the union bound, we obtain
\begin{align*}
\Pr \paren{ \sup_{\beta_1, \ldots, \beta_k \in \real^{d+1}} \eucnorm{ P_{\Xi^j(\beta_1,\dots, \beta_k)} z_{S^j} }^2 \geq (d + 1) + t } &\leq | \Sspace^j | e^{-c t / (d + 1)},
\end{align*}
and substituting $t = c k (d+1) \log (n/d)$ and performing some algebra establishes bound~\eqref{eq:noise-2}.

In order to establish bound~\eqref{eq:noise-1}, we once again consider the random variable $\sum_{j = 1}^k \eucnorm{ P_{\Xi^j(\beta_1,\dots, \beta_k)} z_{S^j} }^2$ for a fixed set of parameters $\{ \beta_1,\dots, \beta_k \}$. Note that this is the sum of $k$ independent sub-exponential random variables and can be thought of as a quadratic form of the entire vector $z$. So once again from the Hanson-Wright inequality, we have
\begin{align*}
\Pr \paren{ \sup_{\beta_1, \ldots, \beta_k \in \real^{d+1}} \sum_{j = 1}^k \; \eucnorm{ P_{\Xi^j(\beta_1,\dots, \beta_k)} z_{S^j} }^2 \geq k (d + 1) + t } &\leq e^{- c t / k (d + 1) }
\end{align*}
for all $t \geq 0$.

Also define the set of all possible partitions of the $n$ points via the max-affine function; we have the set
\begin{align*}
\Sspace &= \paren{ S_1(\beta_1, \ldots, \beta_k), \ldots, S_k(\beta_1, \ldots, \beta_k) : \beta_1, \ldots, \beta_k \in \real^{d+1} }.
\end{align*}
Lemma~\ref{lem:arg_max_class} yields the bound $| \Sspace | \leq 2^{c k d \log (kd) \log (n/kd)}$, and combining a union bound with the high probability bound above establishes bound~\eqref{eq:noise-1} after some algebraic manipulation.
\qed

\subsubsection{Proof of Lemma~\ref{lem:lin-affine}}

Let $\gamma_v = v - \vstar$; we have
\begin{align*}
\ind{ \inprod{\xi_i}{v} \cdot \inprod{\xi_i}{\vstar} \leq 0 } \inprod{\xi_i}{\vstar}^2 &\leq \ind{ \inprod{\xi_i}{v} \cdot \inprod{\xi_i}{\vstar} \leq 0 } \inprod{\xi_i}{\gamma_v}^2 \\
&\leq \ind{ \inprod{\xi_i}{\gamma_v}^2 \geq \inprod{\xi_i}{\vstar}^2 } \inprod{\xi_i}{\gamma_v}^2.
\end{align*}
Define the (random) set $K_v = \{i: \inprod{\xi_i}{\gamma_v}^2 > \inprod{\xi_i}{\vstar}^2  \}$; we have the bound
\begin{align*}
\frac{1}{n} \sum_{i = 1}^n \ind{ \inprod{\xi_i}{v} \cdot \inprod{\xi_i}{\vstar} \leq 0 } \inprod{\xi_i}{\vstar}^2 \leq \frac{1}{n} \| \Xi_{K_v} \gamma_v \|^2.
\end{align*}
We now show that the quantity $\| \Xi_{K_v} \gamma_v \|^2$ is bounded uniformly for all $v \in \Bspace_{\vstar}(r)$ for small enough $r$. 
Recall that $\ustar$ is the ``linear" portion of $\vstar$, and let $m = \max\{ d, 10 \log n, n \cdot (16 r \cdot \sqrt{\log (1 / r )} \}$ (note that $m$ depends implicitly on $r$). We claim that for all $r \in (0, 1/24]$, we have
\begin{subequations}
\begin{align}
\Pr \left\{ \sup_{v \in \Bspace_{\vstar}(r)} |K_v| > m \right\} &\leq 4e^{-c \max \paren{ d, 10 \log n }} + cne^{-c' n }, \text{ and } \label{clm:set1a}\\
\Pr \left\{ \bigcup_{\substack{T \subseteq [n]: \\ |T| \leq m}} \sup_{ \substack{ \omega \in \real^{d + 1} \\ \omega \neq 0}}  \frac{\| \Xi_T \omega \|^2}{\| \omega \|^2} \geq (d + 16m \log(n/m) ) \right\} &\leq e^{-c \max \paren{ d, 10 \log n }}. \label{clm:set2a}
\end{align}
\end{subequations}
Taking these claims as given, the proof of the lemma is immediate, since $\frac{n}{m} \leq \frac{1}{16 r \log (1/r)}$, so that $\log (n / m) \leq C \log(1 / r)$.

\paragraph{Proof of claim~\eqref{clm:set1a}:}

By definition of the set $K_v$, we have
\begin{align*}
\Pr \{ \sup_{v \in \Bspace_{\vstar}(r) } |K_v| > m \} &\leq \sum_{ \substack{T \subseteq [n]: \\ |T| > m}} \Pr \left\{ \exists v \in \Bspace_{\vstar}(r) : \| \Xi_T \gamma_v \|^2 \geq \| \Xi_T \vstar\|^2 \right\} \\
&= \sum_{ \substack{T \subseteq [n]: \\ |T| > m}} \Pr \left\{ \exists v \in \Bspace_{\vstar}(r) : \frac{\| \gamma_v \|^2}{ \| \ustar \|^2 } \frac{\| \Xi_T \gamma_v \|^2}{\| \gamma_v \|^2} \geq \frac{\| \Xi_T \vstar\|^2}{\| \ustar \|^2} \right\} \\
&\leq \sum_{ \substack{T \subseteq [n]: \\ |T| > m}} \Pr \left\{ \exists v \in \Bspace_{\vstar}(r) : r^2 \frac{\| \Xi_T \gamma_v \|^2}{\| \gamma_v \|^2} \geq \frac{\| \Xi_T \vstar\|^2}{\| \ustar \|^2} \right\} \\
&\leq \sum_{\substack{T \subseteq [n]: \\ |T| > m}} \Bigg( \Pr \left\{ \exists v \in \Bspace_{\vstar}(r) : \frac{\| \Xi_T \gamma_v \|^2}{\| \gamma_v \|^2} \geq (\sqrt{d} + \sqrt{|T|} + t_T)^2 \right\} \\
&\qquad \qquad \quad + \Pr \left\{ \frac{\| \Xi_T \vstar\|^2}{\| \ustar \|^2} \leq r^2 (\sqrt{d} + \sqrt{|T|} + t_T)^2 \right\} \Bigg),
\end{align*}
where the final step follows by the union bound and holds for all positive scalars $\{t_T\}_{T \subseteq [n]}$.
For some \emph{fixed} subset $T$ of size $\ell$, we have the tail bounds
\begin{subequations}
\begin{align}
\Pr \paren{ \sup_{ \substack{ \omega \in \real^{d+ 1} \\ \omega \neq 0} } \frac{\| \Xi_T \omega \|^2}{\|\omega \|^2} (\sqrt{d} + \sqrt{\ell} + t)^2 } &\stackrel{\1}{\leq} 2e^{-t^2/2}, \text{ for all } t \geq 0, \text{ and } \label{eq:tb1}\\
\Pr \paren{ \frac{\| \Xi_T \vstar\|^2}{\| \ustar\|^2} \leq \delta \ell } &\stackrel{\2}{\leq} (e \delta)^{\ell/2} \text{ for all } \delta \geq 0, \label{eq:tb2}
\end{align}
\end{subequations}
where step $\1$ follows from the sub-Gaussianity of the covariate matrix (see Lemma~\ref{lem:sgmax}), and step $\2$ from a tail bound for the non-central $\chi^2$ distribution (see Lemma~\ref{lem:lowertail}).

Substituting these bounds yields
\begin{align*}
\Pr \{ \sup_{v \in \Bspace_{\vstar}(r) } |K_v| > m \} &\leq \sum_{\ell = m+1}^n  \binom{n}{\ell} \left[ 2e^{-t_\ell^2/2} + \left( e  r^2 \cdot \frac{(\sqrt{d} + \sqrt{\ell} + t_\ell)^2}{\ell} \right)^{\ell/2} \right] \\
&\leq \sum_{\ell = m+1}^n  \binom{n}{\ell} \left[ 2e^{-t_\ell^2/2} + \left( 2  r \cdot \frac{\sqrt{d} + \sqrt{\ell} + t_\ell}{\sqrt{\ell}} \right)^{\ell} \right].
\end{align*}
Recall that $t_\ell$ was a free (non-negative) variable to be chosen. We now split the proof into two cases and choose this parameter differently for the two cases.

\paragraph{Case 1, $m \leq \ell < n/e$:}

 Substituting the choice $t_\ell = 4 \sqrt{\ell \log (n / \ell) }$, we obtain
\begin{align*}
\binom{n}{\ell} \left[ 2e^{-t_\ell^2/2} + \left( 2  r \cdot \frac{\sqrt{d} + \sqrt{\ell} + t_\ell}{\sqrt{\ell}} \right)^{\ell} \right] 
&\leq \left( \frac{n}{\ell} \right)^{-c\ell} + \binom{n}{\ell} \cdot \left( 2  r \cdot \frac{\sqrt{d} + 5 \sqrt{\ell \log (n / \ell)} }{\sqrt{\ell} } \right)^{\ell} \\
&\stackrel{\1}{\leq}  \left( \frac{n}{\ell} \right)^{-c\ell} + \binom{n}{\ell} \cdot \left( 2  r \cdot (1 + 5 \sqrt{\log (n / \ell)} ) \right)^{\ell} \\
&\stackrel{\2}{\leq}  \left( \frac{n}{\ell} \right)^{-c\ell} + \binom{n}{\ell} \cdot \left(12 r \cdot \sqrt{\log (n / \ell)} \right)^{\ell} \\
&\leq  \left( \frac{n}{\ell} \right)^{-c\ell} + \left(12 \left(\frac{e n}{\ell}\right) r \cdot \sqrt{\log (n / \ell)} \right)^{\ell},
\end{align*}
where step $\1$ follows from the bound $m \geq d$, and step $\2$ from the bound $\ell \leq n/e$.

Now note that the second term is only problematic for small $\ell$. For all $\ell \geq m = n \cdot (16 r \cdot \sqrt{\log (1 / r) } )$, we have
\begin{align*}
\left(12 \left(\frac{e n}{\ell}\right) r \cdot \sqrt{\log (n / \ell)} \right)^{\ell} &\leq \left( 3/4 \right)^{\ell}.
\end{align*}
The first term, on the other hand, satisfies the bound $\left( \frac{n}{\ell} \right)^{-c\ell} \leq \left( 3/4 \right)^{\ell}$ for sufficiently large $n$. 

\paragraph{Case 2, $\ell \geq n/e$:}
In this case, setting $t_{\ell} = 2\sqrt{n}$ for each $\ell$ yields the bound
\begin{align*}
\binom{n}{\ell} \left[ 2e^{-t_\ell^2/2} + \left( 2  r \cdot \frac{\sqrt{d} + \sqrt{\ell} + t_\ell}{\sqrt{\ell}} \right)^{\ell} \right] &\leq 2 \binom{n}{n/2} e^{-2n} + \left( 12  r  \right)^{\ell} \\
&\leq c e^{-c' n},
\end{align*}
where we have used the fact that $d \leq n/2$ and $r \leq 1/24$.

\vspace{2mm}

Putting together the pieces from both cases, we have shown that for all $r \in (0, 1/24]$, we have
\begin{align*}
\Pr \{ \sup_{v \in \Bspace_{\vstar}(r) } |K_v| > m \} &\leq c n e^{-c' n} + \sum_{\ell = m+1}^{n/e} (3/4)^\ell \\
&\leq c n e^{-c' n} + 4 (3/4)^{\max \paren{ d, 10 \log n }},
\end{align*}
thus completing the proof of the claim.

\paragraph{Proof of claim~\eqref{clm:set2a}:}

The proof of this claim follows immediately from the steps used to establish the previous claim. In particular, writing
\begin{align*}
&\Pr \left\{ \bigcup_{\substack{T \subseteq [n]: \\ |T| \leq m }} \bigcup_{\omega: \| \omega\| = 1 } \| \Xi_T \omega \|^2 \geq d + 16m \log (n/m)\right\} \\
&\qquad \qquad \leq \Pr \left\{ \bigcup_{\substack{T \subseteq [n]: \\ |T| \leq m }} \bigcup_{\omega: \| \omega\| = 1 } \| \Xi_T \omega \|^2 \geq \left(\sqrt{d} + \sqrt{m} + \sqrt{4 m \log(n/m)} \right)^2 \right\} \\
&\qquad \qquad \leq \sum_{\ell = 1}^m \Pr \left\{ \bigcup_{\substack{T \subseteq [n]: \\ |T| = \ell }} \bigcup_{\omega: \| \omega \| = 1 } \| \Xi_T \omega \|^2 \geq \left(\sqrt{d} + \sqrt{m} + \sqrt{4 m \log(n/m)} \right)^2 \right\} \\
&\qquad \qquad \stackrel{\4}{\leq} 2 \sum_{\ell = 1}^m \binom{n}{\ell} \exp\{ - 2 m \log (n/m) \} \\
&\qquad \qquad \leq 2 \left( \frac{n}{m} \right)^{-cm} \leq 2 e^{-c\max \paren{ d, 10 \log n } },
\end{align*}
where step $\4$ follows from the tail bound~\eqref{eq:tb1}.
\qed

\subsubsection{Proof of Lemma~\ref{lem:sing}} \label{sec:truncmom}
The lemma follows from some structural results on the truncated Gaussian distribution. Using the shorthand $\vol \defn \vol(K)$ and letting $\psi$ denote the $d$-dimensional Gaussian density, consider a random vector $\tau$ drawn from the distribution having density $h(y) = \frac{1}{\vol} \psi(y) \ind{ y \in K }$, and denote its mean and second moment matrix by $\mutil$ and $\Sigmatil$, respectively. Also denote the recentered random variable by $\tautil = \tau - \mutil$. We claim that
\begin{subequations}
\begin{align}
\| \mutil \|^2 &\leq C \log \left(1 / \vol \right), \label{clm:mean} \\
C \vol^2 \cdot I \preceq &\Sigmatil \preceq \left( 1 + C \log(1 / \vol ) \right) I, \text{ and } \label{clm:sm} \\
\tautil \text{ is $c$-sub-Gaussian } &\text{for a universal constant } c. \label{clm:sg}
\end{align} 
\end{subequations}
Taking these claims as given for the moment, let us prove the lemma.

The claims~\eqref{clm:mean} and~\eqref{clm:sg} taken together imply that the random variable $\tau$ is sub-Gaussian with $\psi_2$ parameter $\zeta^2 \leq 2c^2 + 2C \log \left(1 / \vol \right)$. Now consider $m$ i.i.d. draws of $\tau$ given by $\{\tau_i\}_{i = 1}^m$; standard results (see, e.g., Vershynin~\cite[Remark 5.40]{vershynin2010introduction}, or Wainwright~\cite[Theorem 6.2]{wainwright2019high}) yield the bound
\begin{align*}
\Pr \left\{ \opnorm{ \frac{1}{m} \sum_{i = 1}^m \tau_i \tau_i^\top - \Sigmatil } \geq \zeta^2 \left( \frac{d}{m} + \sqrt{\frac{d}{m}} + \delta \right) \right\} \leq 2 \exp\left( -c n \min\{ \delta, \delta^2\} \right).
\end{align*}
Using this bound along with claim~\eqref{clm:sm} and Weyl's inequality yields
\begin{align}
\lambda_{\min} \left( \frac{1}{m} \sum_{i = 1}^m \tau_i \tau_i^\top \right) \geq C \vol^2 - \zeta^2 \left( \frac{d}{m} + \sqrt{\frac{d}{m}} + \delta \right) \label{eq:weyl}
\end{align}
with probability greater than $1 - 2 \exp\left( -c n \min\{ \delta, \delta^2\} \right)$.

Furthermore, when $n$ samples are drawn from a standard Gaussian distribution, the number $m$ of them that fall in the set $K$ satisfies $m \geq \frac{1}{2} n \cdot \vol$ with high probability. In particular, this follows from a straightforward binomial tail bound, which yields
\begin{align}
\Pr \left\{ m \leq \frac{n \cdot \vol }{2} \right\} \leq \exp( - c n \cdot \vol). \label{eq:bintail}
\end{align}
Recall our choice $n \geq C d \frac{\log^2 (1 / \vol)}{\vol^3}$, which in conjunction with the bound~\eqref{eq:bintail} ensures that $C \vol^2 \geq \frac{1}{8} \sigma^2 \sqrt{\frac{\numdim}{m} }$ with high probability. 
 Setting $\delta = C \vol^2 / \sigma^2$ in inequality~\eqref{eq:weyl}, we have
\begin{align*}
\lambda_{\min} \left( \frac{1}{m} \sum_{i = 1}^m \tau_i \tau_i^\top \right) \geq \frac{C}{2} \vol^2
\end{align*}
with probability greater than $1 - 2 \exp \left( - c n \vol^4 / \sigma^4 \right)$.
Putting together the pieces thus proves the lemma. It remains to show the various claims. \qed

\paragraph{Proof of claim~\eqref{clm:mean}}
Let $\tau_{\Aspace}$ denote a random variable formed as a result of truncating the Gaussian distribution to a (general) set $\Aspace$ with volume $\vol$. Letting $\mu_\Aspace$ denote its mean, the dual norm definition of the $\ell_2$ norm yields
\begin{align*}
\| \mu_\Aspace \| &= \sup_{v \in \Sd } \inprod{v}{\mu_\Aspace} \\
&\leq \sup_{v \in \Sd } \E |\inprod{v}{\tau_\Aspace}|.
\end{align*}
Let us now evaluate an upper bound on the quantity $\E |\inprod{v}{\tau_\Aspace}|$. In the calculation, for any $d$-dimensional vector $y$, we use the shorthand $y_v \defn v^\top y$ and $y_{\setminus v} \defn U^\top_{\setminus v} y$ for a matrix $U_{\setminus v} \in \real^{d \times (d-1) }$ having orthonormal columns that span the subspace orthogonal to $v$. Letting $\Aspace_v \subseteq \real$ denote the projection of $\Aspace$ onto the direction $v$, define the set $\Aspace_{\setminus v} (w) \subseteq \real^{d-1}$ via 
\begin{align*}
\Aspace_{\setminus v} (w) = \{ y_{\setminus v} \in \real^{d-1}: y \in \Aspace \text{ and } y_v = w \}.
\end{align*} 
Letting $\psi_d$ denote the $d$-dimensional standard Gaussian pdf, we have
\begin{align}
\E |\inprod{v}{\tau_\Aspace}| &= \frac{1}{\vol} \int_{y \in \Aspace} |y^\top v| \psi_d(y) dy \notag \\
&= \frac{1}{\vol} \int_{y \in \Aspace} |y_v| \psi(y_v) \psi_{d-1}(y_{\setminus v}) dy \notag \\
&= \frac{1}{\vol} \int_{y_v \in \Aspace_v} |y_v| \psi(y_v) \underbrace{ \left( \int_{y_{\setminus v} \in \Aspace_{\setminus v}(y_v)} \psi_{d-1}(y_{\setminus v} \in \Aspace_{\setminus v}(y_v)) dy_{\setminus v} \right)}_{f(y_v)} dy_v \notag \\
&\stackrel{\1}{\leq} \frac{1}{\vol} \int_{y_v \in \Aspace_v} |y_v| \psi(y_v) dy_v, \label{eq:smbd}
\end{align}
where step $\1$ follows since $f(y_v) \leq 1$ point-wise. On the other hand, we have
\begin{align}
\vol = \int_{y_v \in \Aspace_v} \psi(y_v) \left(\int_{y_{\setminus v} \in \Aspace_{\setminus v}(y_v)} \psi_{d-1} dy_{\setminus v} \right) dy_v
\leq \int_{y_v \in \Aspace_v} \psi(y_v) dy_v. \label{eq:volbd}
\end{align}
Combining inequalities~\eqref{eq:smbd} and~\eqref{eq:volbd} and letting $w = y_v$, an upper bound on $\| \mu_{\tau} \|$ can be obtained by solving the one-dimensional problem given by
\begin{align*}
\| \mu_{\tau} \| \leq &\sup_{\Sspace \subseteq \real} \; \frac{1}{\vol} \int_{w \in \Sspace} |w| \psi(w) dw \\
&\text{ s.t. } \int_{w \in \Sspace} \psi(w) dw \geq \vol.
\end{align*}
It can be verified that the optimal solution to the problem above is given by choosing the truncation set $\Sspace = (\infty, -\beta) \cup [\beta, \infty)$ for some threshold $\beta > 0$. 
With this choice, the constraint can be written as
\begin{align*}
\vol \leq \int_{|w| \geq \beta} \psi(w) dw \leq 2 \sqrt{ \frac{2}{\pi} } \frac{1}{\beta} e^{-\beta^2/2},
\end{align*}
where we have used a standard Gaussian tail bound. Simplifying yields the bound
\begin{align*}
\beta \leq 2 \sqrt{\log (C / \vol)}.
\end{align*}
Furthermore, we have
\begin{align*}
\frac{1}{\vol} \int_{|w| \geq \beta} |w| \psi(w) dw &= \frac{C}{\vol} e^{-\beta^2 /2} \\
&\stackrel{\2}{\lesssim} \frac{\beta^3}{\beta^2 - 1} \\
&\leq c \sqrt{\log (1 / \vol)},
\end{align*}
where step $\2$ follows from the bound $\Pr\{Z \geq z\} \geq \psi(z) \left(\frac{1}{z} - \frac{1}{z^3} \right)$ valid for a standard Gaussian variate $Z$.
Putting together the pieces, we have
\begin{align*}
\| \mutil \|^2 \leq c \log (1 / \vol).
\end{align*}
\qed

\paragraph{Proof of claim~\eqref{clm:sm}}
Let us first show the upper bound. Writing $\cov(\tau)$ for the covariance matrix, we have
\begin{align*}
\opnorm{ \Sigmatil } &\leq \opnorm{ \cov(\tau)} + \| \mutil \|^2 \\
&\stackrel{\3}{\leq} \opnorm{ I} + C \log (1 / \vol ),
\end{align*}
where step $\3$ follows from the fact that $\cov(\tau) \preceq \cov(Z)$, since truncating a Gaussian to a convex set reduces its variance along all directions~\cite{kanter1977reduction, vempala2010learning}.

We now proceed to the lower bound.
Let $\Prob_K$ denote the Gaussian distribution truncated to the set $K$.
Recall that we denoted the probability that a Gaussian random variable falls in the set $K$ by $\vol(K)$; use the shorthand $\vol = \vol(K)$.
Define the polynomial
\begin{align*}
p_u (x) = \inprod{ x - \E_{X \sim \Prob_K}[X] }{u}^2;
\end{align*}
note that we are interested in a lower bound on $\inf_{u \in \Sd} \E_{X \sim \Prob_K} [p_u(X) ]$.

For $\delta > 0$, define the set 
\begin{align*}
S_{\delta} \defn \{ x \in \real^d: p_u (x) \leq \delta\} \subseteq \real^d.
\end{align*}
Letting $Z$ denote a $d$-dimensional standard Gaussian random vector and using the shorthand $\alpha \defn \E_{X \sim \Prob_K}[X]$, we have
\begin{align}
\Pr \{ Z \in S_\delta \} &= \Pr \paren{ \inprod{ Z - \alpha}{u}^2 \leq \delta } \\
&= \Pr \paren{ \inprod{\alpha}{u} - \sqrt{\delta} \leq \inprod{Z}{u} \leq \inprod{\alpha}{u} + \sqrt{\delta} } \\
&= \int_{\inprod{\alpha}{u} - \sqrt{\delta}}^{\inprod{\alpha}{u} + \sqrt{\delta}} \psi(x) dx \leq \sqrt{\frac{2}{\pi} \delta}, \label{eq:pbd}
\end{align}
where in the final step, we have used the fact that $\psi(x) \leq 1/\sqrt{2\pi}$ for all $x \in \real$.
Consequently, we have
\begin{align*}
\E_{X \sim \Prob_K} [p_u (X)] &= \frac{1}{\vol} \E_Z \left[ p_u (Z) \ind{ Z \in K} \right] \\
&\geq \frac{1}{\vol} \E_Z \left[ p_u (Z) \ind{ Z \in K \cap S_{\delta}^c} \right] \\
&\stackrel{\4}{\geq} \frac{1}{\vol} \E_Z \left[ \delta \ind{ Z \in K \cap S_{\delta}^c} \right] \\
&= \frac{\delta}{\vol} \Pr \{ Z \in K \cap S_{\delta}^c \} \\
&\stackrel{\5}{\geq} \delta \frac{\vol - \sqrt{\frac{2}{\pi} \delta} }{\vol}.
\end{align*}
Here, step $\4$ follows from the definition of the set $S_\delta$, which ensures that $p_u (x) \geq \delta$ for all $x \in S^c_{\delta}$. Step $\5$ follows as a consequence of equation~\eqref{eq:pbd}, since 
\[
\Pr\{ Z \in K \cap S_{\delta}^c \} = \Pr \{ Z \in K\} - \Pr\{Z \in S_\delta\} \geq \vol - \sqrt{\frac{2}{\pi} \delta}.
\]
Finally, choosing $\delta = c \vol^2$ for a suitably small constant $c$, we have
$\E_{X \sim \Prob_K} [p_u (X)] \geq C \vol^2$ for a fixed $u \in \Sd$. Since $u$ was chosen arbitrarily, this proves the claim. \qed

\paragraph{Proof of claim~\eqref{clm:sg}}
Since the random variable $\xi$ is obtained by truncating a Gaussian random variable to a convex set, it is $1$-strongly log-concave. Thus, standard results~\cite[Theorem 2.15]{ledoux2001concentration} show that the random variable $\xitil$ is $c$-sub-Gaussian.
\qed

\subsubsection{Proof of Lemma~\ref{lem:wedge}}

For a pair of $d+1$-dimensional vectors $(v, v')$, denote by
 \begin{align}
 n_{W(v,v')}= \# \{i:\meas_i \in W(v,v')\} \label{eq:nw}
\end{align}  
the random variable that counts the number of points that fall within the wedge $W(v,v')$; recall our notation $W_{\delta}$ for the set of all wedges with Gaussian volume less than $\delta$. Since each wedge is formed by the intersection of two hyperplanes, applying Lemmas~\ref{lem:uniform-convergence} and~\ref{lem:sauer-shelah} in conjunction yields that
there are universal constants $(c, c', C)$ such that
\begin{align} \label{eqn:wedge_card}
\sup_{W \in \mathcal{W}_\delta} n_{W} \leq c \delta \numobs
\end{align}
with probability exceeding $1- \exp(- c' \numobs \delta^2)$, provided $\numobs \geq \frac{C}{\delta^2}  \numdim \log(\numobs/\numdim)$. In words, the maximum number of points that fall in \emph{any} wedge of volume $\delta$ is linear in $\delta n$ with high probability.

It thus suffices to bound, simultaneously, the maximum singular value of every sub-matrix of $\AppXmat$ having (at most) $c \delta \numobs$ rows. For a \emph{fixed} subset $S$ of size $c \delta \numobs$, standard bounds for Gaussian random matrices (see, e.g., \cite{vershynin2010introduction})) yield the bound
\begin{align*}
\lambda_{\max} \left (\sum_{i\in S} \appmeas_i \appmeas_i^T \right ) \leq c_1(\delta \numobs + \numdim + t)
\end{align*}
with probability exceeding $1-2\exp(-C_1t)$.

Furthermore, there are at most $c\delta n \cdot \binom{n}{c\delta n}$ subsets of size at most $c\delta n$; taking a union bound over all such subsets yields the bound
\begin{align*}
\Pr \left\{ \max_{S: |S| \leq c\delta n} \lambda_{\max} \left (\sum_{i\in S} \appmeas_i \appmeas_i^{\top} \right) \geq c_1(\delta \numobs + \numdim + t) \right\} &\leq 2c \delta n \cdot \binom{n}{c\delta n} \exp(-C_1t). 
\end{align*}
Making the choice $t = 2c \delta n \log (1 / c\delta)$ and putting together the pieces proves the lemma.
\qed


\section{Proof of Theorem~\ref{thm:sbgen}} \label{sec:pf-thm2}

Once again, let us begin by stating a concise mathematical statement. With the shorthand
\begin{align*}
\delta^{\mathsf{sb}}_{n, \sigma}(d, k, \pi_{\min}) \defn \sigma^2 \frac{kd}{n \pimin^{1 + 2\sball^{-1} }} \log(kd) \log(n/kd)
\end{align*}
and with the rest of the notation remaining the same as before, the theorem claims that there exist constants such that if condition~\eqref{eq:n-req2} is satisfied, then we have
\begin{align} \label{eq:aditya-bound2}
&\Pr \paren{ \max_{t \geq 1} \sup_{\appsigstar_1, \ldots, \appsigstar_k \in \Bvol(\pimin, \Delta, \kappa) } \; \vartheta_t \left( \Csig^{(2)} \left(\frac{\pi_{\min}^{1 + 2 \sball^{-1}}}{ \kappa k}\right)^{\sball^{-1}} ; \paren{ \appsigstar_j }_{j = 1}^k \right) \geq \Csig^{(1)} \delta^{\mathsf{sb}}_{n, \sigma}(d, k, \pi_{\min}) } \notag \\
&\qquad \qquad \qquad \qquad \leq 1 - c_1 \paren{ \frac{k^2}{n^7} + \exp \left(- c_2 n \pi_{\min}^2 \right) }
\end{align}

We now proceed to a proof of this claim.
In order to promote readability, we keep the structure of the proof the same as that of Theorem~\ref{thm:mt}, while sketching the differences. We assume again that the scalar $c^* = 1$ without loss of generality. The result once again follows as a consequence of the following one-step proposition. 

\begin{proposition} \label{thm:onestep-cor}
Suppose that Assumption~\ref{assptn:gen} holds. Then there exists a tuple of universal constants $(c_1, c_2)$ and another tuple of constants $(\Csig^{(1)}, \Csig^{(2)})$ depending only on the tuple $(\eta, \sball, \cs)$ such that

(a) If the sample size satisfies the bound $n \geq c_1 \max \paren{ d, 10 \log n } \max \paren{ \frac{1 + \sball^{-1}}{\pimin}, \frac{k}{\pi_{\min}^2} \log (n / d) }$, then for any set of parameters $\paren{\appsigstar_j}_{j = 1}^k \in \Bvol(\pimin, \Delta, \kappa)$ and $\paren{ \beta_j }_{j = 1}^k$ satisfying 
\begin{subequations}
\begin{align}
\max_{1 \leq j \neq j' \leq k} \; \left( \frac{\left\| v_{j, j'} - \vstar_{j, j'} \right\|}{\| \sigstar_j -\sigstar_{j'} \| } \right)^{\sball} \log^{1 + \sball} \left( \frac{\| \sigstar_j -\sigstar_{j'} \| }{\left\| v_{j, j'} - \vstar_{j, j'} \right\|} \right) &\leq \Csig^{(2)} \left( \frac{\pi_{\min}^{1 + 2 \sball^{-1} }}{k \kappa} \right), \label{eq:init-thm2-i}
\end{align}
we have, simultaneously for all pairs $1 \leq j \neq \ell \leq k$, the bound
\begin{align}
&\frac{\left\| v^+_{j, \ell} - \vstar_{j, \ell} \right\|^2}{\| \sigstar_j -\sigstar_{\ell} \|^2} \leq \Csig^{(1)} \max \paren{ \frac{d \sball \kappa}{\pi_{\min}^{1 + 2 \sball^{-1}} n}, \frac{1}{4k} } \left( \sum_{j' = 1}^k \frac{ \eucnorm{ v_{j, j'} - \vstar_{j, j'} }^2}{ \| \sigstar_j -\sigstar_{j'} \|^2 } + \frac{ \eucnorm{ v_{\ell, j'} - \vstar_{\ell, j'} }^2}{ \| \sigstar_\ell -\sigstar_{j'} \|^2 } \right) \notag \\
& \qquad \qquad \qquad \qquad + \Csig^{(1)} \sigma^2 \frac{kd}{\pi^{1 + 2 \sball^{-1}}_{\min} n} \log(n/d) \label{eq:1step-thm2-i}
\end{align}
with probability exceeding $ 1 - c_1 \paren{ \frac{k^2}{n^7} + \exp \left(- c_2 n \pi_{\min}^2 \right) }$.
\end{subequations}

(b) If the sample size satisfies the bound 
\[
n \geq c_1 \max\paren{ \max \paren{ d, 10 \log n } \max \paren{ \frac{1 + \sball^{-1}}{\pimin}, \frac{k}{\pi_{\min}^2} \log (n / d) }, \Csig^{(1)} \frac{kd}{\pi_{\min}^{1 + 2\sball^{-1}}}}, 
\]
then for any set of parameters $\paren{\appsigstar_j}_{j = 1}^k \in \Bvol(\pimin, \Delta, \kappa)$ and $\paren{ \beta_j }_{j = 1}^k$ satisfying 
\begin{subequations}
\begin{align}
\max_{1 \leq j \neq j' \leq k} \; \left( \frac{\left\| v_{j, j'} - \vstar_{j, j'} \right\|}{\| \sigstar_j -\sigstar_{j'} \| } \right)^{\sball} \log^{1 + \sball} \left( \frac{\| \sigstar_j -\sigstar_{j'} \| }{\left\| v_{j, j'} - \vstar_{j, j'} \right\|} \right) &\leq \Csig^{(2)} \left( \frac{\pi_{\min}^{1 + 2 \sball^{-1} }}{k} \right), \label{eq:init-thm2-ii}
\end{align}
we have the overall estimation error bound
\begin{align}
\sum_{i = 1}^k \| \beta^+_j - \appsigstar_j \|^2 &\leq \frac{3}{4} \cdot \left( \sum_{i = 1}^k \| \beta_j - \appsigstar_j \|^2 \right) + \Csig^{(1)} \sigma^2 \frac{kd}{\pi^{1 + 2\sball^{-1}}_{\min} n} \log(kd) \log(n/kd) \label{eq:1step-thm2-ii}
\end{align}
\end{subequations}
with probability exceeding $ 1 - c_1 \paren{ \frac{k^2}{n^7} + \exp \left(- c_2 n \pi_{\min}^2 \right) }$.
\end{proposition}

The theorem follows from this proposition by a similar method to the proof of Theorem~\ref{thm:mt}, except that it now holds uniformly for all choices of the true parameters $\paren{\appsigstar_j}_{j = 1}^k \in \Bvol(\pimin, \Delta, \kappa)$.

\subsection{Proof of Proposition~\ref{thm:onestep-cor}}

Similarly to before, let $r_a$ be the largest scalar in the interval $[0, e^{- (1 + \sball^{-1}) } ]$ such that $r_a^{\sball} \log^{1 + \sball} (1/ r_a) \leq \Csig^{(2)} \left( \frac{\pimin^{1 + 2 \sball^{-1}}}{k \kappa} \right)$, and let $r_b$ be the largest scalar in that interval with $r_b^{\sball} \log^{1 + \sball} (1/ r_b) \leq \Csig^{(2)} \left( \frac{\pimin^{1 + 2 \sball^{-1}}}{k} \right)$.

With the same notation as in the proof of Theorem~\ref{thm:mt}, we arrive at the bound
\begin{align*}
\lambda_{\min} \left( \AppXmat^\top_{S_j} \AppXmat_{S_j} \right) \cdot \| \beta^+_j - \appsigstar_j \|^2 &\lesssim \sum_{j': j' \neq j} \sum_{i = 1}^n  \ind{ \inprod{\appmeas_i}{v_{j,j'}} \cdot \inprod{\appmeas_i}{\vstar_{j,j'}} \leq 0 } \inprod{\appmeas_i}{\vstar_{j,j'}}^2 + \| P_{S_j} \epsilon_{S_j} \|^2.
\end{align*}
The second term on the RHS of this bound is handled exactly as before. We now make two
claims to bound the remaining two terms in the bound.

Recall that Assumption~\ref{assptn:gen} holds, and use the shorthand $\etatil \defn \max \paren{1, \eta}$. Our first claim bounds the indicator quantities under this assumption. In particular, we claim that there exists a tuple of universal constants $(C, c_1, C', c')$ such that for any $r \leq 1/24$, we have
\begin{subequations}
\begin{align}
&\Pr \Big\{ \exists \appsigstar_1, \ldots, \appsigstar_k \in \real^{d+1} \text{ and } \beta_1, \ldots, \beta_k \text{ such that } v_{j, j'} \in \Bspace_{\vstar} (r) \text{ for all } 1 \leq j \neq j' \leq k: \notag \\
& \quad \quad \sum_{j': j' \neq j} \sum_{i = 1}^n  \ind{ \inprod{\appmeas_i}{v_{j, j'}} \cdot \inprod{\appmeas_i}{\vstar_{j, j'}} \leq 0 } \inprod{\meas_i}{\vstar_{j, j'}}^2 \notag \\
&\quad \qquad \geq C \etatil^2 \sum_{j': j' \neq j} \| v_{j, j'} - \vstar_{j, j'} \|^2 \max\paren{ (1 + \sball^{-1} ) \numdim, \; n \cdot (1 + \sball) \paren{ (\etatil r)^{\sball} \log^{\sball + 1} (1/ \etatil r)} }  \Big\} \notag \\ 
&\quad \quad \qquad \qquad \qquad \qquad \qquad \leq c_1 \binom{k}{2} \paren{ ne^{-C'n} + ce^{-c'\max \paren{ d, 10 \log n }} }. \label{clm:twoc}
\end{align}
The singular values of sub-matrices of $\AppXmat$ are then handled by the following claim: that there exist universal constants $(C, c, c')$ and a constant $\Csig'$ that depends only the tuple $(\eta, \sball, \cs)$, such that if  $n \geq C d \max \paren{ \frac{1 + \sball^{-1} }{\pi_{\min}}, \frac{k}{\pimin^2} \log (n / kd) }$, then
\begin{align}
&\Pr \left\{\inf_{\appsigstar_1, \ldots, \appsigstar_k \in \Bvol(\pimin, \Delta, \kappa)} \; \; \inf_{ \substack{ \beta_1, \ldots, \beta_k \in \real^{d + 1}: \\ v_{j, j'} \in \Bspace_{\vstar_{j, j'} }(r_b) } }
\; \; \min_{j \in [k] } \; \; \eigmin \left( \Xi^j(\beta_1, \ldots, \beta_k)^\top \cdot \Xi^j(\beta_1, \ldots, \beta_k) \right) \leq \Csig' \pi_{\min}^{1 + 2 \sball^{-1} } n \right\} \notag \\
&\qquad \qquad \qquad \leq c \exp( - c' n \cdot \pi_{\min}^2). \label{clm:threec}
\end{align}
\end{subequations}

The proof of the proposition from these claims follows exactly as in the proof of Theorem~\ref{thm:mt}, so we skip the details.
\qed

\noindent We now prove both the given claims in turn.

\paragraph{Proof of claim~\eqref{clm:twoc}:} This claim hinges on the following lemma that parallels Lemma~\ref{lem:lin-affine}. 

\begin{lemma} \label{lem:lin-affine-c}
Suppose that Assumption~\ref{assptn:gen} holds. Then, there exist universal constants $(C, c_1, c_2)$ such for all positive scalars $r \leq 1/24$, we have
\begin{align*}
\sup_{\substack{\vstar \in \real^{d+1} \\ v \in \Bspace_{\vstar}(r) } } & \left( \frac{1}{n} \sum_{i = 1}^n \ind{ \inprod{\xi_i}{v} \cdot \inprod{\xi_i}{\vstar} \leq 0 } \inprod{\xi_i}{\vstar}^2 \right) / \| v - \vstar \|^2 \\
& \qquad \qquad \leq C \etatil^2  \cdot \max \left\{ (1 + \sball^{-1} ) \frac{\max \paren{ d, 10 \log n }}{n}, (1 + \sball) \left( \etatil r \right)^{\sball} \cdot \log^{\sball + 1} \left( \frac{1}{\etatil r} \right) \right\}
\end{align*}
with probability exceeding $1 - c_1 n^{-8}$, where again, we adopt the convention that $0/0 = 0$.
\end{lemma}
As before, this lemma immediately establishes the claim in conjunction with a union bound over all $\binom{k}{2}$ pairs of parameters. \qed

\paragraph{Proof of claim~\eqref{clm:threec}:} This claim differs markedly from the previous claim, in that we require a uniform bound on singular values uniformly over all choices of  the true parameters $\paren{ \appsigstar }_{j = 1}^k \in \Bvol(\pimin, \Delta, \kappa)$. In contrast to the previous claim, we proceed by bounding the minimum singular values of \emph{all} sub-matrices of $\Xi$ of a certain size, and then show that each sub-matrix $\Xi^j$ encountered over the course of the algorithm has a certain size with high probability.

In the following lemma, we use the shorthand $\Csig \defn 4 \cs^2 \max\paren{ 9(\etatil)^2 \sball, 1}$.
\begin{lemma} \label{lem:minsingvol}
Suppose that Assumption~\ref{assptn:gen} holds, and that for a scalar $\alpha \in (0, 1)$, the sample size obeys the lower bound $n \geq \alpha^{-1} \max \paren{ 4d, \frac{d + 1}{\sball}}$. Then we have
\begin{align*}
\min_{S: |S| = \alpha n} \lambda_{\min} \left( \Xi_S^\top \Xi_S \right) \geq  \frac{1}{\Csig \log \Csig  + 2 C \sball^{-1} \log (e^2 / \alpha) } \left( \frac{\alpha}{e^2} \right)^{2 / \sball} \cdot \alpha n
\end{align*}
with probability greater than $1 - 3 \exp \left( -\alpha \numobs \right)$.
\end{lemma}

We combine this lemma with a lower bound on the size of each subset.
\begin{lemma} \label{lem:subsize}
Suppose that $n \geq 4\frac{kd}{\pimin^2} \log (n / kd)$. Then we have
\begin{align*}
\inf_{\appsigstar_1, \ldots, \appsigstar_k \in \Bvol(\pimin, \Delta, \kappa)} \; \; \inf_{ \substack{ \beta_1, \ldots, \beta_k \in \real^{d + 1}: \\ v_{j, j'} \in \Bspace_{\vstar_{j, j'} }(r_b) } }
\; \; \min_{j \in [k] } \; \; | S_j (\beta_1, \ldots, \beta_k) | \geq n \cdot \frac{\pimin}{4}
\end{align*}
with probability exceeding $1 - 2 \exp ( - c n \pimin^2 )$.
\end{lemma}

We are now ready to proceed to a proof of claim~\eqref{clm:threec}.
Note that the condition of the theorem guarantees that we have $n \geq \max \paren{ 4 \alpha^{-1} d, \alpha^{-1} \frac{d + 1}{\sball}, 4\frac{kd}{\pimin^2} \log (n / kd) }$ provided $\alpha \geq c \pimin^2$. 

Choosing $\alpha = \pimin / 4$ and conditioning on the intersection of the pair of events guaranteed by Lemmas~\ref{lem:minsingvol} and~\ref{lem:subsize}, we have
\begin{align*}
&\inf_{\appsigstar_1, \ldots, \appsigstar_k \in \Bvol(\pimin, \Delta, \kappa)} \; \; \inf_{ \substack{ \beta_1, \ldots, \beta_k \in \real^{d + 1}: \\ v_{j, j'} \in \Bspace_{\vstar_{j, j'} }(r_b) } }
\; \; \min_{j \in [k] } \; \; \eigmin \left( \Xi^j (\beta_1, \ldots, \beta_k)^\top \Xi^j (\beta_1, \ldots, \beta_k) \right) \\
&\qquad \geq \frac{1}{\Csig \log \Csig  + 2 C \sball^{-1} \log (4e^2 / \pimin) } \left( \frac{\pimin}{4e^2} \right)^{2 / \sball} \cdot \frac{\pimin n}{4}.
\end{align*}  
Combining this bound with the various high probability statements completes the proof of the claim. \qed

\noindent Having proved the claims, we turn to proofs of the three technical lemmas.

\subsubsection{Proof of Lemma~\ref{lem:lin-affine-c}}

Consider a fixed pair $(v, \vstar)$, and as before, let $\gamma_v = v - \vstar$; we have
\begin{align*}
\ind{ \inprod{\xi_i}{v} \cdot \inprod{\xi_i}{\vstar} \leq 0 } \inprod{\xi_i}{\vstar}^2 &\leq \ind{ \inprod{\xi_i}{v} \cdot \inprod{\xi_i}{\vstar} \leq 0 } \inprod{\xi_i}{\gamma_v}^2 \\
&\leq \ind{ \inprod{\xi_i}{\gamma_v}^2 \geq \inprod{\xi_i}{\vstar}^2 } \inprod{\xi_i}{\gamma_v}^2.
\end{align*}
Define the (random) set $K_v = \{i: \inprod{\xi_i}{\gamma}^2 \geq \inprod{\xi_i}{\vstar}^2  \}$, and note that this quantity implicitly depends on $\vstar$ as well, which is no longer fixed. We have the bound
\begin{align*}
\frac{1}{n}\sum_{i=1}^n \ind{ \inprod{\xi_i}{v} \cdot \inprod{\xi_i}{\vstar} \leq 0 } \inprod{\xi_i}{\vstar}^2  \leq \frac{1}{n} \| \Xi_{K_v} \gamma_v \|^2.
\end{align*}
We now show that the quantity $\| \Xi_K \gamma_v \|^2$ is bounded as desired for all pairs $(\vstar, v)$ such that $v \in \Bspace_{\vstar}(r)$. 
Recall that $\ustar$ is the ``linear" portion of $\vstar$, and
let 
\[
m = \max \paren{ 4d, (d + 1)/\sball, C (1 + \sball^{-1}) \log n,  n \left(C r \etatil \cdot \log \left(\frac{1}{r \etatil}\right) \right)^{\sball} },
\]
which implicitly depends on $r$.
We claim that for all $0 \leq r \leq 1/24$, we have
\begin{subequations}
\begin{align}
\Pr \left\{ \sup_{\substack{ \vstar \in \real^{d+1} \\ v \in \Bspace_{\vstar} (r) }   } |K_v| > m \right\} &\leq \frac{3}{1 - (3/4)^{\sball} } (3/4)^{ \max \paren{ d, 10 \log n } } + cne^{-c' n }, \text{ and } \label{clm:set1a-c}\\
\Pr \paren{ \bigcup_{\substack{T \subseteq [n]: \\ |T| \leq m}} \sup_{\omega \in \real^{d + 1}} \frac{\| \Xi_T \omega \|^2}{\| \omega \|^2} \geq  d + 4m \etatil^2 \log(n/m)  } &\leq \binom{n}{m}^{-1}. \label{clm:set2a-c}
\end{align}
\end{subequations}

Taking these claims as given, the proof of the lemma is immediate, since $\frac{n}{m} \leq \left( Cr \log (1/r \etatil) \right)^{-\sball}$, so that $\log (en / m) \leq C (1 + \sball ) \log(1 / r)$.

We now proceed to the proofs of the two claims. The proof of claim~\eqref{clm:set2a-c} follows from claim~\eqref{clm:set1a-c} similarly to before by using Lemma~\ref{lem:sgmax}, so we dedicate the rest of the proof to establishing claim~\eqref{clm:set1a-c}.

\paragraph{Proof of claim~\eqref{clm:set1a-c}:}
By definition of the set $K_v$, we have
\begin{align*}
\Pr \paren{\sup_{\substack{ \vstar \in \real^{d+1} \\ v \in \Bspace_{\vstar} (r) }   }  |K_v| > m } &\leq \sum_{ \substack{T \subseteq [n]: \\ |T| > m}} \Pr \left\{ \exists \vstar \in \real^{d+1}, v \in \Bspace_{\vstar}(r): \| \Xi_T \gamma_v \|^2 \geq \| \Xi_T \vstar\|^2 \right\} \\
&= \sum_{ \substack{T \subseteq [n]: \\ |T| > m}} \Pr \left\{ \exists \vstar \in \real^{d+1}, v \in \Bspace_{\vstar}(r) : \frac{\| \gamma_v \|^2}{ \| \ustar \|^2 } \frac{\| \Xi_T \gamma_v \|^2}{\| \gamma_v \|^2} \geq \frac{\| \Xi_T \vstar\|^2}{\| \ustar \|^2} \right\} \\
&\leq \sum_{ \substack{T \subseteq [n]: \\ |T| > m}} \Pr \left\{ \exists \vstar \in \real^{d+1}, v \in \Bspace_{\vstar}(r) : r^2 \frac{\| \Xi_T \gamma_v \|^2}{\| \gamma_v \|^2} \geq \frac{\| \Xi_T \vstar\|^2}{\| \ustar \|^2} \right\} \\
&\leq \sum_{\substack{T \subseteq [n]: \\ |T| > m}} \Bigg( \Pr \left\{ \exists \vstar \in \real^{d+1}, v \in \Bspace_{\vstar}(r) : \frac{\| \Xi_T \gamma_v \|^2}{\| \gamma_v \|^2} \geq |T| + \etatil^2 (\sqrt{d | T |} + d + |T| t_T) \right\} \\
&\qquad \qquad \quad + \Pr \left\{ \frac{\| \Xi_T \vstar\|^2}{\| \ustar \|^2} \leq r^2 \paren{|T| + \etatil^2 (\sqrt{d | T |} + d + |T| t_T)} \right\} \Bigg)
\end{align*}
where the final step follows from the union bound and holds for all $\{t_T\}_{T \subseteq n}$. For a \emph{fixed} subset $T$ of size $\ell \geq m$, we have the tail bounds
\begin{subequations}
\begin{align}
\Pr \{ \sup_{\omega \in \real^{d + 1}} \frac{\| \Xi_T \omega \|^2}{\| \omega \|^2 } \geq \ell + \etatil^2 (\sqrt{\ell d} + d + \ell t )  \} &\leq 2e^{-\ell \min \{ t, t^2 \} }, \text{ and } \label{eq:tb1-c}\\
\Pr \{ \inf_{\vstar \in \real^{d + 1}} \frac{\| \Xi_T \vstar\|^2}{\| \ustar \|^2 } \leq \delta \ell \} &\leq 3 \left( 4 \cs^2  \max\paren{ 9 (\etatil)^2 \sball, 1} \delta \log (1 / \delta) \right)^{\ell \sball/2}, \label{eq:tb2-c}
\end{align}
\end{subequations}
where inequality~\eqref{eq:tb1-c} follows from Lemma~\ref{lem:sgmax}, and inequality~\eqref{eq:tb2-c} from Lemma~\ref{lem:minsingsb} since we have $m \geq \max \paren{ \frac{d+ 1}{\sball}, 4d }$.

Now use the shorthand $\Csig \defn 4 \cs^2 \max\paren{ 9(\etatil)^2 \sball, 1}$ as before.
Substituting these bounds and letting $t_\ell \geq 1$ be a parameter to be chosen, we have
\begin{align*}
&\Pr \{ |K_v| > m \} \\
&\qquad \leq \sum_{\ell = m+1}^n  \binom{n}{\ell} \left[ 2e^{- \ell t_\ell } + 3 \left( \Csig r^2 \cdot \frac{\ell + \etatil^2 (\sqrt{\ell d} + d + \ell t_\ell )}{\ell} \log \left( \frac{\ell}{r^2 \ell + r^2 \etatil^2 (\sqrt{\ell d} + d + \ell t_\ell )} \right) \right)^{ \sball \ell/2 } \right].
\end{align*}
We now split the proof into two cases and choose $t_\ell$ differently for the two cases.

\paragraph{Case 1, $m \leq \ell < n/e$:}

 Substituting the choice $t_\ell = 4 \log (n / \ell) $, we obtain
\begin{align*}
\left( \frac{\ell + \etatil^2 (\sqrt{\ell d} + d + \ell t_\ell )}{\ell} \right)^{1/2} &\leq  \frac{\sqrt{\ell} + 5 \etatil \left( \sqrt{\ell \log (n / \ell)} + \sqrt{d} \right) }{\sqrt{\ell} } \\
&\stackrel{\1}\leq 1 + 6 \etatil \sqrt{\log (n / \ell)} \\
&\stackrel{\2}\leq 7 \etatil \cdot \sqrt{\log (n / \ell)}.
\end{align*}
where step $\1$ follows from the bound $m \geq d$, and step $\2$ from the bound $\etatil \geq 1$.
Putting together the pieces and noting that the map $x \mapsto x \log (1/x)$ is an increasing on the interval $(0, e^{-1})$, we have, for each $\ell$ in this set, the bound
\begin{align*}
\Pr \{ |K_v| = \ell \} &\leq 2 \left( \frac{en}{\ell} \right)^{-c\ell} + 3 \binom{n}{\ell} \left( 14 \Csig^{1/2} r \etatil \sqrt{\log (n / \ell) \cdot \log \frac{1}{7 r \etatil \sqrt{\log (n / \ell)}}} \right)^{\sball \ell}.
\end{align*}
For all $\ell \geq n (C r \etatil \log (1 / r \etatil) )^{\sball}$ for a sufficiently large constant $C$, the second term is bounded by $3\left( 3/4 \right)^{\sball \ell}$.

\paragraph{Case 2, $\ell \geq n/e$:}
The argument for this case is identical to before (with the choice $t_{\ell} = 2n / \ell$). Setting $\Csig'$ to be a sufficiently small $\eta$-dependent constant we obtain, for each $\ell$ in this range and for all $r \leq \Csig'$,
\begin{align*}
\left(3 \Csig r^2 \frac{\ell + \etatil^2 (\sqrt{\ell d} + d + \ell t_\ell )}{\ell} \right) \leq 1/3,
\end{align*}
where we also use the fact that $d \leq n/2$.
Once again, using the non-decreasing nature of the map $x \mapsto x \log (1/x)$ on the interval $(0, e^{-1})$, we have
\begin{align*}
&\binom{n}{\ell} \left[ 2e^{- \ell t_\ell } + 3 \left( \Csig r^2 \cdot \frac{\ell + \etatil^2 (\sqrt{\ell d} + d + \ell t_\ell )}{\ell} \log \left( \frac{\ell}{r^2 \ell + r^2 \etatil^2 (\sqrt{\ell d} + d + \ell t_\ell )} \right) \right)^{ \sball \ell/2 } \right] \\
&\qquad \qquad \qquad \leq 2 e^{-cn} + 3 \left( \sqrt{\frac{1}{3}} \right)^{\sball \ell}.
\end{align*}

\vspace{3mm}

Putting together the pieces from both cases, we have shown that for all $r \leq \Csig'$, we obtain
\begin{align*}
\Pr \{ |K_v| > m \} &\leq 2 n e^{-c n} + 3 \sum_{\ell = m+1}^{n} (3/4)^{\sball \ell} \\
&\leq 2 n e^{-c n} + n p^{10 \sball^{-1} \log n },
\end{align*}
where $p = (3/4)^{\sball}$. This completes the proof of the claim.
\qed

\subsubsection{Proof of Lemma~\ref{lem:minsingvol}}

The proof of this lemma follows from Lemma~\ref{lem:minsingsb} in conjunction with the union bound. In particular, we have
\begin{align*}
\Pr \paren{ \min_{S: |S| = \alpha n} \lambda_{\min} \left( \Xi_S^\top \Xi_S \right) \leq  \psi \alpha \cdot n } &\leq 3 \binom{n}{\alpha n} \left( \Csig \psi \log (1 / \psi) \right)^{\sball \alpha n} \\
&\leq 3 \left( e \frac{ \left( \Csig \psi \log (1 / \psi) \right)^{\sball} }{ \alpha } \right)^{\alpha n}.
\end{align*}

Finally, setting $\psi = \frac{1}{\Csig \log \Csig  + 2 C \sball^{-1} \log (e / \alpha) } \left( \frac{\alpha}{e^2} \right)^{2 / \sball}$ and performing some algebraic manipulation yields the claimed bound. 
\qed

\subsubsection{Proof of Lemma~\ref{lem:subsize}}

Note that we have $n \geq 4 \frac{kd}{\pimin^2} \log (n / kd)$. Noting that the $S_j$ can be thought of as the indicator vector corresponding to the intersection of $k$ halfspaces, applying Lemmas~\ref{lem:uniform-convergence} and~\ref{lem:sauer-shelah} in conjunction yields the bound
\begin{align*}
\inf_{\appsigstar_1, \ldots, \appsigstar_k \in \Bvol(\pimin, \Delta, \kappa)} |S_j (\appsigstar_1, \ldots, \appsigstar_k) | \geq n \cdot \frac{\pimin}{2}
\end{align*}
with probability exceeding $1 - ce^{-c' n \pimin^2}$. Furthermore, we have
\begin{align*}
|S_j (\beta_1, \ldots, \beta_k)| \geq |S_j (\appsigstar_1, \ldots, \appsigstar_k) | - | S_j (\appsigstar_1, \ldots, \appsigstar_k) \Delta S_j (\beta_1, \ldots, \beta_k)|,
\end{align*}
and using the notation for a wedge $W$ from before and the notation for $n_W$ from equation~\eqref{eq:nw}, we have
\begin{align*}
| S_j (\appsigstar_1, \ldots, \appsigstar_k) \Delta S_j (\beta_1, \ldots, \beta_k)| &\stackrel{\1}{\leq} \sum_{j' = 1}^k n_{W(v_{j, j'}, \vstar_{j, j'}) } \\
&\stackrel{\2}{\leq} 2 \sum_{j' = 1}^k n \vol( W(v_{j, j'}, \vstar_{j, j'}) ),
\end{align*}
where step $\1$ follows from the bound~\eqref{eq:symmdiff}, and step $\2$ follows once again from Lemma~\ref{lem:uniform-convergence} since $n$ is large enough. The volume of a wedge under a distribution satisfying Assumption~\ref{assptn:gen} is upper bounded in Appendix~\ref{app:volbd}. Applying Lemma~\ref{lem:vol_gen}, we have
\begin{align*}
|S_j (\beta_1, \ldots, \beta_k)| &\geq n \cdot \frac{\pimin}{2} - \sum_{j' = 1}^k \left( \Csig'' \left( \frac{ \eucnorm{v_{j, j'} - \vstar_{j, j'} } }{ \eucnorm{ \sigstar_j - \sigstar_{j'} } } \right)^{2} \log \left( \frac{ \eucnorm{ \sigstar_j - \sigstar_{j'} }}{ \eucnorm{v_{j, j'} - \vstar_{j, j'} } } \right) \right)^{\sball} \\
&\stackrel{\3}{\geq} n \cdot \frac{\pimin}{2} - k \cdot n\frac{\pimin}{4k} \\
&\geq \frac{\pimin n}{4},
\end{align*}
where in step $\3$, we have used condition~\eqref{eq:init-thm2-ii} satisfied by all the parameters $v_{j, j'}$, by which we have
\begin{align*}
\frac{\left\| v_{j, j'} - \vstar_{j, j'} \right\|}{\| \sigstar_j -\sigstar_{j'} \| } \log \left( \frac{\| \sigstar_j -\sigstar_{j'} \| }{\left\| v_{j, j'} - \vstar_{j, j'} \right\|} \right) &\leq \Csig^{(1)} \left(\frac{\pi_{\min}^{1 + 2 \sball^{-1}}}{k}\right)^{\sball^{-1}},
\end{align*}
which further implies, for a sufficiently small constant $\Csig^{(1)}$, that
\begin{align*}
\frac{\left\| v_{j, j'} - \vstar_{j, j'} \right\|}{\| \sigstar_j -\sigstar_{j'} \| } \log^{1/2} \left( \frac{\| \sigstar_j -\sigstar_{j'} \| }{\left\| v_{j, j'} - \vstar_{j, j'} \right\|} \right) &\leq \left( \frac{\pimin}{4k} \right)^{\sball^{-1}/2}.
\end{align*}
Since these steps held for an arbitrary index $j$, the proof of the lemma is complete.
\qed


\section{Proof of Corollary~\ref{cor:PR}} \label{sec:pf-cor1}

Since the proof is more or less subsumed by Theorem~\ref{thm:sbgen}, we only sketch the details. Let $\theta$ denote the current iterate and $\theta^+$ denote the next iterate. Proceeding as before, we have the deterministic bound
\begin{align*}
\| \Xmat (\theta^+ - \thetastar) \|^2 \lesssim \sum_{i = 1}^n \ind{ \inprod{\meas_i}{\theta} \inprod{\meas_i}{\thetastar} \leq 0 } \inprod{\meas_i}{\thetastar}^2 + \| P_X \epsilon \|^2,
\end{align*}
and so this case is even simpler than the max-affine setting, since we no longer select specific sub-matrices of $\Xmat$ on which to invert a linear system. Standard bounds on sub-Gaussian random matrices yield
\begin{align} \label{eq:sgtb}
\Pr \paren{ \opnorm{ \Xmat^\top \Xmat - n } \geq \eta^2 (d +  4 \sqrt{\log n}) } \leq 2 n^{-8}.
\end{align}
Assuming that $n / \eta^2 \geq 2 (d + 4 \sqrt{\log n})$, we now apply Lemmas~\ref{lem:lin-affine-c} and~\ref{lem:noise} along with the bound~\eqref{eq:sgtb}
to obtain that simultaneously for all pairs $(\thetastar, \theta)$ satisfying $\| \theta - \thetastar \| / \| \thetastar \| \leq r \leq 1/24$
, there are universal constants $(c_1, c_2)$ such that we have the one-step bound
\begin{align*}
\| \theta^+ - \thetastar \|^2 &\leq c_1 \| \theta - \thetastar \|^2 \paren{ \eta^2  \cdot \max \left\{ (1 + \sball^{-1} ) \frac{\max \paren{ d, 10 \log n }}{n}, (1 + \sball) \left( \eta r \right)^{\sball} \cdot \log^{\sball + 1} \left( \frac{1}{\eta r} \right) \right\} } \\
&\qquad \qquad + c_1 \sigma^2 \frac{d}{n} \log \left( \frac{n}{d} \right),
\end{align*}
with probability exceeding $1 - c_1 n^{-7}$. Choosing a small enough scalar $r$ (depending on the tuple $(\eta, \sball, \cs)$) and a large enough $n$ to make the quantity within the braces less than $3/4$ completes the proof, since the the above bound can be recursively applied when we also have the condition $n \geq \Csig^{(1)} \frac{\sigma^2}{\| \thetastar \|^2} d \log (n / d)$.

\section{Proof of Theorem~\ref{thm:pca}} \label{sec:pf-thm3}

We dedicate the first portion of the proof to a precise definition of the quantity $\gamma$.

Let $\Thetastar \in \real^{k \times d}$ denote
a matrix with rows $(\sigstar_j)^T, j = 1, \dots, k$ and let $\Sigma =
\Thetastar (\Thetastar)^\top \in \real^{k 
  \times k}$. We employ the decomposition $\Thetastar = \Astar (\Ustar)^\top$, where $\Astar \in \R^{k \times k}$ is the invertible matrix of coefficients and $\Ustar \in \R^{\numdim \times k}$ is a matrix of orthonormal columns.
  Note that for $X \sim N(0, I_d)$, the vector in $\R^k$ with $j$-th component
 $\inprod{X}{\theta_j^*} + b_j^*$ is distributed as $Z + b^*$ where $Z \sim N(0, \Sigma)$ and the vector $\bstar \in \real^k$ collects the scalars $\paren{\bstar_j}_{j = 1}^k$ in its entries. For $Z
\sim \NORMAL(0, \Sigma)$, let   
\begin{align}
\rho = \frac{\E \left[\max(Z + \bstar) Z^\top \Sigma^{-1} \ones  \right]}{\sqrt{\E \left[ ( \max(Z + \bstar ) )^2 \right] \cdot \E \left[ ( Z^\top \Sigma^{-1} \ones)^2 \right] } } \label{eq:rho}
\end{align}
denote the correlation coefficient between the maximum and a particular linear combination of a multivariate Gaussian distribution. Variants of such quantities have been studied extensively in the statistical literature (see, e.g., James~\cite{james2007limit}). For our purposes, the fact that $\max(Z + \bstar) Z \neq 0$ for any finite $\bstar$, coupled with a full-rank $\Sigma$, ensure that $\rho \neq 0$ for any fixed $k$. Also define the positive scalar 
$
\varrho \defn \sqrt{ \EE [ (\max( Z + \bstar) )^2 ]},
$
which tracks the average size of our observations.
Also recall the quantity $\varsigma$ defined in the main section.

For each $j \in [k]$ consider the zero-mean Gaussian random vector with covariance $(\ones \cdot e_j^\top - I) \Astar (\Astar)^\top (\ones \cdot e_j^\top - I)^\top$. This is effectively a Gaussian that lives in $k - 1$ dimensions, with density that we denote by $\widetilde{\psi}_j (x_1, x_2, \ldots, x_{j - 1}, 0, x_{j + 1}, \ldots x_k)$ at point $(x_1, x_2, \ldots, x_{j - 1}, 0, x_{j + 1}, \ldots x_k)$ (the density is not defined elsewhere). Truncate this random vector to the region $\{ x_i \geq \bstar_i - \bstar_j : i \in [k] \}$; this results in the \emph{truncated} Gaussian density $\psi_j (x_1, x_2, \ldots, x_{j - 1}, 0, x_{j + 1}, \ldots x_k)$ for each $j \in [k]$.

For any $x \in \real^k$ such that $x_j = 0$, define
\begin{align} \label{eq:marginal-density}
F^j_i (x) = \int_{\bstar_1 - \bstar_j}^\infty \cdots \int_{\bstar_{i-1} - \bstar_j}^\infty \int_{\bstar_{i+1} - \bstar_j}^\infty \cdots \int_{\bstar_k - \bstar_j}^\infty \psi_j (x_1, \ldots, x_{i-1}, x, x_{i+1}, \ldots, x_k) dx_{k} \ldots dx_{i+1} dx_{i-1} \ldots dx_1
\end{align}
to be the $i$-th marginal density of this truncated Gaussian evaluated at the point $x$, with the convention that $F^j_j(\cdot) = 0$ everywhere. Also define the vector $F^j$ by setting its $i$-th entry to $(F^j)_i = F^j_i( \bstar_i - \bstar_j)$.

Now let $P$ denote the matrix with entries
\begin{align*}
P_{i, j} = 
\begin{cases}
(F^j)_i / \sum_{k \neq j} (F^j)_k &\text{ if } i \neq j \\
0 	   &\text{ otherwise}.
\end{cases}
\end{align*}
Note that the matrix $P$ is the transition matrix of an irreducible, aperiodic Markov chain, with one eigenvalue equal to $1$. Consequently, the matrix $I - P$ is rank $k - 1$. 
With this setup in place, let
\begin{align} \label{eq:gammadef-1}
\gamma \defn \min \bigg \lbrace \rho^2 \varrho^2, \min_{j \in k} \left( \sum_{k \neq j} (F^j)_k \right) \lambda_{k}( \Sigma ) \cdot \sqrt{ \lambda_{k-1} \left( (I - P^\top)(I - P) \right) } \bigg \rbrace
\end{align}
denote a positive scalar that will serve as a bound on our eigengap.
%


Let $M_1 = \EE\left[ \max(\Theta^* X + \bstar) X \right]$ and $M_2 = \EE\left[ \max(\Theta^* X + \bstar ) (XX^\top - I_d )\right]$ denote the expectations of the first and second moment estimators, respectively. 

For a random variable $W \sim \NORMAL(\bstar, \Sigma)$, we often use the shorthand
\begin{align*}
\{W_j = \max \} \defn \{ W_j \geq W_i \text{ for all } 1 \leq i \leq k \}.
\end{align*}
%
Finally, collect the probabilities $\{\pi_j\}_{j = 1}^k$ defined in equation~\eqref{eq:pi_def} in a vector $\pi \in \real^k$. We use $\ones$ to denote the all-ones vector in $k$ dimensions.


We are ready to state our two main lemmas.
\begin{lemma} \label{lem:expmom}
(a) The first moment satisfies
\begin{align*}
M_1 = (\Theta^*)^\top \pi \quad \text{ and } \quad \inprod{M_1}{(\Theta^*)^\top \Sigma^{-1} \ones} = \rho \varrho \eucnorm{(\Theta^*)^\top \Sigma^{-1} \ones}.
\end{align*}
(b) The second moment satisfies
\begin{align*}
 M_2 \succeq 0, \quad M_2 (\Theta^*)^\top \Sigma^{-1} \ones = 0, \quad \rank(M_2) = k - 1 \quad \text{ and }
 \end{align*}
 \begin{align*}
  \lambda_{k-1}(M_2) \geq \min_{j \in k} \left( \sum_{k \neq j} (F^j)_k \right) \lambda_{k}( \Sigma ) \cdot \sqrt{ \lambda_{k-1} \left( (I - P^\top)(I - P) \right) }.
\end{align*}
\end{lemma}

We combine this lemma with a result that shows that the empirical moments concentrate about their expectations. 
\begin{lemma} \label{lem:dev}
For an absolute constant $C$, we have
\begin{subequations}
\begin{align}
\Pr \paren{ \eucnorm{ \Mhat_1 - M_1}^2 \geq C_1  \left( \sigma^2 + \varsigma^2  \right)\frac{d \log^2 (nk)}{n} } &\leq 5dn^{-12}, \text{ and} \label{eq:mom1}\\
\Pr \paren{ \opnorm{ \Mhat_2 - M_2}^2 \geq C\left (\sigma^2 + \varsigma^2  \right ) \frac{d \log^3 (nk) }{n} } &\leq 5dn^{-12}. \label{eq:mom2}
\end{align}
\end{subequations}
\end{lemma}

Lemma~\ref{lem:expmom} is proved at the end of this section, and Lemma~\ref{lem:dev} is proved in Appendix~\ref{app:pf-moments}.
For now, we take both lemmas as given and proceed to a proof of Theorem~\ref{thm:pca}.

Recall the matrix $\Mhat = \Mhat_1 \otimes \Mhat_1 + \Mhat_2$ and let $M = M_1 \otimes M_1 + M_2$. By Lemma~\ref{lem:expmom}, the matrix $M$ is positive semidefinite with $k$ non-zero eigenvalues. 
In particular, using the shorthand $\bar{\theta} \defn (\Thetastar)^\top \Sigma^{-1} \ones$, we have $\bar{\theta} \in {\sf nullspace}(M_2)$, and so
\begin{align*}
\bar{\theta}^\top M \bar{\theta} = \inprod{\bar{\theta}}{M_1}^2 = \rho^2 \varrho^2 \| \bar{\theta} \|^2,
\end{align*}
where the final inequality follows by part (a) of Lemma~\ref{lem:expmom}.

Thus, there is a $k$-dimensional subspace orthogonal to the nullspace of $M$ (and so the range of $M$ is $k$ dimensional). For any unit vector $v$ in this subspace, we have
\begin{align*}
v^\top M v \geq \min \{\rho^2 \varrho^2, \lambda_{k -1}(M_2) \}.
\end{align*}
Thus, the $k$th eigenvalue of $M$ satisfies 
\[
\lambda_k(M) \geq \min \bigg \lbrace \rho^2 \varrho^2, \min_{j \in k} \left( \sum_{k \neq j} (F^j)_k \right) \lambda_{k}( \Sigma ) \cdot \sqrt{ \lambda_{k-1} \left( (I - P^\top)(I - P) \right) } \bigg \rbrace = \gamma, 
\]
where the equality follows by definition~\eqref{eq:gammadef-1}. By Lemma~\ref{lem:dev}, we have 
\begin{align*}
\opnorm{ \Mhat - M}^2 &\leq 2 \opnorm{ \Mhat_2 - M_2}^2 + 2 \opnorm{ \Mhat_1 \otimes \Mhat_1 - M_1 M_1^\top}^2 \\
&\leq 2 C\left (\sigma^2 + \varsigma^2 \log^2(nk) \right ) \frac{d \log (nk) }{n}  + 8 \eucnorm{ \Mhat_1 - M_1}^2 \eucnorm{M_1}^2 + 2 \eucnorm{ \Mhat_1 - M_1}^4 \\
&\leq C' \left (\sigma^2 + \varsigma^2 \log^2(nk) \right ) \frac{d \log (nk) }{n} ,
\end{align*}
where the last two inequalities each hold with probability greater than $1 - 2n^{-10}$.

We denote the estimated and true eigenspaces by $\Uhat$ and $\Ustar$, respectively. Applying~\cite[Theorem 2]{yu2014useful} yields the bound
\begin{align*}
\fronorm{\Ustar (\Ustar)^\top - \Uhat \Uhat^\top}^2 \leq C \left (\frac{\sigma^2 + \varsigma^2 }{\gamma^2} \right ) \frac{k d \log^3 (nk) }{n},
\end{align*}
thereby proving the required result.
\qed

\noindent We now proceed to a proof of Lemma~\ref{lem:expmom}.

\subsection{Proof of Lemma~\ref{lem:expmom}}

Recall our decomposition $\Theta^* = \Astar (\Ustar)^\top$, where $\Ustar \in \real^{\numdim \times k}$ is a matrix of orthonormal columns, and $\Astar \in \real^{k \times k}$ is an invertible matrix of coefficients. Since we are always concerned with random variables of the form $\Theta^* X$ with $X$ Gaussian, we may assume without loss of generality by the rotation invariance of the Gaussian distribution that $\Ustar = [e^d_1 \; e^d_2 \ldots \; e^d_k]$, where $e^d_i$ denotes the $i$th standard basis vector in $\real^d$.

We let $X_{i}^j = (X_i, X_{i + 1}, \ldots, X_j)$ denote a sub-vector of the random vector $X$, so that by the above argument, we have $\Theta^* X \stackrel{d}{=} \Astar X_1^k$.

\paragraph{Calculating $M_1$:} Using the shorthand $Z = \Astar X_1^k$, we have
\begin{align*}
M_1 &= \E [ \max(\Theta^* X + \bstar) X] \\
&= \Ustar \E [ \max(\Astar X_1^k + \bstar) X] \\
&= \Ustar (\Astar)^{-1} \E [\max(Z + \bstar) Z].
\end{align*}
Now using Stein's lemma\footnote{One can also derive $M_1 = (\Theta^*)^{\top} \pi$
  directly applying Stein's lemma $\E X f(X) = \E \nabla f(X)$ to
  $f(x) := \max(\Theta^* X + b^*)$ so that $\nabla f(x)$ equals $\theta_j^*$
whenever $x$ belongs to the region when $j$ is maximized.}, by a calculation similar to the one performed also in Seigel~\cite{siegel1993surprising} and Liu~\cite{liu1994siegel}, we have
\begin{align*}
\E [\max(Z + \bstar) Z] = \Sigma \pi,
\end{align*}
where $\pi \in \real^k$ is the vector of probabilities, the $j$-th of which is given by equation~\eqref{eq:pi_def}, and we have used $\Sigma = \Astar (\Astar)^\top = (\Theta^*) (\Theta^*)^\top$ to denote the covariance matrix of $Z$.

Therefore, we have the first moment
\begin{align*}
M_1 &= \Ustar (\Astar)^{-1} \Astar (\Astar)^\top \pi = (\Theta^*)^\top \pi. 
\end{align*}

\paragraph{Correlation bound:} By computation, we have
\begin{align*}
\inprod{M_1}{(\Theta^*)^\top \Sigma^{-1} \ones} = \E \left[ \max(Z + \bstar) \inprod{Z}{\Sigma^{-1} \ones} \right] &\stackrel{\1}{=} \rho \cdot \sqrt{ \E  \left[ (\max(Z + \bstar))^2 \right] \cdot \E \left[ \inprod{Z}{ \Sigma^{-1} \ones}^2 \right] } \\
&\stackrel{\2}{=} \rho \varrho \cdot \eucnorm{ (\Theta^*)^\top \Sigma^{-1} \ones}, 
\end{align*}
where step $\1$ follows from the definition~\eqref{eq:rho} of the quantity $\rho$, and step $\2$ from explicitly calculating the expectation and recalling the definition of $\varrho$.

\paragraph{Positive semidefiniteness of $M_2$:} For some $u \in \real^{\numdim}$, let $f(X) = \max(\Theta^* X + b)$ and $g_u(X) = \inprod{u}{X}^2$. Since $g_u$ is an even function, we have $\E [g_u(X) X] = 0$. Furthermore, since both $f$ and $g_u$ are convex, applying Lemma~\ref{lem:correlation} (see Appendix~\ref{app:thm3}) yields the bound
\begin{align*}
\E [f(X) g_u(X)] \geq \E[f(X)] \E[g_u(X)],
\end{align*}
so that substituting yields the bound
\begin{align*}
u^\top \E [\max(\Theta^* X + b) XX^\top ] u \geq u^\top \E[ \max(\Theta^* X + b) I] u.
\end{align*}
Since this holds for all $u \in \real^d$, we have shown that the matrix $\E[ \max(\Theta^* X + b) (XX^\top - I)]$ is positive semidefinite.

\paragraph{Calculating $M_2$:} We now use Stein's lemma to compute an explicit expression for the moment $M_2$. By the preceding substitution, we have
\begin{align*}
M_2 &= \E \left[ \max(\Astar X_1^k + \bstar)
\begin{bmatrix}
X_1^k (X_1^k)^\top - I_k & X_1^k (X_{k+1}^d)^\top \\
\vspace{2mm}
X_{k+1}^d (X_1^k)^\top & X_{k+1}^d (X_{k+1}^d)^\top - I_{d - k}
\end{bmatrix}
\right] \\
&=
\begin{bmatrix}
\E \left[ \max(\Astar X_1^k + \bstar) (X_1^k (X_1^k)^\top - I_k) \right] & 0 \\
0 & 0
\end{bmatrix}
\end{align*}
Once again using the substitution $Z = \Astar X_1^k$ and $\Sigma = \Astar (\Astar)^\top$, we have
\begin{align*}
M_2 = \Ustar (\Astar)^{-1} \E \left[ \max(Z + \bstar) (Z Z^\top - \Sigma) \right] (\Astar)^{-\top} (\Ustar)^\top,
\end{align*}
and applying Stein's lemma yields
\begin{align*}
\E \left[ \max(Z + \bstar) (Z Z^\top - \Sigma) \right] = \Sigma \Pi^\top = \Pi \Sigma,
\end{align*}
where $\Pi \in \real^{k \times k}$ denotes a matrix with entry $i, j$ given by $\Pi_{i, j} = \E [Z_i \ind{ Z_j + \bstar_j = \max} ]$, and the final equality follows by symmetry of the matrix.

Simplifying further, we have
\begin{align*}
M_2 = \Ustar (\Astar)^{-1} \Pi \Astar (\Ustar)^\top.
\end{align*}


\paragraph{Nullspace of $M_2$:} Notice that $\Pi \ones = \E [Z] = 0$, so that 
\begin{align*}
M_2 (\Theta^*)^\top \Sigma^{-1} \ones = \Ustar (\Astar)^{-1} \Pi \Astar (\Ustar)^\top \Ustar (\Astar)^{\top} \Sigma^{-1} \ones = 0.
\end{align*}

\paragraph{Rank of $M_2$ and bound on $\lambda_{k-1}(M_2)$:} By the previous claim, we have $\rank(M_2) \leq k-1$. Furthermore, the matrix $M_2$ has $d - k$ eigenvalues equal to zero, and the other $k$ of its eigenvalues equal to those of $\Pi$, all of which are positive (by the PSD property of $M_2$), and at least one of which is zero. Thus, it suffices to work with the eigenvalues of $\Pi$; in particular, a lower bound on $\lambda_{k - 1}(\Pi)$ directly implies a lower bound on $\lambda_{k - 1}(M_2)$.

Let us first show that $\lambda_{k - 1}(\Pi) > 0$. Since we know that a zero-eigenvector of $\Pi$ is the all-ones vector $\ones$, it suffices to show that $x^\top \Pi x \neq 0$ when $\inprod{x}{\ones} = 0$. We use the shorthand $x \perp \ones$ to denote any such vector. 

We now explicitly evaluate the entries of the matrix $\Pi$. We denote the $j$th column of this matrix by $\Pi_j$.
We have
\begin{align}
\Pi_j &= \E[ Z \ind{Z_j + \bstar_j = \max}] \notag \\
&= \E[ \ones \cdot Z_j \ind{Z_j + \bstar_j = \max}] - \E [ (\ones \cdot Z_j - Z) \ind{Z_j + b_j = \max} ] \notag \\
&= \ones \cdot \E[ Z_j \ind{Z_j + \bstar_j = \max}] - \E [ (\ones \cdot Z_j - Z) \ind{Z_j + \bstar_j = \max} ]. \label{eq:suffexp}
\end{align} 
For any $x \perp \ones$, we have $x^\top \ones \E [Z \ind{Z + \bstar = \max} ]^\top \ones = 0$, so that in order to show that $x^\top \Pi x \neq 0$, it suffices to consider just the second term in the expression~\eqref{eq:suffexp}.

In order to focus on this term, consider the matrix $\Phi$ with column $j$ given by
\begin{align*}
\Phi_j = \E [ (\ones \cdot Z_j - Z) \ind{Z_j - Z \geq \bstar - \bstar_j } ].
\end{align*}
where the indicator random variable above is computed element-wise.
We are interested in evaluating the eigenvalues of the matrix $-\Phi$.

The quantity $\Phi_j$ can be viewed as the first moment of a (lower) truncated, multivariate Gaussian with (original) covariance matrix 
\begin{align*}
\kappa_j = (\ones \cdot e_j^\top - I) \Astar (\Astar)^\top (\ones \cdot e_j^\top - I)^\top.
\end{align*} 
Recalling the column vectors $F^j$ defined (in equation~\eqref{eq:marginal-density}) for each $j \in [k]$ and applying~\cite[(11)]{manjunath2009moments} (see also Tallis~\cite{tallis1961moment} for a similar classical result), we may explicitly evaluate the vector $\Phi_j$, as
\begin{align*}
\Phi_j &= \kappa_j F^j \\
&\stackrel{\3}{=} (\ones \cdot e_j^\top - I) \Astar (\Astar)^\top G_j
\end{align*}
where in step $\3$, we have let $G_j$ denote a vector in $\real^k$ with entry $i$ given by 
\begin{align*}
(G_j)_i = 
\begin{cases}
- (F^j)_i \text{ if } j \neq i \\
\sum_{k \neq j} (F^j)_k \text{ otherwise}.
\end{cases}
\end{align*}
Letting $G \in \real^{k \times k}$ denote the matrix with $G_j$ as its $j$th column, and for $x \perp \ones$, we have
\begin{align*}
x^\top (- \Phi) x = x^\top \Sigma G x,
\end{align*}
since once again, for each $x \perp \ones$, we have $x^\top \ones \cdot e_j^\top \Astar (\Astar)^\top (\ones \cdot e_j^\top - I)^\top x = 0$. 

Now consider the matrix $\Sigma G$. In order to show the claimed bound, it suffices to show that $x^\top \Sigma G x \neq 0$ if $x \perp \ones$. We show this by combining two claims:

\noindent {\bf Claim 1:} The nullspace of $G$ is one-dimensional. \\
{\bf Claim 2:} Both the left and right eigenvectors of $\Sigma G$ that correspond to this nullspace are not orthogonal to the $\ones$ vector.

We show both claims concurrently.
The nullspace of $G$ is clearly non-trivial, since $\ones^\top G = 0$. Let us first show, by contradiction, that the left eigenvector corresponding to this nullspace dimension is not orthogonal to the all-ones vector. Toward that $x_{\ell}$ denote the aforementioned left eigenvector which also satisfies $\inprod{x_{\ell}}{\ones} = 0$.
By virtue of being a left eigenvector, $x_{\ell}$ satisfies $\Sigma x_{\ell} = \ones$, or in other words, $x_{\ell} = \Sigma^{-1} \ones$.  Since $x_{\ell} \perp \ones$, we have $\ones^\top \Sigma \ones = 0$, but this contradicts the positive definiteness of $\Sigma$.

It remains to establish that the the null-space of $G$ is in fact only one-dimensional, and that its right eigenvector is not orthogonal to the all-ones vector.
Notice that we may write the matrix as
\begin{align*}
G = (I - P^\top) \diag(G),
\end{align*}
where we have let $P$ denote the matrix with entries
\begin{align*}
P_{i, j} = 
\begin{cases}
(F^j)_i / \sum_{k \neq j} (F^j)_k &\text{ if } i \neq j \\
0 	   &\text{ otherwise}.
\end{cases}
\end{align*}
Since all of the entries of $P$ are positive and sum to $1$ along the rows, the matrix $P$ matrix can be viewed as the transition matrix of a Markov chain. Furthermore, since this Markov chain communicates, it is irreducible and aperiodic, with only one eigenvalue equal to $1$. Thus, the matrix $I - P^\top$ is rank $k - 1$, thereby establishing that the nullspace of $G$ is one-dimensional. Furthermore, the right eigenvector $x_r$ of $G$ is a non-negative vector by the Perron-Frobenius theorem, so that it cannot satisfy $\inprod{x_r}{\ones} = 0$.

We have thus established both claims, which together show that
$\lambda_{k - 1}(M_2) \neq 0$. Further noting that the matrix $M_2$ is
positive definite, we have 
\begin{align*}
\lambda_{k - 1} (M_2) \geq \min_{j \in [k]} G_{j, j} \cdot \lambda_{\min} (\Sigma) \sqrt{ \lambda_{k-1} [(I - P^\top) (I - P)]},
\end{align*}
and this completes the proof of the claim, and consequently, the
lemma. 
\qed

\section{Proof of Theorem~\ref{thm:rs}} \label{sec:pf-thm4}

Recall the matrix $\Vhat$ formed by appending a standard basis vector to $\Uhat$. First, we show that there is a point among the randomly chosen initializations that is sufficiently close to 
the true parameters. Toward that end, let $c_0 \defn r + \Deltamax$ and define 
$\beta^{\ell}_j = \Vhat \nu^{\ell}_j$ for each $j \in [k]$ and $\ell \in [M]$. Let
\begin{align*}
\ellsharp \defn \argmin_{\ell \in [M]} \paren{ \max_{j \in [k]} \| c_0 \beta^{\ell}_j - \appsigstar_j \| },
\end{align*}
and define the event
\begin{align*}
\mathcal{E}_1 (M, r) \defn \paren{ \max_{j \in [k]} \| c_0 \beta^{\ellsharp}_j - \appsigstar_j \| \geq r + \Deltamax \opnorm{ \Uhat \Uhat^\top - U^* (U^*)^\top } };
\end{align*}
in words $\Espace_1(M, r)$ is the event that \emph{none} of the randomly initialized points (when scaled by a fixed constant $c_0$) is close to the true parameters. The following lemma bounds the probability of such an event provided $M$ is sufficiently large.

\begin{lemma} \label{lem:closept}
If $M \geq \left( 1 + \frac{\Deltamax}{r} \right)^{k^2} \log (1 / \delta)$, then $\Pr \paren{ \Espace_1 (M, r) } \leq \delta$.
\end{lemma}

Taking the lemma as given, let us now proceed to the proof of the theorem.
Define the shorthand
\[
\Pspace(\beta_1, \ldots, \beta_k) \defn \frac{2}{n} \sum_{i = n/2 + 1}^n \left( \max_{j \in [k]} \; \inprod{\appmeas_i}{\beta_j} - \max_{j \in [k]} \; \inprod{\appmeas_i}{\appsigstar_j} \right)^2
\]
for each set of parameters $\beta_1, \ldots, \beta_k \in \real^{d + 1}$.

For each $\ell \in [M]$, let 
\[
c_{\ell} \defn \argmin_{c \geq 0} \frac{2}{n} \sum_{i = n/2 + 1}^n \left( y_i - c \max_{j \in [k]} \; \inprod{\appmeas_i}{\beta^{\ell}_j} \right),
\]
and recall that $\ell^*$ is the index returned by the algorithm. Also note that trivially, we have $c_{\ell^*} > 0$ with probability tending to $1$ exponentially in $n$, so that this pathological case in which
the initial partition is random can be ignored.

Due to sample splitting, the parameters $\beta^\ell_j$ are independent of the noise sequence $\{ \epsilon_i \}_{i = n/2 + 1}^n$. Thus, applying Lemma~\ref{lem:rays} from Appendix~\ref{sec:rays} yields the bound
\begin{align*}
\Pr\paren{ \Pspace(c_{\ell^*} \beta^{\ell^*}_1, \ldots, c_{\ell^*} \beta^{\ell^*}_k) \geq c_1 \paren{ \min_{\substack{c \geq 0 \\ \ell \in [M] } } \Pspace(c \beta^{\ell}_1, \ldots, c \beta^{\ell}_k) + \frac{\sigma^2 t ( \sqrt{\log M} + c_1 )}{n} }} \leq e^{-c_2 n t(\sqrt{\log M}+ c_1)},
\end{align*}
valid for all $t \geq \sqrt{\log M} + c_1$ and suitable universal constants $c_1$ and $c_2$.
Setting $t = \sqrt{\log M} + c_1$ and on this event, we have
\begin{align*}
\Pspace(c_{\ell^*} \beta^{\ell^*}_1, \ldots, c_{\ell^*} \beta^{\ell^*}_k) \leq c_1 \Pspace(c_0 \beta^{\ellsharp}_1, \ldots, c_0 \beta^{\ellsharp}_k) + c_1 \frac{\sigma^2 \log M}{n} 
\end{align*}
with probability greater than $1 - e^{-c_2 n}$.

To complete the proof, let $C(\pimin, k) \defn c_2 \log(k/\pi_{\min}) \left( \frac{k}{\pimin} \right)^5$ for a suitable constant $c_2$ and apply Lemma~\ref{lem:pred_est_gen} twice (note that here we use the assumption $n \geq Cd \frac{k}{\pimin}$) to obtain
\begin{align*}
\sum_{j \in [k]} \min_{j' \in [k]} \| \appsigstar_j - c_{\ell^*} \beta^{\ell^*}_{j'} \|^2 &\leq C(\pimin, k) \cdot \Pspace(c_{\ell^*} \beta^{\ell^*}_1, \ldots, \beta^{\ell^*}_k) \\
&\leq C(\pimin, k) \cdot \paren{ \Pspace(c_0 \beta^{\ellsharp}_1, \ldots, c_0 \beta^{\ellsharp}_k) + \frac{\sigma^2 \log M }{n} } \\
&\leq C(\pimin, k) \cdot \paren{ 2 \sum_{j = 1}^k \| c_0 \beta^{\ellsharp}_j - \appsigstar_j \|^2 + \frac{\sigma^2 \log M }{n} } \\
&\leq C(\pimin, k) \cdot \paren{ 2k \max_{j \in [k]} \| c_0 \beta^{\ellsharp}_j - \appsigstar_j \|^2 + \frac{\sigma^2 \log M }{n} }\\
&\stackrel{\2}{\leq} C(\pimin, k) \paren{ 4k \left( r^2 + \Deltamax^2 \opnorm{ \Uhat \Uhat^\top - U^* (U^*)^\top}^2 \right) + \frac{\sigma^2 \log M }{n} }
\end{align*}
on an event of suitably high probability, where step $\2$ follows from Lemma~\ref{lem:closept} and on the event $\Espace^c_1 (M, r)$.

Finally, note that provided the RHS above is less than $\Delta^2 / 4$, each minimum on the LHS is attained for a unique index $j'$. This condition is ensured by the sample size assumption of the theorem; thus, we have
\begin{align*}
\min_{c > 0} \; \dist\left( \paren{ c \appsig{0}_j }_{j = 1}^k, \paren{ \appsigstar_j}_{j = 1}^k \right) \leq C(\pimin, k) \paren{ 4k \left( r^2 + \Deltamax^2 \opnorm{ \Uhat \Uhat^\top - U^* (U^*)^\top }^2 \right) + \frac{\sigma^2 \log M }{n} }.
\end{align*}
Combining the various probability bounds then completes the proof.
\qed

\subsection{Proof of Lemma~\ref{lem:closept}}

Recall that $U^*$ is a matrix of orthonormal columns spanning the $k$-dimensional subspace spanned by the vectors $\{\sigstar_1, \ldots, \sigstar_k \}$. Define the matrix 
\begin{align*}
V^* = 
\begin{bmatrix}
U^* & 0 \\
0 & 1
\end{bmatrix};
\end{align*}
for each $j \in [k]$, we have $\beta^*_j = V^* \nu^*_j$ for some vector $\nu^*_j \in \real^{k+1}$. Also
define the rotation matrix 
\begin{align*}
O = 
\begin{bmatrix}
\Uhat^\top U^* & 0 \\
0 & 1 
\end{bmatrix},
\end{align*}
so that $\Vhat O - V^*  = \begin{bmatrix} \Uhat \Uhat^\top U^*  - U^* & 0 \\ 0 & 0 \end{bmatrix}$ and we have $\|\Vhat O - V^*\| = \| \Uhat \Uhat^\top - U^* (U^*)^\top \|$ for any unitarily invariant norm $\| \cdot \|$.

Now for each $j \in [k]$ and $\ell \in [M]$, applying the triangle inequality yields
\begin{align*}
\| c_0 \beta^{\ell}_j - \beta^*_j \| &\leq \| c_0 \Vhat O \nu^{\ell}_j - \Vhat O \nu^*_j \| + \| \Vhat O \nu^*_j - V^* \nu^*_j \| \\
&\leq \| c_0 \nu^{\ell}_j - \nu^*_j \| + \| \nu^*_j \| \opnorm{ \Vhat O - V^* } \\
&\leq \| c_0 \nu^{\ell}_j - \nu^*_j \| + \Deltamax \opnorm{ \Uhat \Uhat^\top - U^* (U^*)^\top }.
\end{align*}
For each pair $(\ell, j)$, define the event
\[
\Espace^{\ell}_j(r) \defn \paren{ \| c_0 \nu^{\ell}_j - \nu^{\ell}_j \| \leq r }.
\]
We claim that if $M \geq \left( 1 + \frac{\Deltamax}{r} \right)^{k^2} \log (1 / \delta)$, we have
\begin{align} \label{clm:prob}
\Pr \paren{ \cup_{\ell \in [M]} \cap_{j \in [k]} \Espace^{\ell}_j(r) } \geq 1 - \delta.
\end{align}
Indeed, such a claim suffices, since it implies that
\begin{align*}
\min_{\ell \in [M]} \max_{j \in [k]} \| c_0 \beta^{\ell}_j - \beta^*_j \| \leq r + \Deltamax \opnorm{ \Uhat \Uhat^\top - U^* (U^*)^\top }
\end{align*}
with probability exceeding $1 - \delta$, thereby proving the theorem. It remains to establish claim~\eqref{clm:prob}.

Denote by $p$ the probability with which for a fixed pair $(\ell, j)$, we have
$\| c_0 \nu^{\ell}_j - \nu^{\ell}_j \| \leq r$. This is the ratio of the volume of the $\ell_2$-ball of radius $r$ and the $\ell_2$-ball of radius $c_0$, and so we have $p = \left( \frac{r}{r + \Deltamax} \right)^{k}$. Thus, we have
\begin{align*}
\Pr \paren{ \cap_{\ell \in [M]} \left( \cap_{j \in [k]} \Espace^{\ell}_j(r) \right)^c } &\leq (1 - p^k)^M \\
&\leq e^{-p^k M} \\
&\stackrel{\1}{\leq} \delta,
\end{align*}
where step $\1$ holds provided $M \geq \frac{1}{p^k} \log (1 / \delta)$. Putting together the pieces completes the proof.
\qed


\section{Technical results concerning the global LSE} \label{pfs:lse}

In this section, we provide a proof of the existence of the global least squares estimator that was stated in the main text. We also state and prove a lemma that shows that the global LSE is a fixed point of the AM update under a mild technical condition.

\subsection{Proof of Lemma \ref{lse.exist}}
Fix data $(x_1, y_1), \dots, (x_n, y_n)$ and let 
\begin{equation*}
  L(\gamma_1, \dots, \gamma_k) := \sum_{i=1}^n \left(y_i - \max_{j \in
    [k]} \inprod{\xi_i}{\gamma_j} \right)^2
\end{equation*}
denote the objective function in \eqref{intro:opt} with $\xi_i :=
(x_i, 1)$. The goal is to show that a global minimizer of $L(\gamma_1,
\dots, \gamma_k)$ over $\gamma_1, \dots, \gamma_k \in \R^{d+1}$
exists. For $\gamma_1, \dots, \gamma_k \in \R^{d+1}$, let 
$S_1^{\gamma}, \dots, S_k^{\gamma}$ denote a fixed partition of $[n]$
having the property that 
\begin{equation*}
  \inprod{\xi_i}{\gamma_j} = \max_{u \in [k]} \inprod{\xi_i}
    {\gamma_u} \qt{for every $j \in [k]$ and $i \in
    S_j^{\gamma}$}. 
\end{equation*}
Also, let $\widehat{\beta}_1^{\gamma}, \dots,
\widehat{\beta}_k^{\gamma}$ denote the solution to the following
constrained least squares problem: 
\begin{equation*}
\begin{aligned}
& \underset{\beta_1, \dots, \beta_k}{\text{minimize}}
& & \sum_{j=1}^k \sum_{i \in S_j^{\gamma}} \left(y_i - \inprod{\xi_i}
    {\beta_j} \right)^2\\ 
& \text{subject to}
& & \inprod{\xi_i}{\beta_j} \geq \inprod{\xi_i}{\beta_u}, u,
j \in [k], i \in S_j^{\gamma}. 
\end{aligned}
\end{equation*}
Note that the above quadratic problem is feasible as $\gamma_1,
\dots, \gamma_k$ satisfies the constraint and, consequently,
$\widehat{\beta}_1^{\gamma}, \dots, \widehat{\beta}_k^{\gamma}$
exists uniquely for every $\gamma_1, \dots, \gamma_k \in
\R^{d+1}$. Note further that, by construction, 
\begin{equation*}
  L\left(\widehat{\beta}_1^{\gamma}, \dots, \widehat{\beta}_k^{\gamma}
  \right) \leq L(\gamma_1, \dots, \gamma_k). 
\end{equation*}
and that the set 
\begin{equation*}
\Delta :=  \left\{(\widehat{\beta}^{\gamma}_1, \dots,
    \widehat{\beta}_k^{\gamma}): \gamma_1, \dots, \gamma_k \in \R^{d+1} \right\}
\end{equation*}
is finite because  $\widehat{\beta}^{\gamma}_1, \dots,
    \widehat{\beta}_k^{\gamma}$ depends on $\gamma_1, \dots, \gamma_k$
    only through the partition $S_1^{\gamma}, \dots, S_k^{\gamma}$ and
    the number of possible such partitions of $[n]$ is obviously
    finite. Finally, it is evident that
    \begin{equation*}
      (\widehat{\beta}_{1}^{(\mathsf{ls})}, \dots,
      \widehat{\beta}_{k}^{(\mathsf{ls})} ) = \argmin_{(\beta_1, \dots,
        \beta_k) \in \Delta}  L(\beta_1, \dots, \beta_k)
    \end{equation*}
    is a global minimizer of $L(\gamma_1, \dots, \gamma_k)$ as
    \begin{equation*}
      L\left(\widehat{\beta}_{1}^{(\mathsf{ls})}, \dots,
      \widehat{\beta}_{k}^{(\mathsf{ls})}  \right) \leq   L\left(\widehat{\beta}_1^{\gamma}, \dots, \widehat{\beta}_k^{\gamma}
  \right) \leq L(\gamma_1, \dots, \gamma_k)
\end{equation*}
for every $\gamma_1, \dots, \gamma_k$. This concludes the proof of
Lemma \ref{lse.exist}.

\subsection{Fixed point of AM update}

The following lemma establishes that the global LSE is a fixed point of the AM update under a mild technical condition.

\begin{lemma}\label{lse.fixed}
  Consider the global least squares estimator~\eqref{intro:opt}. Suppose that the $k$ values $\inprod{\xi_i}{\widehat{\beta}_j^{\mathsf{ls}}}$ for $j = 1, \dots, k$ are distinct for  each $i \in [n]$. Then
  \begin{equation}\label{lse.fixed.eq}
\widehat{\beta}_j^{(\mathsf{ls})} \in \argmin_{\beta \in \R^{d+1}}
\sum_{i \in S_{j}(\widehat{\beta}_1^{(\mathsf{ls})}, \dots,
  \widehat{\beta}_k^{(\mathsf{ls})})} \left(y_i - \left<\xi_i, \beta
  \right> \right)^2 \qt{for every $j \in [k]$}. 
\end{equation}
\end{lemma}

\begin{proof}
It is clearly enough to prove \eqref{lse.fixed.eq} for $j = 1$. Suppose that $\widehat{\beta}_1^{(\mathsf{ls})}$
does not minimize the least squares criterion over
$S_{1}(\widehat{\beta}_1^{(\mathsf{ls})}, \dots,
\widehat{\beta}_k^{(\mathsf{ls})})$. Let 
\begin{equation*}
\widehat{\gamma}_1^{(\mathsf{ls})} \in \argmin_{\beta \in \R^{d+1}}
\sum_{i \in S_{1}(\widehat{\beta}_1^{(\mathsf{ls})}, \dots,
  \widehat{\beta}_k^{(\mathsf{ls})})} \left(y_i - \inprod{\xi_i}{\beta} \right)^2
\end{equation*}
be any other least squares minimizer over $S_{1}(\widehat{\beta}_1^{(\mathsf{ls})}, \dots,
\widehat{\beta}_k^{(\mathsf{ls})})$ and let, for $\epsilon > 0$, 
\begin{equation*}
  \widetilde{\beta}_1 := \widehat{\beta}_1^{(\mathsf{ls})} + \epsilon
  \left(\widehat{\gamma}_1^{(\mathsf{ls})} -
    \widehat{\beta}_1^{(\mathsf{ls})} \right). 
\end{equation*}
When $\epsilon > 0$ is sufficiently small, we have
\begin{equation*}
S_{j}(\widetilde{\beta}_1, \widehat{\beta}_2^{(\mathsf{ls})} \dots,
  \widehat{\beta}_k^{(\mathsf{ls})}) = S_{j}(\widehat{\beta}_1^{(\mathsf{ls})}, \dots,
  \widehat{\beta}_k^{(\mathsf{ls})}) \qt{for every $j \in [k]$}
\end{equation*}
due to the no ties assumption and the fact that
$\widetilde{\beta}_1$  and $\widehat{\beta}_1^{(\mathsf{ls})}$ can be
made arbitrarily close as $\epsilon$ becomes small.  Thus, if 
\begin{equation*}
  U(\beta_1, \dots, \beta_k) := \sum_{i=1}^n \left(y_i - \max_{j \in
      [k]} \inprod{\xi_i}{\beta_j} \right)^2 = \sum_{j \in [k]} \; 
  \sum_{i \in S_j(\beta_1, \dots, \beta_k)} \left(y_i - \inprod{\xi_i}
      {\beta_j} \right)^2, 
\end{equation*}
then
\begin{align*}
  U(\widetilde{\beta}_1, \widehat{\beta}_2^{(\mathsf{ls})} \dots,
  \widehat{\beta}_k^{(\mathsf{ls})}) &=    \sum_{i \in S_1(\widetilde{\beta}_1, \widehat{\beta}_2^{(\mathsf{ls})} \dots,
  \widehat{\beta}_k^{(\mathsf{ls})})} \left(y_i -
                                       \inprod{\xi_i}{ 
      \widetilde{\beta}_1} \right)^2 + \sum_{j \ge 2}
  \sum_{i \in S_j(\widetilde{\beta}_1, \widehat{\beta}_2^{(\mathsf{ls})} \dots,
  \widehat{\beta}_k^{(\mathsf{ls})})} \left(y_i - \inprod{\xi_i}
                                       {\widehat{\beta}_j^{(\mathsf{ls})} } \right)^2 \\
  &= \sum_{i \in S_1(\widehat{\beta}_1^{(\mathsf{ls})}, \widehat{\beta}_2^{(\mathsf{ls})} \dots,
  \widehat{\beta}_k^{(\mathsf{ls})})} \left(y_i -
                                       \inprod{\xi_i} 
      {\widetilde{\beta}_1 } \right)^2 + \sum_{j \ge 2}
  \sum_{i \in S_j(\widehat{\beta}_1^{(\mathsf{ls})}, \widehat{\beta}_2^{(\mathsf{ls})} \dots,
  \widehat{\beta}_k^{(\mathsf{ls})})} \left(y_i - \inprod{\xi_i}
                                       {\widehat{\beta}_j^{(\mathsf{ls})}} \right)^2 \\
  &< \sum_{i \in S_1(\widehat{\beta}_1^{(\mathsf{ls})}, \widehat{\beta}_2^{(\mathsf{ls})} \dots,
  \widehat{\beta}_k^{(\mathsf{ls})})} \left(y_i -
                                       \inprod{\xi_i} 
      {\widehat{\beta}_1^{(\mathsf{ls})} } \right)^2 + \sum_{j \ge 2}
  \sum_{i \in S_j(\widehat{\beta}_1^{(\mathsf{ls})}, \widehat{\beta}_2^{(\mathsf{ls})} \dots,
  \widehat{\beta}_k^{(\mathsf{ls})})} \left(y_i - \inprod{\xi_i}
                                       {\widehat{\beta}_j^{(\mathsf{ls})}
    } \right)^2 \\ &= U(\widehat{\beta}_1^{(\mathsf{ls})}, \widehat{\beta}_2^{(\mathsf{ls})} \dots,
  \widehat{\beta}_k^{(\mathsf{ls})})
\end{align*}
where the strict inequality above comes from the fact that
$\widetilde{\beta}_{1}$ is closer to the least squares solution
$\widehat{\gamma}^{(\mathsf{ls})}_1$ compared to
$\widehat{\beta}^{(\mathsf{ls})}_1$. This leads to a contradiction as
the criterion function is smaller than its value at a global
minimizer, thereby concluding the proof. 
\end{proof}


\section{Small ball properties and examples} \label{app:sb}

We begin with a technical lemma taken from Rudelson and Vershynin~\cite[Corollary 1.4]{rudelson2014small} that shows that product measures of small-ball distributions also satisfy the small-ball condition. 

\begin{lemma}[\cite{rudelson2014small}] \label{lem:sbprod}
For $P_X$ satisfying the $(\sball, \cs)$-small-ball property~\eqref{eq:sb} and $x_1, \ldots x_m \stackrel{i.i.d.}{\sim} P_X$, there is a universal constant $C$ such that we have
\begin{align} \label{eq:sbm}
\sup_{\substack{u \in \Sd \\ w \in \real} } \Pr \paren{ \sum_{i = 1}^m (\inprod{x_i}{u} + w )^2 \leq \delta m } &\leq (C \cdot \cs \delta)^{m \sball} \text{ for all } \delta > 0.
\end{align}
\end{lemma}

We also verify that log-concave distributions and the standard Gaussian distribution satisfy the small-ball condition~\eqref{eq:sb}.

\subsection{Log-concave distribution} \label{sec:sblc}

The following result, taken from Carbery and Wright~\cite[Theorem 8]{carbery2001distributional}, provides a small-ball bound for log-concave distributions almost directly.

\begin{lemma}[\cite{carbery2001distributional}] \label{lem:cw}
Let $p: \real^d \mapsto \real$ denote polynomial of degree (at most) $\ell$, let $X$ denote a $d$-dimensional random vector drawn from a log-concave distribution, and let $S_{\delta} = \{ x \in \real^d: |p(x)| \leq \delta \}$ for each $\delta > 0$. Then for each $q > 0$, we have
\begin{align*}
\Pr \{ X \in S_\delta \} \cdot \left( \E_{X} \left[ |p(X)|^{q/\ell} \right] \right)^{1/q} \leq C q \delta^{1/\ell}. 
\end{align*}
\end{lemma}

For a unit norm vector $u \in \Sd$, setting $p(x)= (\inprod{x}{u} + w)^2$ and $q=2$, we obtain
\begin{align*}
\Pr \left( |\inprod{X}{u} + w |^2 \leq \delta \right ) \leq C \frac{1}{1 + w^2}\delta^{1/2} \leq C \delta^{1/2},
\end{align*}
where we have used the fact that $\E[\inprod{X}{u}^2 ] = 1$ since $P_X$ is isotropic. Since this holds for each pair $(u, w) \in \Sd \times \real$, we have verified that $X$ satisfies small ball condition with $(\sball, \cs) = (1/2, C^2)$. 


\subsection{Standard Gaussian distribution} \label{sec:sbg}

This is the canonical example of a sub-Gaussian distribution satisfying a small-ball condition. Suppose $X$ is a standard Gaussian random variable. For a unit vector $u \in \Sd$, this implies that $\inprod{X}{u}^2 + w$ is a central $\chi^2$ random variable with $1$ degree of freedom. 

\begin{lemma} \label{lem:lowertail}
Let $Z_\ell$ and $Z'_{\ell}$ denote central and non-central $\chi^2$ random variables with $\ell$ degrees of freedom, respectively. Then for all $p \in [0, \ell]$, we have
\begin{align}
\Pr \{Z'_\ell \leq p\} \leq \Pr \{Z_\ell \leq p\} \leq \left( \frac{p}{\ell} \exp\left(1 - \frac{p}{\ell} \right)\right)^{\ell/2} = \exp\left( - \frac{\ell}{2} \left[ \log \frac{\ell}{p} + \frac{p}{\ell} - 1 \right] \right)\label{eq:lowertail}
\end{align}
\end{lemma}

Applying the above lemma for $\ell=1$ and $t  = \delta$, we obtain
\begin{align*}
\Pr \paren{ |\inprod{X}{u} + w |^2 \leq \delta } \leq \left ( e \delta \right )^{1/2}.
\end{align*}
Hence the standard Gaussian satisfies the small-ball condition with $(\sball, \cs) = (1/2, e)$. 

For completeness, we provide a proof of Lemma~\ref{lem:lowertail} below.

\subsubsection{Proof of Lemma~\ref{lem:lowertail}}
The fact that $Z'_{\ell} \stackrel{st.}{\leq} Z_\ell$ follows from standard results (see, e.g.,~\cite{}) that guarantee that central $\chi^2$ random variables stochastically dominate their non-central counterparts.

The tail bound is a simple consequence of the Chernoff bound. In particular, we have for all $\lambda > 0$ that
\begin{align}
\Pr \{ Z_\ell \leq p \} &= \Pr \{ \exp(-\lambda Z_\ell) \geq \exp(-\lambda p)\}  \nonumber\\
&\leq \exp(\lambda p) \E \left[ \exp(-\lambda Z_\ell) \right] \nonumber\\
&= \exp(\lambda p) (1 + 2\lambda)^{-\frac{\ell}{2}}. \label{eq:lasteq}
\end{align}
where in the last step, we have used $\E \left[ \exp(-\lambda Z_\ell) \right] = (1 + 2\lambda)^{-\frac{\ell}{2}}$, which is valid for all $\lambda > -1/2$. Minimizing the last expression over $\lambda > 0$ then yields the choice $\lambda^* = \frac{1}{2} \left( \frac{\ell}{p} - 1 \right)$, which is greater than $0$ for all $0 \leq p \leq \ell$. Substituting this choice back into equation~\eqref{eq:lasteq} proves the lemma.
\qed



\section{Fundamental limits} \label{app:lbs}
In this section, we present two lower bounds: one on the minimax risk of parameter estimation, and another on the risk of the least squares estimator with side-information.

\subsection{Minimax lower bounds}
\label{subsec:minimax}

Recall our notation $\Thetastar$ for the matrix whose rows consist of the parameters $\sigstar_1, \ldots, \sigstar_k$. Assume that the intercepts $b^*_1,\ldots,b^*_k$ are identically zero, so that $\appmeas_i = \meas_i$ and $\AppXmat = \Xmat$. For a fixed matrix $\Xmat$, consider the observation model
\begin{align}
\label{eq:model_minimax}
y = \max \left( \Xmat (\Thetastar)^\top \right) + \epsilon,
\end{align} 
where $y \in \R^\numobs$, the noise $\epsilon \sim \NORMAL(0,\sigma^2 I_n)$ is chosen independently of $\Xmat$, and the $\max$ function is computed row-wise. 

\begin{proposition} \label{prop:minimax}
There is an absolute constant $C$ such that the minimax risk of estimation satisfies
\begin{align*}
\inf_{\widehat{\Theta}} \; \sup_{\Thetastar \in \real^{k \times d} } \mathbb{E} \left[ \frac{1}{n} \fronorm{ \Xmat (\widehat{\Theta} - \Thetastar)^\top }^2 \right ] \geq C \frac{\sigma^2 k d}{n}.
\end{align*}
\end{proposition}
Here, the expectation is taken over the noise $\epsilon$, and infimum is over all measurable functions of the observations $(X, y)$. Indeed, when $\Xmat$ is a random Gaussian matrix, it is well conditioned and has singular values of the order $\sqrt{n}$, so that this bound immediately yields
\begin{align*}
\inf_{\widehat{\Theta}} \; \sup_{\Thetastar \in \real^{k \times d} } \mathbb{E} \left[ \frac{1}{n} \fronorm{ \widehat{\Theta} - \Thetastar }^2 \right ] \geq C \frac{\sigma^2 k d}{n}.
\end{align*}
Let us now provide a proof of the proposition.
\begin{proof}
The proof is based on a standard application of Fano's inequality (see, e.g., Wainwright \cite[Chapter 15]{wainwright2019high} and Tsybakov~\cite[Chapter 2]{tsybakov_nonparametric}). 
For a tolerance level $\delta > 0$ to be chosen, we choose the local set 
\begin{align*}
F= \left \lbrace \Xmat \Theta^\top \in \R^{n \times k} \,\, \bigg | \,\, \fronorm{\Xmat \Theta^\top } \leq 4\delta  \sqrt{k n} \right \rbrace
\end{align*}
 and let $\paren{ X (\Theta^1)^\top, \ldots, X (\Theta^M)^\top }$
be a $2\delta  \sqrt{ kn}$-packing of the set in the Frobenius norm. This can be achieved by packing the $j$-th column $Q_j := \left \lbrace \Xmat \theta_j \,\, | \norms{\Xmat \theta_j}_2 \leq 4 \delta \sqrt{n} \right \rbrace$ at level $2\delta \sqrt{n}$ in $\ell_2$ norm for all $j \in [k]$. Standard results yield the bound $\log M \geq C_1 \cdot k d \log 2$.

For each $i \neq j$, we have
\begin{align}
\label{eq:bounded_minimax}
2\delta \sqrt{k} \leq \frac{\fronorm{ \Xmat (\Theta^i -\Theta^j)^\top }}{\sqrt{n}} \leq 8 \delta \sqrt{k}.
\end{align}
Let $\mathbb{P}_j = \NORMAL \left( \max( \Xmat (\Theta^j)^\top), \sigma^2 I_n \right)$ denote the distribution of the observation vector $y$ when the true parameter is $\Theta^j$. We thus obtain
\begin{align*}
\kull{\mathbb{P}_j}{\mathbb{P}_i} =\frac{1}{2\sigma^2} \norms{\max( \Xmat (\Theta^j)^\top \Xmat) - \max( \Xmat (\Theta^i)^\top ) }_2^2 \leq \frac{1}{2 \sigma^2} \fronorm{ \Xmat (\Theta^j -\Theta^i)^\top }^2,
\end{align*}
where the inequality follows since the $\max$ function is $1$-Lipschitz in $\ell_2$ norm. Putting together the pieces yields
\begin{align*}
\kull{\mathbb{P}_j}{\mathbb{P}_i} \leq \frac{32 k  \delta^2 n}{\sigma^2},
\end{align*}
so that the condition
\begin{align*}
\frac{\frac{1}{M^2}\sum_{i,j} \kull{\mathbb{P}_{\Theta^j}}{\mathbb{P}_{\Theta^k}} + \log 2}{\log M} \leq \frac{1}{2}
\end{align*}
is satisfied with the choice $\delta^2 = C \frac{\sigma^2 d}{n}$. Finally, applying Fano's inequality (see, e.g., \cite[Proposition 15.2]{wainwright2019high}) yields the minimax lower bound
\begin{align}
\label{eq:prediction_minimax}
\inf_{\widehat{\Theta}} \sup_{\Thetastar} \mathbb{E} \left[ \frac{1}{n}\fronorm{ \Xmat(\widehat{\Theta} - \Thetastar )^\top }^2 \right ] \geq C \frac{\sigma^2 k d}{n}.
\end{align}
\end{proof}


\subsection{Performance of unconstrained least squares with side-information} \label{sec:covariance}

In this section, we perform an explicit computation when $k = 3$ and $d = 2$ to illustrate the cubic $\pimin$ dependence of the error incurred by the unconstrained least squares estimator, even when provided access to the true partition $\paren{ S_j (\appsigstar_1, \ldots, \appsigstar_3) }_{j = 1}^3$.

We begin by defining our unknown parameters.
For a scalar $\alpha \in (0, \pi/4)$, let
\begin{align*}
\sigstar_1 = \sin (\alpha) \cdot e_1, \qquad \sigstar_2 = \cos (\alpha) \cdot e_2, \qquad \text{ and } \sigstar_3 = - \cos (\alpha) \cdot e_2,
\end{align*}
and set $\bstar_j = 0$ for $j = 1, 2, 3$.

Now an explicit computation yields that the cone on which $\sigstar_1$ attains the maximum is given by
\begin{align*}
\mathcal{C}_1 \defn \paren{ x \in \real^2: \inprod{x}{\sigstar_1} \geq \max_{j \in [k] } \inprod{x}{\sigstar_j} } =
 \paren{x \in \real^2: x_1 \geq 0, \; |x_2| \leq x_1 \tan (\alpha) }.
\end{align*}
Now consider a Gaussian random vector in $\real^2$ truncated to that cone.
In particular, consider a two-dimensional random variable $W$ with density
$\psi(x) \ind{ x \in \mathcal{C}_1 } / \vol(\mathcal{C}_1)$, where $\psi$ is the two-dimensional standard Gaussian density and $\vol (S)$ denotes the Gaussian volume of a set $S$. Note that we have $\vol(\mathcal{C}_1) = \alpha / \pi$ by construction.

Let us now compute the second order statistics of $W$, using polar coordinates with $R^2$ denoting a $\chi^2_2$ random variable.
%
%
%
%
%
The individual second moments take the form
\begin{align*}
\E[W_1^2] = \frac{\pi}{\alpha} \E[R^2] \left( \frac{1}{2 \pi} \int_{-\alpha}^\alpha \cos^2 \phi d\phi \right) = 1,
\end{align*}
and
\begin{align*}
\E[W_2^2] = \frac{\pi}{\alpha} \E[R^2] \left( \frac{1}{2 \pi} \int_{-\alpha}^\alpha \sin^2 \phi d\phi \right) 
= \frac{1}{\alpha} \left( \alpha - \sin (2 \alpha)/2 \right) \sim \alpha^2.
\end{align*}

On the other hand, the cross terms are given by
\begin{align*}
\E[W_1 W_2] = \frac{\pi}{\alpha} \E[R^2] \left( \frac{1}{2 \pi} \int_{-\alpha}^\alpha \sin(\phi) \cos(\phi) d\phi \right) = 0.
\end{align*}
Thus, it can be verified that for all $\alpha \in [0, \pi/4]$, the second moment matrix of $W$ has a tuple of singular values $(1, c \alpha^2 )$ for an absolute constant $c$. 

Let us now use this calculation to reason about the least squares estimator. Drawing $n$ samples from the Gaussian distribution on $\real^2$, we expect $n_1 \sim \frac{\alpha}{\pi} n$ of them to fall in the set $\mathcal{C}_1$ with high probability. Collect these samples as rows of a matrix $\Xmat_1$. 
When $n$ is large enough, i.e., on the order of $\alpha^{-3}$, standard bounds (as in Section~\ref{sec:truncmom}) can be applied to explicitly evaluate the singular values of the matrix $\frac{1}{n_1} \Xmat_1^\top \Xmat_1$. In particular, we have
\begin{align*}
\lambda_1 \left( \frac{1}{n_1} \Xmat_1^\top \Xmat_1 \right) = c' \qquad \text{ and } \qquad \lambda_2 \left( \frac{1}{n_1} \Xmat_1^\top \Xmat_1 \right) = c \alpha^2.
\end{align*}
We now provide the $n_1 \times 2$ matrix $\Xmat_1$ as side information to a procedure whose goal is to estimate the unknown parameters. Clearly, given this matrix, a natural procedure to run in order to estimate $\sigstar_1$ is the (unconstrained) least squares estimator on these samples, which we denote by $\thetahat_1$. As is well known, the rate obtained (in the fixed design setting) by this estimator with $\sigma$-sub-Gaussian noise is given by
\begin{align*}
\mathbb{E} \left[ \| \thetahat_1 - \sigstar_1 \|^2 \right] &= \sigma^2 \tr (\Xmat_1^\top \Xmat_1)^{-1} \\
&= \sigma^2 \frac{1}{n_1} \left( c \alpha^{-2} + c' \right) \\
&\sim \sigma^2 \frac{1}{\alpha^3 n},
\end{align*}
where the last two relations hold with exponentially high probability in $n$.
We have thus shown that the unconstrained least squares estimator (even when provided with additional side information) attains an error having cubic dependence on $\alpha \sim \pimin$. While this does not constitute an information theoretic lower bound, our calculation provides some evidence for the fact that, at least when viewed in isolation, the dependence of our statistical error bound~\eqref{eq:stat-err} on $\pimin$ is optimal for Gaussian covariates.



\section{Background and technical lemmas used in the proofs of Theorems~\ref{thm:mt} and~\ref{thm:sbgen}}

In this section, we collect statements and proofs of some technical lemmas used in the proofs of our results concerning the AM algorithm.

\subsection{Bounds on the ``volumes" of wedges in $\real^d$} \label{app:volbd}

For a pair of scalars $(w, w')$ and $d$-dimensional vectors $(u, u')$, recall that we define the \emph{wedge} formed by the $d+1$-dimensional vectors $v = (u, \; w)$ and  $v' = (u', \; w')$ as the region
\begin{align*}
W(v, v') = \{ x \in \real^d: (\inprod{x}{u} + w) \cdot (\inprod{x}{u'} + w') \leq 0\}.
\end{align*}
Note that the wedge is a purely geometric object. 

For any set $\mathcal{C} \subseteq \real^d$, let 
\[
\vol_{P_X} (\mathcal{C}) = \Pr_{X \sim P_X } \paren{ X \in \Cspace }
\]
denote the volume of the set under the measure corresponding to the covariate distribution.


We now bound the volume of a wedge for both the Gaussian distribution and for distributions satisfying the small-ball condition.

\begin{lemma} \label{lem:vol_gauss}
Suppose that $P_X = \NORMAL(0, I_d)$, and that for a pair of scalars $(w, w')$, $d$-dimensional vectors $(u, u')$, and $v = (u, \; w)$ and $v' = (u', \; w')$, we have $\frac{\| v - v' \|}{\| u \|} < 1/2$. Then, there is a positive constant $C$ such that
\begin{align*}
\vol_{P_X} ( W(v, v') ) &\leq C \frac{\| v - v' \|}{\| u \|} \log^{1/2} \left(\frac{2 \| u \|}{\| v - v' \|}  \right).
\end{align*}
\end{lemma}

The above lemma has an analogue when the distribution $P_X$ is $\eta$ sub-Gaussian and satisfies a $(\sball, \cs)$-small-ball condition. 

\begin{lemma} \label{lem:vol_gen}
Suppose that Assumption~\ref{assptn:gen} holds, and that for a pair of scalars $(w, w')$, $d$-dimensional vectors $(u, u')$, and $v = (u, \; w)$ and $v' = (u', \; w')$, we have $\frac{\| v - v' \|}{\| u \|} < 1/2$. Then, there is a positive constant $C$ such that
\begin{align*}
\vol_{P_X}( W(v, v') ) &\leq C \left( \Csig' \frac{\| v - v' \|^2}{\| u \|^2} \log \left(\frac{\| u \|}{\| v - v' \|}  \right) \right)^{\sball},
\end{align*}
where $\Csig'$ is a constant that depends only on the tuple $(\eta, \sball, \cs)$.
\end{lemma}
\subsubsection{Proof of Lemma~\ref{lem:vol_gauss}}

We use the shorthand $\vol_{\NORMAL}$ to denote the Gaussian volume. Using the notation $\xi = (x, \; 1) \in \real^{d + 1}$ to denote the appended covariate, we have
\begin{align*}
\vol_{\NORMAL} ( W(v, v') ) = \Pr \paren{ \inprod{\xi}{v} \cdot \inprod{\xi}{v'} \leq 0 },
\end{align*}
where the probability is computed with respect to Gaussian measure.

In order to prove a bound on this probability, we begin by bounding the associated indicator random variable as
\begin{align}
\ind{\inprod{\xi}{v} \cdot \inprod{\xi}{v'} \leq 0} &= \ind{\inprod{\xi}{v' - v}^2 \geq \inprod{\xi}{v}^2} \notag \\
& \leq \ind{\inprod{\xi}{v' - v}^2 \geq t} + \ind{\inprod{\xi}{v}^2 \leq t}, \label{eq:prob_split_intercept}
\end{align}
where inequality~\eqref{eq:prob_split_intercept} holds for all $t \geq 0$. In order to bound the expectation of the second term, we write
\begin{align*}
\Pr \paren{ \inprod{\xi}{v}^2 \leq t } &= \Pr \paren{ \norms{u}^2 \chi_{nc}^2 \leq t } \\
&\stackrel{\1}{\leq} \left (\frac{et}{\norms{u}^2}\right )^{1/2}
\end{align*}
where $\chi_{nc}^2$ is a non-central chi-square random variable centered at $\frac{w}{\norms{u}}$, and step $\1$ follows from standard $\chi^2$ tail bounds (see Lemma~\ref{lem:lowertail}).

It remains to control the expectation of the first term on the RHS of inequality~\eqref{eq:prob_split_intercept}. We have
\begin{align*}
\Pr \paren{\inprod{\xi}{v' - v}^2 \geq t} &\leq \Pr \paren{ 2\inprod{x}{u' - u}^2 + 2 (w' - w)^2 \geq t} \\
&\leq \Pr \paren{ \norms{u-u'}^2 \chi^2 \geq \frac{t}{2}-\norms{v-v'}^2 }.
\end{align*}
Now, invoking a standard sub-exponential tail bound on the upper tail of a $\chi^2$ random variable yields
\begin{align*}
\Pr \paren{\inprod{\xi}{v' - v}^2 \geq t} &\leq c_1 \exp \bigg ( - \frac{c_2}{\norms{u-u'}^2} \bigg \lbrace \frac{t}{2}- \norms{v-v'}^2 \bigg \rbrace \bigg ) \\
& \leq c_1 \exp \bigg ( - \frac{c_2}{\norms{v - v'}^2} \bigg \lbrace \frac{t}{2}- \norms{v - v'}^2 \bigg \rbrace \bigg ).
\end{align*}
Putting all the pieces together, we obtain
\begin{align*}
\vol( W(v, v') )  \leq c_1 \exp \bigg ( - \frac{c_2}{\norms{v - v'}^2} \bigg \lbrace \frac{t}{2}- \norms{v - v'}^2 \bigg \rbrace \bigg ) + \left (\frac{et}{\norms{u}^2}\right )^{1/2}.
\end{align*}
Substituting $t = 2 c \norms{v - v'}^2  \log (2 \| u \| / \| v - v' \| ) $, which is a valid choice provided $\frac{\| v - v' \|}{\| u \|} < 1/2$, yields the desired result.
\qed

\subsubsection{Proof of Lemma~\ref{lem:vol_gen}}

With the same notation as the previous proof, inequality~\eqref{eq:prob_split_intercept} still applies to this setting.
However, we now have
\begin{align*}
\Pr \paren{ \inprod{\xi}{v}^2 \leq t } \leq \left (\frac{\cs t}{\norms{u}^2}\right )^{\sball},
\end{align*}
since $P_X$ satisfies a small-ball condition.

Furthermore, the sub-Gaussianity of the covariate distribution yields the upper tail bound
\begin{align*}
\Pr \paren{\inprod{\xi}{v' - v}^2 \geq t} &\leq c_1 \exp \bigg ( - \frac{c_2}{\eta^2 . \norms{u-u'}^2} \bigg \lbrace \frac{t}{2}- \norms{v-v'}^2 \bigg \rbrace \bigg ) \\
& \leq c_1 \exp \bigg ( - \frac{c_2}{\eta^2. \norms{v - v'}^2} \bigg \lbrace \frac{t}{2}- \norms{v - v'}^2 \bigg \rbrace \bigg ).
\end{align*}

Putting all the pieces together, we obtain
\begin{align*}
\vol_{P_X}( W(v, v') )  \leq c_1 \exp \bigg ( - \frac{c_2}{\eta^2 . \norms{v - v'}^2} \bigg \lbrace \frac{t}{2}- \norms{v - v'}^2 \bigg \rbrace \bigg ) + \left (\frac{\cs t}{\norms{u}^2}\right )^{\sball}.
\end{align*}
Substituting $t = 2 c \sball \eta^2 \norms{v - v'}^2 \left( \log (\| u \| / \| v - v' \| ) \right)$ yields the desired result.
\qed

\subsection{Uniform bounds on singular values of (sub-)matrices}

We now state and prove two technical lemmas that bound the maximum and minimum singular values of a matrix whose rows are drawn from a sub-Gaussian distribution obeying the small-ball property. Our results on the minimum singular value are similar in spirit to those of Rudelson and Vershynin~\cite{rudelson2008littlewood}, but proved under a slightly different set of assumptions.

\begin{lemma} \label{lem:sgmax}
Suppose that the covariates are drawn i.i.d. from a $\eta$-sub-Gaussian distribution. Then for a fixed subset $S \in [n]$ of size $\ell$ and each $t \geq 0$, we have
\begin{align*}
\Pr \paren{ \eigmax \left( \Xi_S^\top \Xi_S \right) \geq  \ell + \etatil^2 (\sqrt{\ell d} + d + \ell t) } &\leq 2e^{-\ell \min \{ t, t^2 \} },
\end{align*}
where $\etatil = \max\paren{\eta, 1}$.
\end{lemma}

The second lemma controls the minimum singular value of any sub-matrix of $\Xi$ that has sufficiently many rows. Recall that distribution $P_X$ is isotropic and $\eta$-sub-Gaussian, and satisfies the $(\sball, \cs)$ small-ball condition~\eqref{eq:sb}.
\begin{lemma} \label{lem:minsingsb}
Suppose that Assumption~\ref{assptn:gen} holds, and that $\ell \geq \max\paren{ 4d, \frac{d + 1}{\sball} }$. Let $\etatil = \max\{\eta, 1\}$. Then for a fixed subset $S \in [n]$ of size $\ell$ and for each positive $\epsilon < \min\{ (\etatil)^2 \sball, e^{-4 / \sball} \}$, we have
\begin{align*}
\Pr \paren{ \eigmin \left( \Xi_S^\top \Xi_S \right) \leq \ell \epsilon } &\leq 3 \left( 4 \cs^2  \max\paren{ 9 \etatil^2 \sball, 1} \epsilon \log (1 / \epsilon) \right)^{\ell \sball/2}.
\end{align*}
\end{lemma}

\subsubsection{Proof of Lemma~\ref{lem:sgmax}}

Let $\{z_i\}_{i = 1}^{\ell}$ denote i.i.d. Rademacher variables, and collect these in an $\ell$-dimensional vector $z$. Let $D = \diag(z)$ denote a diagonal matrix, and note that by unitary invariance of the singular values, the singular values of the matrix $\widetilde{\Xi}_S = D \Xi_S$ are the same as those of $\Xi_S$.

By construction, the matrix $\widetilde{\Xi}_S$ has i.i.d. rows, and the $i$-th row is given by $z_i (\meas_i, \; 1)$. 
For a $d+1$ dimensional vector $\widetilde{\lambda} = (\lambda, \; w)$ with $\lambda \in \real^d$ and $w \in \real$, we have
\begin{align*}
\mathbb{E} \left[ \exp ( \inprod{ \widetilde{\lambda}} {z_i (\meas_i,\; 1)} ) \right] &= \frac{e^w}{2} \cdot \mathbb{E} \left[ \exp ( \inprod{\lambda}{\meas_i} ) \right] +  \frac{e^{-w}}{2} \cdot \mathbb{E} \left[ \exp ( - \inprod{\lambda}{\meas_i} ) \right] \\
&= \exp (\|\lambda\|^2 \eta^2/2 ) \cdot \frac{1}{2} \left( e^{w} + e^{-w} \right) \\
&\leq \exp (\|\lambda\|^2 \eta^2/2 ) \cdot \exp (w^2/2) \leq \exp (\| \widetilde{\lambda} \|^2 \etatil^2 / 2 ).
\end{align*}
where we have used the fact that $\meas_i$ is zero-mean and $\eta$ sub-Gaussian.

Since the rows of $\widetilde{\Xi}_S$ are i.i.d., zero-mean, and $\etatil$-sub-Gaussian, applying~\cite[Theorem 6.2]{wainwright2019high} immediately yields the lemma.
 \qed

\subsubsection{Proof of Lemma~\ref{lem:minsingsb}}
We let $M$ denote the $\ell \times (d+1)$ matrix $\Xi_S$. By the variational characterization of the minimum eigenvalue, we have
\[
\eigmin \left( M^\top M \right) = \inf_{v \in \mathbb{S}^d } \eucnorm{ M v }^2 = \left( \inf_{v \in \mathbb{S}^d } \eucnorm{ M v } \right)^2.
\]
Let $Z_v = \eucnorm{ M v }$ denote a random process indexed by $v$, and let $Z = \inf_{v \in \mathbb{S}^d } Z_v$; we are interested bounding the lower tail of $Z$. Consider a $\rho$-covering $\paren{ v^1, \ldots, v^N}$ of the set $\mathbb{S}^{d}$ in $\ell_2$ norm, with $N \leq (1 + 2 / \rho)^{d+1}$. Letting $v^j$ be the closest element of the cover to $v$, we have
\begin{align*}
Z_v \geq Z_{v^j} - |Z_{v} - Z_{v^j}| \geq Z_{v^j} - \opnorm{M} \cdot \rho, 
\end{align*}
so that we have the bound $Z \geq \min_{j \in [N]} Z_{v^j} - \opnorm{M} \cdot \rho$. We have thus reduced the infimum over the unit shell to a finite minimum.

We thus have
\begin{align*}
\Pr \paren{ \eigmin \left( M^\top M \right) \leq \ell \epsilon } &= \Pr \paren{ Z^2 \leq \ell \epsilon } \\
&= \Pr \paren{ Z \leq \sqrt{\ell \epsilon} } \\
&\leq \Pr \paren{ \min_{j \in [N]} Z_{v^j} \leq 2 \sqrt{\ell \epsilon} } + \Pr \paren{ \opnorm{ M} \rho \geq \sqrt{\ell \epsilon} } \\
&\leq N \Pr \paren{ Z_{v^j} \leq 2 \sqrt{\ell \epsilon} } + \Pr \paren{ \opnorm{ M} \geq \sqrt{\ell \epsilon} / \rho },
\end{align*}
where we have used the union bound in each of the last two steps.

Now note that for each $j \in [N]$, the small ball condition yields the bound
\begin{align*}
\Pr \paren{ Z_{v^j} \leq 2 \sqrt{\ell \epsilon} } = \Pr \paren{ \| M v^j \|^2 \leq 2 \ell \epsilon } \leq (2 C \cs \epsilon)^{\ell \sball},
\end{align*}
where we have used Lemma~\ref{lem:sbprod} to reason about the product measure of small-ball distributions.

Furthermore, since $M$ has $\etatil$ sub-Gaussian rows, we may apply Lemma~\ref{lem:sgmax} to obtain
\begin{align*}
\Pr \paren{ \opnorm{ M} \geq \sqrt{\ell \epsilon} / \rho } \leq 2 \exp \left(  - \ell \paren{ \frac{\epsilon}{\rho^2 \etatil^2} - \etatil^{-2} - \frac{d + \sqrt{\ell d}}{\ell} } \right),
\end{align*}
which holds provided 
$
\paren{ \frac{\epsilon}{\rho^2 \etatil^2} - \etatil^{-2} - \frac{d + \sqrt{\ell d}}{\ell} } \geq 1. 
$
When $\epsilon \leq e^{-4/\sball}$, the choice $\rho = \rho_0 \defn \sqrt{\frac{\epsilon}{\sball \etatil^2} } \log^{-1} (1 / \epsilon)$ ensures that
\[
\frac{\epsilon}{\rho_0^2 \etatil^2} - \etatil^{-2} - \frac{d + \sqrt{\ell d}}{\ell} \stackrel{\1}{\geq} \sball \log (1 / \epsilon) - 2 \geq \frac{\sball}{2} \log(1 / \epsilon),
\]
where in step $\1$ we have used the properties $\etatil \geq 1$ and $\ell \geq 4d$.
This yields the bound
\[
\Pr \paren{ \opnorm{ M} \geq \sqrt{\ell \epsilon} / \rho } \leq 2 \epsilon^{\ell \sball / 2},
\]
and putting together the pieces by substituting $N = \left( 1 + \frac{2}{\rho_0} \right)^{d + 1}$, we have
\begin{align*}
\Pr \paren{ \eigmin \left( M^\top M \right) \leq \ell \epsilon } &\leq \left( 1 + \frac{2}{\rho_0} \right)^{d + 1} \left( 2 \cs \epsilon \right)^{\ell \sball} + 2 \epsilon^{\ell \sball / 2}.
\end{align*}
Now note that we have $\epsilon < \min\{ 1, \etatil^2 \sball \}$, so that $\rho_0 \leq 1$, so that $1 + 2 /\rho_0 \leq 3/\rho_0$. Therefore, we have
\begin{align*}
\Pr \paren{ \eigmin \left( M^\top M \right) \leq \ell \epsilon } &\leq (9 \etatil^2 \sball)^{(d + 1)/2} (2 \cs)^{\ell \sball} \epsilon^{\ell \sball - (d + 1)/2 } \log^{d + 1} (1 / \epsilon) + 2 \epsilon^{\ell \sball / 2} \\
&\leq 3 \left( 4 \cs^2  \max\paren{ 9 \etatil^2 \sball, 1} \epsilon \log (1 / \epsilon) \right)^{\ell \sball/2},
\end{align*}
where we have used the fact that $\ell \sball \geq d + 1$. 
This completes the proof.
\qed

\subsection{Growth Functions and Uniform Empirical Concentration}
\label{sec:growth-fns}

We now briefly introduce growth functions and uniform laws derived from them, and refer the interested reader to Mohri et al.~\cite{mohri2018foundations} for a more in-depth exposition on these topics.

We define growth functions in the general multi-class setting~\cite{daniely2012multiclass}. Let $\mathcal{X}$ denote a set, and let $\mathcal{F}$ denote a family of functions mapping $\Xspace \mapsto \{ 0, 1, \ldots, k-1 \}$.
The \emph{growth function} $\Pi_{\mathcal{F}}: \mathbb{N} \to \real$ of $\mathcal{F}$ is defined via
\[
\Pi_{\mathcal{F}}(n) := \max_{x_1,\ldots,x_n \in \mathcal{X}} \abs{\paren{ \{ f (x_1), f (x_2),\ldots, f (x_n) \} ~:~ f \in \mathcal{F} }}.
\]
In words, it is the cardinality of all possible labelings of $n$ points in the set $\Xspace$ by functions in the family $\mathcal{F}$.

A widely studied special case arises in the case $k = 2$, with the class of binary functions. In this case, a natural function class $\mathcal{F}$ is formed by defining $\mathcal{C}$ to be a family of subsets of $\mathcal{X}$, and identifying each set $C \in \mathcal{C}$ with its indicator function $f_C \defn 1_C : \mathcal{X} \to \{0,1\}$. In this case, define $\mathcal{F}_\mathcal{C} = \{ f_C: C \in \mathcal{C} \}$.
%
%
%
%
%
%
%
A bound on the growth function for such binary function provides following guarantee for the uniform convergence for the empirical measures of sets belonging to $\mathcal{C}$.

\begin{lemma}[Theorem 2 in~\cite{vapnik1968uniform}] \label{lem:uniform-convergence}
	Let $\mathcal{C}$ be a family of subsets of a set $\mathcal{X}$. Let $\mu$ be a probability measure on $\mathcal{X}$, and let $\hat{\mu}_m := \frac{1}{m} \sum_{i=1}^m \delta_{X_i}$ be the empirical measure obtained from $m$ independent copies of a random variable $X$ with distribution $\mu$. For every $u$ such that $m \geq 2/u^2$, we have
	\begin{equation}
	\Pr \left\{\sup_{C \in \mathcal{C}} \abs{\hat{\mu}_m(C) - \sigma(C)} \geq u \right\} \leq 4\Pi_{\mathcal{F}_\mathcal{C} } (2m) \exp(-mu^2/16).
	\end{equation}
\end{lemma}


We conclude this section by collecting some results on the growth functions of various function classes. For our development, it will be specialize to the case $\Xspace = \real^{d}$. 

Define the class of binary functions $\mathcal{F}_{\mathcal{H}}$ as the set of all functions of the form
\begin{align*}
f_{\theta, b} (x) \defn \frac{\sgn( \inprod{x}{\theta} + b ) + 1}{2};
\end{align*}
specifically, let $\mathcal{F}_{\mathcal{H}} \defn \paren{ f_{\theta, b} : \theta \in \real^d, b \in \real }$. In particular, these are all functions that can be formed by a $d$-dimensional hyperplane.

Using the shorthand $B_1^k = \{ B_1, \ldots, B_k \}$, define the binary function
\begin{align*}
g_{\theta_1^k, b_1^k }(x) \defn \prod_{i = 1}^k f_{\theta_i, b_i} (x),
\end{align*}
and the binary function class corresponding to the intersection of $k$ hyperplanes
\begin{align*}
\mathcal{G}_{\mathcal{H}^k} \defn \paren{ g_{\theta_1^k, b_1^k }: \theta_1, \ldots, \theta_k \in \real^d \; , \; b_1, \ldots, b_k \in \real }.
\end{align*}

Finally, we are interested in the $\argmax$ function over hyperplanes. Here, define the function
\begin{align*}
m_{\theta_1^k, b_1^k }(x) \defn \argmax_{j \in [k]} \left( \inprod{\theta_j}{x} + b_j \right) - 1,
\end{align*}
mapping $\real^d \mapsto \{0, \ldots, k - 1\}$.
The function class that collects all such functions is given by
\begin{align*}
\mathcal{M}_k \defn \paren{ m_{\theta_1^k, b_1^k }: \theta_1, \ldots, \theta_k \in \real^d\; , \; b_1, \ldots, b_k \in \real }.
\end{align*}

The following results bound the growth functions of each of these function classes. We first consider the function classes $\mathcal{F}_{\mathcal{H}}$ and $\mathcal{G}_{\mathcal{H}^k}$, for which bounds on the VC dimension directly yield bounds on the growth function.

\begin{lemma}[Sauer-Shelah (e.g. Section 3 of Mohri et al.~\cite{mohri2018foundations})] \label{lem:sauer-shelah}
We have
\begin{align}
	\Pi_{\mathcal{F}_{\mathcal{H}}}(n) &\leq \left(\frac{en}{d+1} \right)^{d+1}, \text{ and } \\
	\Pi_{\mathcal{G}_{\mathcal{H}^k}}(n) &\leq \left(\frac{en}{d+1} \right)^{k(d+1)}.
\end{align}
\end{lemma}
The second bound can be improved (see, e.g.~\cite{csikos2018optimal}), but we state the version obtained by a trivial composition of individual halfspaces.

The following bound on the growth function of the class $\mathcal{M}_k$ is also known.

\begin{lemma}[Theorem 3.1 of Daniely et al.~\cite{daniely2012multiclass}] \label{lem:arg_max_class}
For an absolute constant $C$, we have
\[
\Pi_{\mathcal{M}_k} (n) \leq \left(\frac{en}{C k(d+1)\log (kd) } \right)^{C k(d+1) \log (kd)}.
\]	
\end{lemma}

\section{Background and technical lemmas used in the proof of Theorem~\ref{thm:pca}} \label{app:thm3}

We begin by stating a result of Harge~\cite[Theorem 1.2]{harge2004convex} (see also Hu~\cite{hu1997ito}) that guarantees that convex functions of a Gaussian random vector are positively correlated. We state it below in the notation of the current paper.
\begin{lemma}[\cite{harge2004convex}] \label{lem:correlation}
Let $f$ and $g$ be two convex functions on $\real^d$, and let $X$ be a standard $d$-dimensional Gaussian vector. Then
\begin{align}
\E [ f(X) g(X) ] \geq (1 + \inprod{m(g)}{m(f)} ) \E [f (X)] \E [g (X)],
\end{align}
where for any $d$-variate function $h$, we have $m(h) = \frac{\E [X h(X)] }{\E[ h(X)] }$.
\end{lemma}

We also prove Lemma~\ref{lem:dev}, which was used in the proof of Theorem~\ref{thm:pca}.

\subsection{Proof of Lemma~\ref{lem:dev}} \label{app:pf-moments}

We prove each bound separately. First, by the rotation invariance of the Gaussian distribution, we may assume that $\Ustar = [e^d_1 \; \ldots \; e^d_k]$, so that the $\max$ is computed as a function of the $k$ coordinates $X_1, \ldots X_k$.

We also define some events that we make use of repeatedly in the proofs.
For each $i \in [n]$, define the events
\begin{align*}
\Espace_i &= \{ |x_{i, j}| \leq 5 \sqrt{ \log (2nk) } \text{ for all } 1 \leq j \leq k \}, \text{ and} \\
\Fspace_i &= \{ |\epsilon_{i}| \leq 5 \sigma \sqrt{ \log (2n) } \}.
\end{align*}
Note that by standard sub-Gaussian tail bounds, we have $\Pr\{ \Espace_i^c \} \leq 2 n^{-12}$ and $\Pr\{ \Fspace_i^c \} \leq 2n^{-12}$ for each $i \in [n]$. For notational convenience, define for each $i$ the modified covariate $z_i = x_i \cdot \ind{ \Espace_i }$.

We have
\[
|\max(\Theta^* z_i + \bstar)| \leq C \max_{j \in [k]} \| \sigstar_j \|_1 \sqrt{ \log (nk) } + |\bstar_j| \leq \left ( C \sqrt{ \log (nk) } \right) \varsigma
\]
almost surely, where in the second bound, we have used the shorthand $\varsigma = \max_j \left( \| \theta^*_j \|_1 + \| b^*_j \|_1 \right)$ as defined in equation~\eqref{eq:psi}.
With this setup in place, we are now ready to prove both deviation bounds.

\subsubsection{Proof of bound~\eqref{eq:mom1}}

Let us first bound the deviation of the first moment. 
We work with the decomposition
\begin{align*}
\Mhat_1 - M_1 = \frac{2}{n} \sum_{i = 1}^{n/2} \underbrace{ \max(\Theta^* x_i + \bstar) x_i - \E [\max(\Theta^* X + \bstar) X] }_{T^1_i}  + \frac{2}{n} \sum_{i = 1}^{n/2} \underbrace{\epsilon_i x_i}_{T^2_i}.
\end{align*}
By triangle inequality, it suffices to bound the norms of each of the two sums separately. We now use the further decomposition
\begin{align*}
T^1_i &= \underbrace{\max(\Theta^* x_i + \bstar) x_i - \max(\Theta^* z_i + \bstar) z_i}_{P_i} + \underbrace{\max(\Theta^* z_i + \bstar) z_i - \E [\max(\Theta^* z_i + \bstar) z_i ]}_{Q_i} \\
& \qquad +
\underbrace{\E [\max(\Theta^* z_i + \bstar) z_i ] - \E [\max(\Theta^* x_i + \bstar) x_i ]}_{R_i}.
\end{align*}
Since $z_i = x_i$ with probability greater than $1 - 2n^{-12}$, the term $P_i = 0$ on this event. 

Also, for each fixed $j \in [k]$, applying the Hoeffding inequality yields the bound
\begin{align*}
\Pr \paren{ \left| \frac{2}{n} \sum_{i = 1}^{n/2} Q_{i, j} \right| \geq t } \leq 2\exp \paren{ - \frac{nt^2}{8C^2 \varsigma^2 (\log (nk))^2 } }.
\end{align*}
On the other hand, for $j \in [d] \setminus [k]$, we have
\begin{align*}
\left| \frac{2}{n} \sum_{i = 1}^{n/2} Q_{i, j} \right| &\leq \varsigma \frac{2}{n} \sum_{i = 1}^{n/2} z_{i, j} \\
&= \varsigma \left| \frac{2}{n} \sum_{i = 1}^{n/2} x_{i, j} \right|.
\end{align*}
Standard Gaussian tail bounds then yield
\begin{align*}
\Pr \paren{ \left| \frac{2}{n} \sum_{i = 1}^{n/2} Q_{i, j} \right| \geq \varsigma t \sqrt{\log (nk) } } \leq 2\exp \paren{ - \frac{nt^2}{8} }
\end{align*}
for each $t \geq 0$.
Putting together the pieces with a union bound and choosing constants appropriately, we then have
\begin{align*}
\Pr \paren{ \eucnorm{ \frac{2}{n} \sum_{i = 1}^{n/2} Q_i }^2 \geq \frac{1}{n} \cdot C k \varsigma^2 (\log (nk))^2 + \frac{1}{n} \cdot C' (d-k) \varsigma^2 \log (nk) } &\leq 2d n^{-12}.
\end{align*}

It remains to handle the final terms $\{R_i\}_{i = 1}^n$. Note that when $j \notin [k]$, we have $R_{i, j} = 0$. 
It therefore suffices to bound the various $R_{i, j}$ terms when $j \in [k]$. We have
\begin{align*}
|R_{i, j}| &= |\E [\max(\Theta^* z_i + \bstar) z_{i, j} ] - \E [\max(\Theta^* x_i + \bstar) x_{i, j} \ind{\Espace_i} ] - \E [\max(\Theta^* x_i + \bstar) x_{i, j} \ind{\Espace_i^c} ]| \\
&= | \E [\max(\Theta^* x_i + \bstar) x_{i, j} \ind{\Espace_i^c} ] |
\end{align*}
Expanding this further, we have
\begin{align*}
|R_{i, j}| &\leq \E [\max_{\ell \in [k]} ( |\inprod{\sigstar_\ell}{x_i}| + |\bstar_\ell| ) |x_{i, j}| \ind{\Espace_i^c} ] \\
&\leq \E \left[ |x_{i, j}| \| x_i \|_\infty ( \| \Thetastar \|_{1, \infty} + \| \bstar \|_\infty ) \ind{\Espace_i^c} \right] \\
&= \varsigma \E \left[ |x_{i, j}| \| x_i \|_\infty \ind{\Espace_i^c } \right] \\
&\leq \varsigma \sum_{\ell = 1}^k \E \left[ |x_{i, j}| | x_{i, \ell} | \ind{\Espace_i^c } \right].
\end{align*}



Note that for a pair $(X_1, X_2)$ of i.i.d. random variables, Jensen's inequality yields the bounds
\begin{align*}
\E [| X_1 X_2 | \ind{X_1, X_2 \geq \lambda}] &\leq \E [X_1^2 \ind{|X_1| \geq \lambda}] \text{ for all } \lambda \geq 0, \text{ and } \\
\E [|X_1| \ind{|X_1| \geq \lambda}] &\leq \E [X_1^2 \ind{|X_1| \geq \lambda}] \text{ for all } \lambda \geq 1.
\end{align*}
Furthermore, if $X$ is a standard Gaussian random variable, then a simple calculation (see also Burkardt~\cite{burkardt2014truncated}) yields the bound
\begin{align*}
\E [ X^2 \; | \; | X | \geq \lambda ] &\leq \frac{1}{2 \sqrt{2\pi}} \lambda e^{- \lambda^2 / 2}, \text{ for all } \lambda \geq \sqrt{2}.
\end{align*}

Putting together the pieces with with $\lambda = 5 \sqrt{\log (2nk)}$, we have
\begin{align*}
|R_{i, j} |^2 &\leq C k^2 \varsigma^2 \log (nk) (nk)^{-24}, 
\end{align*}
and summing over $j \in [k]$ yields the bound
\begin{align*}
\eucnorm{ \frac{2}{n} \sum_{i = 1}^{n/2} R_i }^2 \leq C k^2 \varsigma^2 \log(nk) (nk)^{-24}.
\end{align*}

Finally, putting together the pieces with a union bound yields the desired bound on the random variable $\eucnorm{ \frac{2}{n} \sum_{i = 1}^{n/2} T^1_{i}} $.

The second term can be bounded more easily; in particular, on the intersection of the events $\paren{\Fspace_i}_{i=1}^n$, we have
\begin{align*}
\eucnorm{ \frac{2}{n} \sum_{i=1}^{n/2} T^2_i}^2 \leq C \sigma^2 \log n \eucnorm{ \frac{2}{n} \sum_{i = 1}^{n/2} x_i }^2
\leq C \sigma^2 \frac{(d+ \log n) \log n}{n},
\end{align*}
where the final bound holds with probability greater than $1 - cn^{-10}$. Finally, putting the bounds together yields the result.

\subsubsection{Proof of bound~\eqref{eq:mom2}}
 
Once again, we decompose the required term as
\begin{align*}
\Mhat_2 - M_2 &= \frac{2}{n} \sum_{i = 1}^n \underbrace{ \max(\Theta^* x_i + \bstar) \left( x_i x_i^\top - I_d \right) }_{\tau^1_i} + \frac{2}{n} \sum_{i = 1}^{n/2} \underbrace{\epsilon_i \left( x_i x_i^\top - I_d \right)}_{\tau^2_i}.
\end{align*}
We use the further decomposition
\begin{align*}
\tau^1_i &= \underbrace{\max(\Theta^* x_i + \bstar) \left( x_i x_i^\top - I_d \right) - \max(\Theta^* z_i + \bstar) \left( z_i z_i^\top - I_d \right) }_{\phi_i} \\
& \qquad + \underbrace{\max(\Theta^* z_i + \bstar) \left( z_i z_i^\top - I_d \right) - \E [\max(\Theta^* z_i + \bstar) \left( z_i z_i^\top - I_d \right) ]}_{\kappa_i} \\
& \qquad +
\underbrace{\E [\max(\Theta^* z_i + \bstar) \left( z_i z_i^\top - I_d \right) ] - \E [\max(\Theta^* x_i + \bstar) \left( x_i x_i^\top - I_d \right) ]}_{\rho_i}.
\end{align*}
As before, since $z_i = x_i$ with probability greater than $1 - 2n^{-12}$, the term $\phi_i = 0$ on this event.

Let us further decompose $\kappa_i$ as
\begin{align*}
\kappa_i &= 
\underbrace{\left(\max(\Theta^* z_i + \bstar) + \varsigma \sqrt{\log (nk) } \right) z_i z_i^\top -\mathbb{E} \bigg [\left(\max(\Theta^* z_i + \bstar) + \varsigma \sqrt{\log (nk) } \right) z_i z_i^\top \bigg]}_{\kappa_i^{(1)}} \\
&\qquad \varsigma \sqrt{\log (nk) } \underbrace{ \mathbb{E}\left[  z_i z_i^\top \right] -   I_d }_{\kappa_i^{(2)}}
+ I_d \; \cdot \; \underbrace{\left( \E [\max(\Theta^* z_i + \bstar) - \max(\Theta^* z_i + \bstar) \right)}_{\kappa^{(3)}_i},
\end{align*}
so that 
\begin{align*}
\opnorm{ \frac{2}{n} \sum_{i = 1}^n \kappa_i} \leq \opnorm{\frac{2}{n} \sum_{i = 1}^n \kappa^{(1)}_i } + \opnorm{\frac{2}{n} \sum_{i = 1}^n \kappa^{(2)}_i } + \left| \frac{2}{n} \sum_{i = 1}^{n / 2} \kappa^{(3)}_i \right|.
\end{align*}
Since $|\max(\Theta^* z_i + \bstar)|\leq C \varsigma \sqrt{\log (nk) } $, the random vector $\sqrt{ \max(\Theta^* z_i + \bstar) + C \varsigma \sqrt{\log (nk) } } z_i$  is well-defined and bounded; sub-Gaussian concentration bounds~\cite{wainwright2019high} can therefore be applied to obtain
\begin{align*}
\Prob \bigg [ \opnorm{ \frac{1}{n} \sum_{i=1}^{n} \kappa_i^{(1)} } \geq c_1 \varsigma^2 (\log (nk))^2 \left \lbrace \sqrt{\frac{\numdim}{\numobs}} + \frac{\numdim}{\numobs} + \delta \right \rbrace \bigg ] \leq c_2 \exp \left( - n \min(\delta,\delta^2) \right)
\end{align*}
where $\varsigma_1 \log (nk) = \max(\Theta^* z_i + \bstar) + \varsigma \sqrt{\log (nk) } \leq 2 \varsigma \log(nk) $. Reasoning similarly for the second term, we have
\begin{align*}
\Prob \bigg [ \opnorm{ \frac{1}{\numobs} \sum_{i=1}^{n} \kappa_i^{(2)} } \geq c_1 \varsigma^2 (\log (nk))^2 \left \lbrace \sqrt{\frac{\numdim}{\numobs}} + \frac{\numdim}{\numobs} + \delta \right \rbrace \bigg ] \leq c_2 \exp \left( - n \min(\delta,\delta^2) \right).
\end{align*}
Combining these bounds setting $\delta = c_1 \sqrt{\frac{\numdim}{\numobs}}$, we have
\begin{align*}
\opnorm{ \frac{2}{\numobs} \sum_{i=1}^{\numobs} \kappa_i^{(1)} } + \opnorm{ \frac{2}{\numobs} \sum_{i=1}^{\numobs} \kappa_i^{(2)} } \leq C \varsigma^2 (\log (nk))^2 \left \lbrace \sqrt{\frac{\numdim}{\numobs}} + \frac{\numdim}{\numobs} \right \rbrace
\end{align*} 
with probability at least $1- c \exp \left( -c' d\right)$.

The term $\kappa_i^{(3)}$, on the other hand, can be controlled directly via Hoeffding's inequality. Since $\max(\Theta^* z_i + \bstar)$ is $C \varsigma \sqrt{\log (nk) }$ sub-Gaussian, we obtain
\begin{align*}
\Prob \left[ \left | \frac{2}{n}\sum_{i=1}^{n/2}\kappa_i^{(3)} \right | \geq \varsigma \sqrt{\log (nk) } t \right ] \leq 2 \exp \left\lbrace -\frac{n t^2}{32} \right\rbrace.
\end{align*}
Choosing $t = c \sqrt{\frac{\numdim + \log n}{\numobs}}$ and putting together all the pieces, we obtain
\begin{align*}
\opnorm{ \frac{2}{n} \sum_{i = 1}^n \kappa_i} \leq C \varsigma^2 (\log (nk))^2 \left  \lbrace \sqrt{\frac{\numdim + \log n}{\numobs}} + \frac{\numdim + \log n}{\numobs} \right \rbrace + c \varsigma \sqrt{\log (nk) } \sqrt{\frac{\numdim}{\numobs}}
\end{align*}
with probability at least $1- c n^{-12}$.

It remains to handle the terms $\lbrace \rho_i \rbrace_{i=1}^{n/2}$, and to do so, we use a similar argument to before. We first bound the absolute value of the $(p,q)$th entry of each matrix as
\begin{align*}
|\rho_i(p,q)| &= |\E [\max(\Theta^* z_i + \bstar) z_i z_i^\top (p,q) ] - \E [\max(\Theta^* x_i + \bstar) x_i x_i^\top (p,q) \ind{\Espace_i} ] \\ &- \E [\max(\Theta^* x_i + \bstar) x_i x_i^\top(p,q) \ind{\Espace_i^c} ]| = | \E [\max(\Theta^* x_i + \bstar) x_i x_i^\top (p,q) \ind{\Espace_i^c} ] |
\end{align*}
Expanding this further, we have
\begin{align*}
|\rho_i (p,q)| &\leq \E [\max_{\ell \in [k]} ( |\inprod{\sigstar_\ell}{x_i}| + |\bstar_\ell| ) | x_{i, p} x_{i, q} | \ind{\Espace_i^c} ] \\
&\leq \varsigma \E \left[ | x_{i, p} x_{i, q} | \| x_i \|_\infty \ind{\Espace_i^c} \right] \\
&\leq \E \left[ | x_{i, p} x_{i, q} | \sum_{\ell \in [k]} | x_{i, \ell} | \ind{\Espace_i^c } \right].
\end{align*}

Also note that $\rho_{p , q} = 0$ unless $p \in [k], q \in [k]$. Hence we finally need to control the terms of the form $\mathbb{E}\left[ |X|^3 \ind{|X| \geq \lambda} \right]$ for a standard Gaussian $X$. Substituting $\lambda =5 \sqrt{\log (nk)}$, a simple calculation of truncated third moment of standard Gaussian (\cite{burkardt2014truncated}) yields 
\begin{align*}
|\rho_i(p,q)| &\leq \varsigma \log^2(nk) (nk)^{-10},
\end{align*}
and proceeding as before provides a strictly lower order bound on $\opnorm{ \rho_i }$ than the remaining terms.




The term $\tau_i^2$ can be bounded more easily. Specifically, on the intersection of the events $\{\mathcal{F}_i \}_{i=1}^{n/2}$, applying \cite[Lemma 6.2]{wainwright2019high}, we have
\begin{align*}
\opnorm{ \frac{2}{n} \sum_{i=1}^{n/2} \tau^2_i}^2 \leq C \sigma^2 \log n \opnorm{ \frac{2}{n} \sum_{i = 1}^{n/2} x_i x_i^\top -I }^2
\leq C \sigma^2 \log n \left\lbrace \frac{d+\log n}{n}+ \frac{(d+\log n)^2}{n^2} \right \rbrace
\end{align*}
where the final bound holds with probability greater than $1 - c n^{-12}$. Finally combining all the terms yield the desired result.
\qed


\section{Background and technical lemmas used in the proof of Theorem~\ref{thm:rs}} \label{app:thm4}

In this section, we collect two technical lemmas that were used to prove Theorem~\ref{thm:rs}.

\subsection{Prediction and estimation error} \label{sec:pred-est}

Here, we connect the prediction error to the estimation error when the covariates are Gaussian, which may be of independent interest. Recall our notation $\dist$ for the minimum distance between parameters obtainable after relabeling.
\begin{lemma}
\label{lem:pred_est_gen}
Suppose that the covariates $\{ \meas_i \}_{i=1}^\numobs$ are drawn i.i.d from the standard Gaussian distribution $\NORMAL(0,I_d)$ and that the true parameters $\paren{\appsigstar_j}_{j = 1}^k$ are fixed. Then, there exists a tuple of universal constants $(c_1, c_2)$ such that simultaneously for all parameters $\beta_1, \ldots, \beta_k \in \mathsf{B_{vol}}(\pi_{\min},\Delta,\kappa)$:
\begin{enumerate}
\item If $n \geq c_1 d$, then we have 
\begin{align*}
\frac{1}{n} \sum_{i =  1}^n \left( \max_{j \in [k]} \inprod{\appmeas_i}{\beta_j} - \max_{j \in [k]} \inprod{\appmeas_i}{\appsigstar_j} \right)^2 & \leq c_1 \dist( \paren{\beta_j}_{j = 1}^k, \paren{\appsigstar}_{j = 1}^k )
\end{align*}
with probability exceeding $1- c_1 \exp(-c_2 \numobs)$.
\item If $n \geq  c_1  d\frac{k}{\pi_{\min}}$, then we have
\begin{align*}
c_2 \log^{-1}(k/\pi_{\min}) \left(\frac{\pi_{\min}}{k}\right)^5  \sum_{j \in [k]} \min_{j' \in [k]} \; \| \appsigstar_j - \beta_{j'} \|^2 \leq
\frac{1}{n} \sum_{i =  1}^n \left( \max_{j \in [k]} \inprod{\appmeas_i}{\beta_j} - \max_{j \in [k]} \inprod{\appmeas_i}{\appsigstar_j} \right)^2 
\end{align*}
with probability exceeding $1- c_1 k \exp(-c_2 n)$.
\end{enumerate}
\end{lemma}

\begin{proof}
 To prove the part 1 of the lemma, we leverage the fact that the $ \max $ function is $1$-Lipschitz with respect to the $\ell_2$-norm. Consequently, we obtain
 \begin{align*}
 \frac{1}{n} \sum_{i =  1}^n \left( \max_{j \in [k]} \inprod{\appmeas_i}{\beta_j} - \max_{j \in [k]} \inprod{\appmeas_i}{\appsigstar_j} \right)^2 & \leq \frac{1}{n} \sum_{i =  1}^n \sum_{j=1}^k \left(\xi_i^\top(\beta_j - \beta^*_j) \right)^2,
 \end{align*}
 where we have ordered the parameters such that $\dist \left(\paren{\beta_j}_{j = 1}^k, \paren{\appsigstar_j}_{j = 1}^k \right)$ is minimized.
We now use the fact that the rows of $\AppXmat$ are $1$-sub-Gaussian (this is restatement of the conclusion of Lemma~\ref{lem:sgmax}) to complete the proof.
 
 We now proceed to a proof of part 2 of the lemma.
 Recall the setup of Appendix~\ref{sec:pf-thm3} along with notation $\left( \{\meas_i \}_{i=1}^\numobs, \Thetastar, b^*, \beta^*,  \right)$. Specifically, we have $\beta^*_j = \left( \theta^*_j,\,\, b^*_j \right)$ and $(\Theta^*)^\top = [\theta^*_1 \,\, \theta^*_2 \ldots \theta^*_k]$. Similarly let $\beta_j = (\theta_j , \,\, b_j) \in \real^{d+1}$ and $\Theta^\top = [\theta_1 \,\, \theta_2 \ldots \theta_k]$.
 In the notation of Section~\ref{sec:pf-thm1}, we define for each pair $(\Theta,b)$, the sets 
\begin{align*}
S_j(\Theta, b) &= \left \{ i \in [\numobs]: \langle \meas_i, \theta_j \rangle +b_j = \max_{j' \in [k]} (\langle \meas_i, \theta_{j'} \rangle + b_{j'})  \right \}, \,\,\, j \in [k].
\end{align*}
We use the shorthand $S^*_j = S_j(\Thetastar,b^*)$ and $\widehat{S}_j = S_j(\Theta , b)$ for the rest of the proof.  By definition, we have
\begin{align*}
\frac{1}{n}\sum_{i=1}^{n} \left ( \max(\Theta \meas_i + b) - \max(\Thetastar \meas_i + b^*) \right )^2 &= \frac{1}{n}\sum_{ \substack{ \ell \in [k] \\ m \in [k]}} \sum_{i \in S^*_{\ell} \cap \widehat{S}_m }  \bigg ( (\inprod{\thetastar_{\ell}}{\meas_i}+b^*_l) - (\inprod{\theta_m}{\meas_i} + b_m) \bigg )^2 \\
&= \frac{1}{n}\sum_{ \substack{ \ell \in [k] \\ m \in [k]}} \sum_{i \in S^*_{\ell} \cap \widehat{S}_m }  \bigg ( \inprod{\beta^*_{\ell}}{\appmeas_i} - \inprod{\beta_m}{\appmeas_i} \bigg )^2 \\
&= \frac{1}{n}\sum_{ \substack{ \ell \in [k] \\ m \in [k]}} \| \widetilde{\Xi}_{\ell, m} (\beta^*_\ell - \beta_m ) \|^2,
\end{align*}
where we have let $\widetilde{\Xi}_{\ell, m}$ denote the sub-matrix of $\Xi$ with rows indexed by the set $S^*_{\ell} \cap \widehat{S}_{m}$. Applying the
Hoeffding bound to $|S^*_{\ell}|$ yields $\Prob(|S^*_{\ell}| \leq \frac{1}{3} \pi_{\min} n ) \leq \exp(-cn)$ for each $\ell \in [k]$. Furthermore, for each $\ell \in [k]$, there exists a corresponding index $m_\ell$ such that $|S^*_{\ell} \cap \widehat{S}_{m_{\ell}}| \geq \frac{1}{k}|S^*_{\ell}|$. In conjunction with the high probability bound on $|S^*_{\ell}|$, we obtain
\begin{align*}
|S^*_{\ell} \cap \widehat{S}_{m_{\ell}}| \geq  \frac{\pi_{\min}}{3k} n,
\end{align*}
with probability exceeding $1 - \exp(-cn)$. Putting together the pieces, we obtain the bound
\begin{align*}
\frac{1}{n}\sum_{i=1}^{n} \left ( \max(\Theta \meas_i + b) - \max(\Thetastar \meas_i + b^*) \right )^2 &\geq \frac{1}{n} \sum_{ \ell \in [k]} \norms{ \widetilde{\Xi}_{\ell, m_{\ell}} (\beta^*_\ell - \beta_{m_\ell})}^2 \\
&\geq \frac{1}{n}\sum_{\ell \in [k]} \lambda_{\min} \left (\widetilde{\Xi}_{\ell, m_\ell}^\top \widetilde{\Xi}_{\ell, m_\ell} \right ) \norms{\beta^*_\ell - \beta_{m_\ell}}^2.
\end{align*}
We now claim that the bound
\begin{align}
\max_{\ell \in [k]} \,\, \lambda_{\min} \left (\widetilde{\Xi}_{\ell, m_\ell}^\top \widetilde{\Xi}_{\ell, m_\ell} \right ) \geq C \frac{1}{1+\log(k/\pi_{\min})}\left(\frac{\pi_{\min}}{k}\right)^5 n \label{eq:min-sing-val}
\end{align}
holds with probability exceeding $1- c k \exp(-c_1 n)$ provided $n \geq C \frac{d k}{\pi_{\min}}$. Taking this claim as given for the moment, we have
\begin{align*}
\frac{1}{n}\sum_{i=1}^{n} \left ( \max(\Theta \meas_i + b) - \max(\Thetastar \meas_i + b^*) \right )^2 & \geq C \frac{1}{1+\log(k/\pi_{\min})}\left(\frac{\pi_{\min}}{k}\right)^5  \sum_{\ell = 1}^k \norms{\beta^*_\ell - \beta_{m_\ell}}^2 \\
& \geq C \log^{-1}(k/\pi_{\min}) \left(\frac{\pi_{\min}}{k}\right)^5  \sum_{j \in [k]} \min_{j' \in [k]} \; \| \appsigstar_j - \beta_{j'} \|^2,
\end{align*}
thereby proving the second part of the lemma. It remains to establish claim~\eqref{eq:min-sing-val}.

\paragraph{Proof of claim~\eqref{eq:min-sing-val}:}
We use Lemma~\ref{lem:minsingvol} to prove a lower bound on $
 \lambda_{\min} \left (\widetilde{\Xi}_{\ell, m_\ell}^\top \widetilde{\Xi}_{\ell, m_\ell} \right )$.
Since the covariates $\{ x_i \}_{i=1}^n$ are drawn from the standard $d$ dimensional Gaussian distribution, Assumption~\ref{assptn:gen} is satisfied with $(\eta, \sball, \cs)= (1,1/2,e)$. Also, the inequality 
\begin{align*}
|S^*_{\ell} \cap \widehat{S}_{m_{\ell}}| \geq  \frac{\pi_{\min}}{3k} n,
\end{align*}
holds with probability at least $1-c \exp(-c_1 n)$. Hence, on the event where Lemma~\ref{lem:minsingvol} holds, we obtain
\begin{align*}
\lambda_{\min} \left (\widetilde{\Xi}_{\ell, m_\ell}^\top \widetilde{\Xi}_{\ell, m_\ell} \right ) \geq C \frac{1}{1+\log(k/\pi_{\min})}\left(\frac{\pi_{\min}}{k}\right)^5 n,
\end{align*}
with probability exceeding $1-c_1 \exp (-c_2 n)$ provided $n \geq C_1 \frac{kd}{\pi_{\min}}$. Finally, taking a union bound over possible values of $\ell$ yields the claim.
\end{proof}

\subsection{Projection onto a finite collection of rays} \label{sec:rays}

Consider a vector $\thetastar \in \real^n$ observed via the observation model
\begin{align*}
y = \thetastar + \epsilon,
\end{align*}
where $\epsilon$ has independent, zero-mean, $\sigma$-sub-Gaussian entries.
For a fixed set of $M$ vectors $\{ \theta_1, \ldots, \theta_M\}$, denote by
$\mathbb{C} \defn \{ c \theta_\ell: c \geq 0, \ell \in [M]\}$ the set of all one-sided rays obtainable with these vectors.

Now consider the projection estimate
\begin{align*}
P_{\mathbb{C}}(y) = \argmin_{\theta \in \mathbb{C}} \| y - \theta \|^2,
\end{align*}
which exists since the projection onto each ray exists. 
The following lemma proves an oracle inequality on the error of such an estimate.
\begin{lemma} \label{lem:rays}
There are universal constants $c$, $C$, $c_1$ and $c_2$ such that 
\begin{align*}
\Pr \paren{ \| P_{\mathbb{C}}(y) - \thetastar \|^2 \geq c \left( \min_{\theta \in \mathbb{C}}\|\theta-\theta^*\|^2 +  \sigma^2 t(\log M + c_1) \right) } \leq c_2 e^{-nt(\sqrt{\log M}+c_1)},
\end{align*}
for all $t \geq C \sigma ( \sqrt{\log M} +c_1)$.
\end{lemma}
\begin{proof}
We follow the standard technique for bounding the error for non-parametric least squares estimators. From the definition, we have
\begin{align*}
P_{\mathbb{C}}(y)=\argmin_{\theta \in \mathbb{C}}\norms{y-\theta}^2.
\end{align*}
We substitute the expression for $y$ and obtain
\begin{align*}
P_{\mathbb{C}}(y)= \argmax_{\theta \in \mathbb{C}} \left[ 2 \langle\epsilon, \theta - \theta^* \rangle - \norms{\theta - \theta^*}^2 \right].
\end{align*}
 To obtain an upper bound on $\|P_{\mathbb{C}}(y) -\theta^* \|^2$, it is sufficient to control the following quantity (e.g. see \cite[Chapter 3]{van_m_est}, \cite[Chapter 13]{wainwright2019high}):
\begin{align*}
 \mathbb{E}\left[ \sup_{\theta \in \mathbb{C}: \|\theta -\theta^*\| \leq \delta} \langle \epsilon, \theta - \theta^* \rangle \right]
\end{align*}
for some $\delta >0$ to be chosen later. Since $\epsilon$ is $\sigma$-sub-Gaussian, we use Dudley's entropy integral to control the term above. We obtain
\begin{align*}
\mathbb{E} \left [ \sup_{\theta \in \mathbb{C}: \| \theta -\theta^*\| \leq \delta} \langle \epsilon, \theta - \theta^* \rangle \right ] \leq C \sigma \int_{0}^\delta \sqrt{ \log N \left(\varepsilon, \{\theta \in \mathbb{C}, \|\theta - \theta^*\| \leq \delta \}, \ell_2 \right)} d\varepsilon,
\end{align*}
where $N(\epsilon, S, \ell_2)$ is the $\epsilon$-covering number of a compact set $S$ in $\ell_2$ norm. Note that $\mathbb{C}$ contains scaled versions of $M$ fixed vectors $\{\theta_1,\ldots,\theta_M \}$. For a fixed $\theta_i$, with $i \in [M]$, the covering number $N\left(\varepsilon, \{c\theta_i : c \in \real,\|\theta_i-\theta^*\|\leq \delta\}, \ell_2 \right)$ is equivalent to the covering number of a bounded interval (in $1$ dimension). Using \cite{vershynin2018high}, this is $(1+\frac{2\delta}{\varepsilon})$. Since there are $M$ such fixed vectors, we obtain
\begin{align*}
N\left(\varepsilon, \{\theta \in \mathbb{C}, \|\theta - \theta^*\| \leq \delta \}, \ell_2 \right)  \leq C_1 M (1+\frac{\delta}{\varepsilon}).
\end{align*}
Substituting, we obtain
\begin{align*}
\mathbb{E} \left [ \sup_{\theta \in \mathbb{C}: \| \theta -\theta^*\| \leq \delta} \langle \epsilon, \theta - \theta^* \rangle \right ] \leq C \sigma \left( \delta \sqrt{\log M} + C_1 \delta \right).
\end{align*}
Now, the critical inequality (\cite[Chapter 13]{wainwright2019high}) takes the form
\begin{align*}
\delta \sigma (\sqrt{\log M} + C_1) \lesssim \delta^2.
\end{align*}
Hence we can choose $\delta = C_2 \sigma ( \sqrt{\log M} +C_1) $. Now, for any $t \geq \delta$, invoking \cite[Theorem 13.2]{wainwright2019high} yields the oracle inequality
\begin{align*}
\|P_\mathbb{C}(y) - \theta^* \|^2 \leq c \left(\|\theta^* - P_\mathbb{C}(\theta^*) \|^2 +  \sigma^2 t (\log M +c_1)\right) = c \left( \min_{\theta \in \mathbb{C}}\|\theta-\theta^*\|^2 +  \sigma^2 t(\log M + c_1) \right),
\end{align*}
with probability exceeding $1-c_2 e^{-nt(\sqrt{\log M}+c_1)}$, which proves the lemma.
\end{proof}


\section{Numerical experiments: Noiseless Sample Complexity} \label{app:sim}

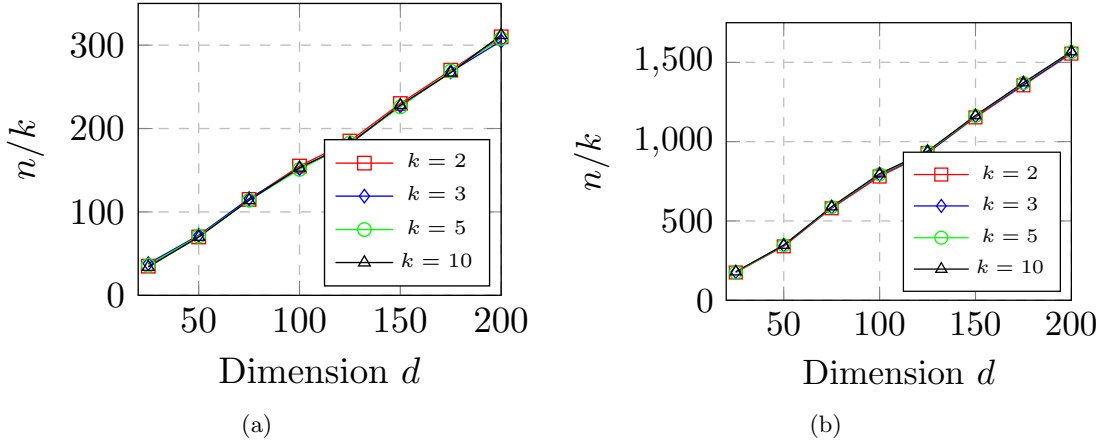
\begin{figure*}[t!]
\centering
\subfigure[]{\resizebox {0.45 \textwidth} {!}{
\begin{tikzpicture}
\begin{axis}[
    legend style={font=\tiny},
    xlabel={Dimension $d$},
    ylabel={$n/k$},
    xmin=20, xmax=200,
    ymin=0, ymax=350,
    legend pos=south east,
    ymajorgrids=true,
    xmajorgrids=true,
    grid style=dashed,
]
 
\addplot[
    color=red,
    mark=square,
    ]
    coordinates {
    (25,35)(50,70)(75,115)(100,155)(125,185)(150,230)(175,270)(200,310)
    };
    \addlegendentry{$k=2$}
    
\addplot[
    color=blue,
    mark=diamond,
    ]
    coordinates {
    (25,38)(50,72)(75,116)(100,152)(125,182)(150,227)(175,268)(200,305)
    };
    \addlegendentry{$k=3$}

\addplot[
    color=green,
    mark=o,
    ]
    coordinates {
    (25,36)(50,71)(75,114)(100,151)(125,183)(150,226)(175,269)(200,308)
    };
    \addlegendentry{$k=5$}
 
 \addplot[
    color=black,
    mark=triangle,
    ]
    coordinates {
    (25,34)(50,70)(75,114)(100,153)(125,182)(150,228)(175,267)(200,312)
    };
    \addlegendentry{$k=10$}
 
\end{axis}
\end{tikzpicture}}}
\subfigure[]{\resizebox {0.45 \textwidth} {!}{
\begin{tikzpicture}
\begin{axis}[
    legend style={font=\tiny},
    xlabel={Dimension $d$},
    ylabel={$n/k$},
    xmin=20, xmax=200,
    ymin=0, ymax=1750,
    legend pos=south east,
    ymajorgrids=true,
    xmajorgrids=true,
    grid style=dashed,
]
 
\addplot[
    color=red,
    mark=square,
    ]
    coordinates {
    (25,175)(50,340)(75,580)(100,780)(125,928)(150,1153)(175,1355)(200,1555)
    };
    \addlegendentry{$k=2$}
    
\addplot[
    color=blue,
    mark=diamond,
    ]
    coordinates {
    (25,178)(50,345)(75,585)(100,790)(125,930)(150,1160)(175,1360)(200,1560)
    };
    \addlegendentry{$k=3$}

\addplot[
    color=green,
    mark=o,
    ]
    coordinates {
    (25,175)(50,347)(75,587)(100,794)(125,935)(150,1162)(175,1369)(200,1565)
    };
    \addlegendentry{$k=5$}
 
 \addplot[
    color=black,
    mark=triangle,
    ]
    coordinates {
    (25,180)(50,348)(75,590)(100,798)(125,940)(150,1165)(175,1372)(200,1568)
    };
    \addlegendentry{$k=10$}
 
\end{axis}
\end{tikzpicture}}}
\caption{Convergence of the alternating minimization in the noiseless setting. Panel (a) shows the sample complexity (number of samples required for exact recovery of the parameters) for Gaussian covariates, when the initial iterates are chosen using a perturbed initialization. In panel (b), we plot the sample complexity when the covariates are drawn i.i.d from $\mathsf{Unif}[-\sqrt{3},\sqrt{3}]^{\otimes d}$. In both plots, that the sample complexity scales linearly with $k d$.}
\label{fig:conv_good_sample}
\end{figure*}

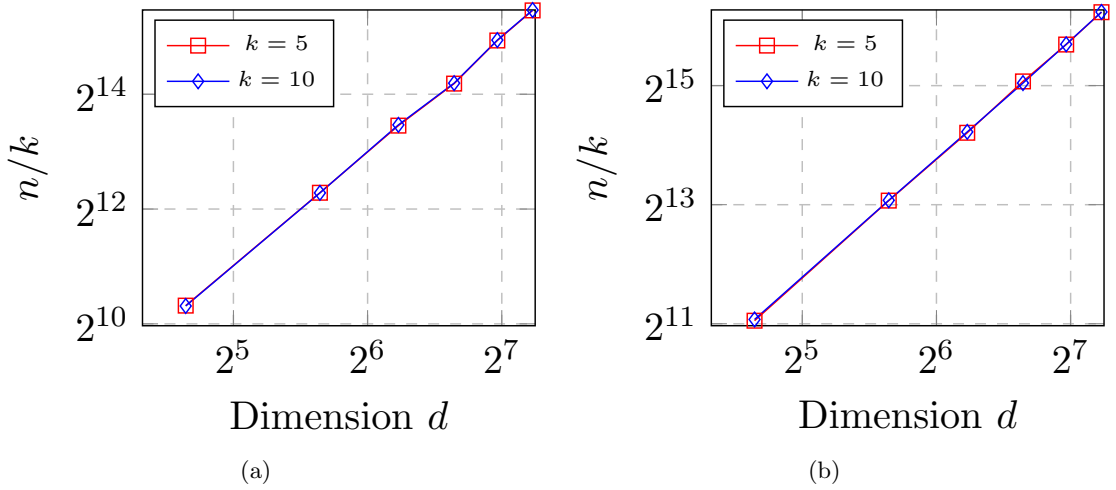
\begin{figure*}[t]
\centering
\subfigure[]{\resizebox {0.45 \textwidth} {!}{
\begin{tikzpicture}
\begin{loglogaxis}[
    log basis x=2,
    log basis y=2,
    legend style={font=\tiny},
    xlabel={Dimension $d$},
    ylabel={$n/k$},
    xmin=20, xmax=152,
    ymin=1000, ymax=45250,
    legend pos=north west,
    ymajorgrids=true,
    xmajorgrids=true,
    grid style=dashed,
]
 
\addplot[
    color=red,
    mark=square,
    ]
    coordinates {
    (25,1275)(50,4980)(75,11200)(100,18600)(125,31300)(150,45000)
    };
    \addlegendentry{$k=5$}
    
\addplot[
    color=blue,
    mark=diamond,
    ]
    coordinates {
    (25,1269)(50,4972)(75,11300)(100,18700)(125,31450)(150,45200)
    };
    \addlegendentry{$k=10$} 
    
 
\end{loglogaxis}
\end{tikzpicture}}}
\subfigure[]{\resizebox {0.45 \textwidth} {!}{
\begin{tikzpicture}
\begin{loglogaxis}[
    log basis x=2,
    log basis y=2,
    legend style={font=\tiny},
    xlabel={Dimension $d$},
    ylabel={$n/k$},
    xmin=20, xmax=152,
    ymin=2000, ymax=79000,
    legend pos=north west,
    ymajorgrids=true,
    xmajorgrids=true,
    grid style=dashed,
]
 
\addplot[
    color=red,
    mark=square,
    ]
    coordinates {
    (25,2120)(50,8580)(75,18900)(100,34370)(125,52790)(150,76930)
    };
    \addlegendentry{$k=5$}
    
\addplot[
    color=blue,
    mark=diamond,
    ]
    coordinates {
    (25,2151)(50,8612)(75,19100)(100,33698)(125,52940)(150,77120)
    };
    \addlegendentry{$k=10$} 
    
 
\end{loglogaxis}
\end{tikzpicture}}}
\caption{AM with Rademacher and $\frac{1}{\sqrt{2.4}}\left(\mathsf{Bin}(10,0.4)-4\right)$ covariates in log-log scale--- the best-fit line to plot (a) has slope $1.94$ and the best-fit line to plot (b) has slope $1.96$, and hence the sample complexity scales linearly with $k$ but quadratically with dimension $d$.}
\label{fig:rademacher}
\end{figure*}

In this section, we present some numerical experiments to illustrate the noiseless sample complexity of the AM algorithm under different covariate assumptions. We fix $\norms{\appsigstar_i}=1$ for all $i \in [k]$, $\sigma =0$ and $\pi_{\min}=1/k$. In particular, we choose $\{\beta^*_i\}_{i=1}^k$ to be the standard $k$-dimensional basis. In this noise-free setting, we say $\appsigstar_i$ is recovered if $\norms{\appsig{t}_i-\appsigstar_i} \leq 0.01$. For a fixed dimension $\numdim$, we run a linear search on the number of samples $\numobs$, such that the empirical probability of success over $100$ trials is more than $0.95$, and output the least such $\numobs$. Panel (a) of Figure~\ref{fig:conv_good_sample} shows the variation of normalized sample complexity (normalized by a factor $k$) for Gaussian covariates with respect to dimension. We observe that the sample complexity obeys $n \propto kd$. In panel (b), we sample the covariates from $\mathsf{Unif}[-\sqrt{3},\sqrt{3}]^{\otimes d}$, which, as discussed in Section~\ref{sec:mainresults}, satisfy the small-ball condition. Here also we observe that $n \propto kd$ for perfect recovery of $\{\beta^*_i\}_{i=1}^k$.

 For the distributions which does not satisfy the small ball condition, the sample complexity for perfect parameter recovery has a quadratic dependence on dimension $d$. In particular, we draw the covariates i.i.d from Rademacher and $\frac{1}{\sqrt{2.4}}\left(\mathsf{Bin}(10,0.4)-4\right)$ distributions. In Figure~\ref{fig:rademacher}, we plot the noiseless sample complexity and dimension on a log scale. Since the slope of the (best-fit) lines in Figure~\ref{fig:rademacher} is very close to $2$, we conclude that the number of samples required for perfect recovery of the parameters is linear in $k$, but quadratic in $d$.



\section{Initialization via PCA and least squares} \label{app:least-squares-init}

We now provide an alternative to the random search algorithm (Algorithm~\ref{algo:rs}) of Section~\ref{sec:setup}. Recall that this is the second step of the initialization algorithm, which solves a full-blown least squares problem for max-affine regression, the details of which is described in Algorithm~\ref{algo:lsk}. We analyze the least squares problem and obtain the following guarantee.
\begin{algorithm}[h] 
\KwIn{Data $\paren{ \appmeas_i, y_i }_{i=n_1}^{n_2}$ and subspace estimate $\Uhat$ formed independently of the data.}

\KwOut{Initial estimator $\paren{ \appsig{0}_j }_{j = 1}^k$.}

\nl Form the tilted, $k$-dimensional covariates $\widetilde{\meas}_i = \widehat{U}^\top \meas_i$ for each $i = n_1 + 1, \ldots, n_2$, and compute the least squares estimator in $k$ dimensions
\begin{align}
\paren{ \widehat{\alpha}_j , \widehat{b}_j }_{j = 1}^k = \argmin_{ \substack{\alpha_1, \ldots, \alpha_k \in \real^k \\ b_1, \ldots, b_k \in \real}  } \; \sum_{i = n_1+1}^{n_2} \left( y_i - \max_{j \in [k]} \; \left( \inprod{\widetilde{\meas}_i}{\alpha_j} + b_j \right) \right)^2. \label{eq:kopt}
\end{align}

\nl Return the $(d+1)$-dimensional parameters $\appsig{0}_j = \left( \widehat{U} \widehat{\alpha}_j \; \; \widehat{b}_j \right)$ for each $j \in [k]$.
\caption{Low-dimensional least squares\label{algo:lsk}}
\end{algorithm}
\begin{theorem}
\label{thm:init}
Suppose that the covariates $\meas_i$ are drawn i.i.d. from a standard Gaussian distribution, and that $\numobs \geq C \sigma^2 \frac{k^2}{\pi_{\min}}$. Then the parameter estimates $\paren{ \appsig{0}_j }_{j=1}^{k}$ returned by Algorithm~\ref{algo:lsk} satisfy  
\begin{align*}
\dist \left( \paren{\appsig{0}_j }_{j=1}^{k}, \paren{\appsigstar_j }_{j=1}^{k} \right) \leq C \opnorm{\Thetastar}^2 \,\, \fronorm{\Uhat \Uhat^\top - \Ustar (\Ustar)^\top}^2 \frac{k^5 \log (nk) \log(k/\pi_{\min})}{\pi_{\min}^5},
\end{align*}
with probability exceeding $1- c n^{-10} - c' \exp(-c_1 \numobs)$.
\end{theorem}
From Theorem~\ref{thm:pca}, we have
\begin{align*}
\fronorm{\Uhat \Uhat^\top - \Ustar (\Ustar)^\top}^2 \leq C \left (\frac{\sigma^2 + \varsigma^2 }{\gamma^2} \right ) \frac{k d \log^3 (nk) }{n},
\end{align*}
with probability greater than $1 - C n^{-10}$. Hence, in conjunction with Theorem~\ref{thm:init}, we obtain the final guarantee
\begin{align*}
\dist \left( \paren{\appsig{0}_j }_{j=1}^{k}, \paren{\appsigstar_j }_{j=1}^{k} \right) \leq C  \opnorm{\Thetastar}^2 \left (\frac{\sigma^2 + \varsigma^2 }{\gamma^2} \right ) \frac{k^6 d \log^4 (nk) \log(k/\pi_{\min})}{\pi_{\min}^5 n},
\end{align*}
with probability at least $1- c n^{-10} - c' \exp(-c_1 \numobs)$. This resembles the final guarantee of the combined PCA and random search based procedure, but notably, the algorithm above is not tractably implementable to the best of our knowledge.

While this least squares algorithm has been analyzed in the past~\cite{guntuboyina2012optimal}, two crucial differences specific to our setting are worth pointing out. First, while prior work provides bounds on the prediction error of the least squares estimator, we are interested in providing guarantees on parameter estimation. Second, our problem is solved with errors-in-variables, since we only expect the approximate relation $\Uhat^\top x_i \approx U^\top x_i$ to hold. Consequently, we prove a general result with both of these intricacies, which may be of broader interest.

Let us set up a general problem of this form, using distinct notation from before for clarity. Assume that $n$ covariate-response pair $(\resp_i,\covarstar_i)_{i = 1}^n$ are generated by the max-affine observation model 
\begin{align*}
\resp_i = \max_{j \in [k]} \left( \inprod{\covarstar_i}{\sigstargen_j} + \sigstarint_j \right) + \epsilon_i,
\end{align*}
where $\paren{\sigstargen_j}_{j = 1}^k$ and $\paren{\sigstarint_j}_{j = 1}^k$ are now the $d$-dimensional and scalar parameters, respectively. We use the notation $\paren{\sigstargen, \sigstarint }$ to denote the $k \times d$ matrix and $k$-vector that collects these parameters, respectively. We denote the augmented parameter as $\left\lbrace \sigstaraug_j = (\sigstargen_j \,\, \, \sigstarint_j) \right \rbrace_{j=1}^k $ and similar to equation~\eqref{eq:pi_min_def}, we define the quantity $\pi_{\min}(\sigstaraug)$.
Assume that the covariates $\paren{ \covarstar_i }_{i = 1}^n$ 
are drawn i.i.d from the Gaussian distribution $\NORMAL(0,I_d)$ and $\epsilon_i$  represents independent noise drawn i.i.d. from a zero-mean, $\sigma$-sub-Gaussian distribution. 

Our goal is to perform least squares estimation under this model, but with errors in variables.
%
In particular, we assume that we do not observe the true covariates, but instead a set of ``erroneous" covariates $\paren{ \covarhat_i }_{i = 1}^n$, with the error in observation $i$ given by
\[
\msdiff_i := \max \left (\sigstargen \covarstar_i +\sigstarint \right ) - \max \left (\sigstargen \covarhat_i +\sigstarint \right ) \text{ for each } i \in [n].
\]
Let $\msdiff$ denote the $n$-dimensional vector that stacks these scalar values.
We now solve the mis-specified least squares problem
\begin{align} \label{eqn:ls_estimate}
\left\lbrace \hat{\siggen}_{\mathsf{ms}}, \hat{\sigint}_{\mathsf{ms}} \right \rbrace = \argmin_{ \substack{\siggen \in \real^{k \times \numdim} \\ \sigint \in \real^k } }  \sum_{i = 1}^n \left(\resp_i - \max (\siggen \covarhat_i +\sigint) \right)^2,
\end{align}
where the maximum is computed element-wise.  

The following proposition bounds the error of these estimates; recall the notation $\dist$ defined in equation~\eqref{eq:distfunc}.

 
\begin{proposition}[Mis-specified least squares]
\label{prop:mis_ls}
Suppose that $\numobs \geq C d\frac{k}{ \pi_{\min}}$. Then provided the mis-specified covariates $\covarhat_i \stackrel{i.i.d.}{\sim} \NORMAL(0, I_d)$, we have
\begin{align*}
\dist \left( (\widehat{\siggen}_{\mathsf{ms}}, \widehat{\sigint}_{\mathsf{ms}} ), ( \sigstargen, \sigstarint) \right) \leq C \log(k/\pi_{\min}) \left(\frac{k}{\pi_{\min}}\right)^5 \bigg (\frac{k^2 (\log k)(\log n)}{n} + \frac{1}{n}\norms{\msdiff}^2  \bigg )
\end{align*} 
with probability at least $1-c \cdot n^{-10}- c_1 k \exp(-c_2 n)$.
\end{proposition}

Taking this proposition as given, let us first prove Theorem~\ref{thm:init}.

\paragraph{Proof of Theorem~\ref{thm:init}:}

Recall the decomposition $\Thetastar = \Astar (\Ustar)^\top$, where $\Astar \in \R^{k \times k}$ is the invertible matrix of coefficients and $\Ustar \in \R^{k \times \numdim}$ is a matrix of orthonormal columns. 
Note that the PCA step returns a subspace $\Uhat$ such that $\Uhat O$ is close to $U^*$ with high probability for some rotation matrix $O$ with $O^\top O = O O^\top = I_k$. Also recall that we construct the $k$-dimensional covariates $\widetilde{x}_i = \Uhat^\top x_i$. In addition, it is helpful to write the observation model of \eqref{eq:model} as 
\begin{align} \label{eqn:obs_rewrite}
y_i = \max \left (A^* O^\top O (\Ustar)^\top x_i + \bstar \right ) + \epsilon_i,
\end{align}
where the maximum is computed element-wise.


In step 2 of the algorithm, we compute the least squares estimate with the covariates $\paren{ \widetilde{x}_i }_{i = n/2 + 1}^n$. Note that conditioned on $\Uhat$, these covariates are Gaussian, since we split samples across the two steps of the algorithm. Applying Proposition~\ref{prop:mis_ls} with $\sigstargen = A^* O^\top$ and $\sigstarint = \bstar$ thus yields the bound
\begin{align} \label{eq:init_with_rho}
\dist \left( \{ \beta^{(0)}_j \}_{j=1}^k, \{\beta^*_j \}_{j=1}^k \right) \leq C \log(k/\pi_{\min}) \left(\frac{k}{\pi_{\min}}\right)^5 \bigg (\frac{k^2 (\log k)(\log n)}{n} + \frac{1}{n}\norms{\msdiff}^2  \bigg )
\end{align}
with probability exceeding $1-c \numobs^{-10} - c_1\exp(-\numobs)$, where $\msdiff$ is a vector with entry $i$ given by 
%
where 
\begin{align*}
\msdiff_i = \max \left (A^* O^\top O (U^*)^\top x_i + \bstar \right ) - \max \left(A^* O^\top \tilde{x}_i + \bstar \right).
\end{align*}
In order to complete the proof, it suffices to bound the term $\norms{\msdiff}^2$.

\paragraph{Bounding $\norms{\msdiff}^2$:}
Using the Lipschitzness of the $\max$ function and substituting $\tilde{x}_i = \Uhat^\top x_i$, we obtain
\begin{align*}
|\msdiff_i|^2 &\leq \norms{A^* O^\top O(U^*)^\top x_i - A^* O^\top \tilde{x}_i}^2 \\
&= \norms{A^* O^\top \left( O (U^*)^\top - \hat{U}^\top\right) x_i}^2 \\
&\leq \opnorm{A^*O^\top}^2 \norms{\left( O (U^*)^\top - \hat{U}^\top\right) x_i}^2 \\
&= \opnorm{A^*}^2 \norms{\left( O (U^*)^\top - \hat{U}^\top\right) x_i}^2,
\end{align*}
where the final step follows since the operator norm is unitarily invariant.
Now note that conditioned on $\Uhat$, the quantity $\left( O (U^*)^\top - \hat{U}^\top\right) x_i$ is a Gaussian random vector. Using the shorthand $V:= O (U^*)^\top - \Uhat^\top $, we have 
\begin{align*}
\mathbb{E} \norms{V x_i }^2 =  \tr \left( V^\top V \mathbb{E} (x_i x_i^\top) \right) = \fronorm{V}^2.
\end{align*}
Using the Hanson-Wright inequality (see \cite[Theorem 2.1]{rudelson2013hanson}), we obtain
\begin{align*}
\Prob \left(\left | \norms{V x_i}- \fronorm{V} \right | \geq t \right) \leq 2 \exp \left( - \frac{ct^2}{\fronorm{V}^2} \right)
\end{align*}
Substituting $t = c_1 \fronorm{V} \log \numobs$, we obtain
\begin{align*}
\norms{\left( O (U^*)^\top - \hat{U}^\top\right) x_i}^2 \leq C \log \numobs \norms{\left( O (U^*)^\top - \hat{U}^\top\right)}^2
\end{align*}
with probability at least $1-c n^{-10}$. Now putting together all the pieces yields the bound 
\begin{align*}
\frac{1}{n}\norms{\msdiff}^2 \leq C \opnorm{A^*}^2 \fronorm{\Uhat \Uhat^\top - \Ustar (\Ustar)^\top}^2 \log(nk),
\end{align*}
with probability at least $1-c n^{-10}$. We substitute the above bound in equation~\eqref{eq:init_with_rho} and use the fact that $\opnorm{A^*}=\opnorm{\Thetastar}$. Hence, the parameter estimates satisfy
\begin{align*}
\dist \left( \{ \beta^{(0)}_j \}_{j=1}^k, \{\beta^*_j \}_{j=1}^k \right) \leq  C \opnorm{\Thetastar}^2 \,\, \fronorm{\Uhat \Uhat^\top - \Ustar (\Ustar)^\top}^2  \frac{k^3 \log (nk)}{\pi_{\min}^3}
\end{align*}
with probability at least $1-c_2\numobs^{-10} - \exp(- c_1 \numobs)$, which proves the theorem.
\qed

\subsection{Proof of Proposition~\ref{prop:mis_ls}}
\label{subsec:mis_ls}

As mentioned before, there are two technical challenges involved in obtaining this result; the first is to handle the mis-specification in the model, and the second is to provide a bound on parameter estimation. We handle each challenge separately.

The following lemma upper bounds the prediction error of a mis-specified least squares problem. Recall the setup of equation~\eqref{eqn:ls_estimate} along with the notation $\left( \{\covarhat_i \}_{i=1}^\numobs, \widehat{\siggen}_{\mathsf{ms}}, \widehat{\sigint}_{\mathsf{ms}},\sigstargen, \sigstarint, \sigstaraug, \varrho, \pi_{\min}(\sigstaraug) \right)$.
 \begin{lemma}
 \label{lem:mis_ls_upper}
For a fixed set of covariates $\{ \covarhat_i \}_{i=1}^\numobs$, the inequality   
 \begin{align*}
 \frac{1}{n}\sum_{i=1}^{n} \bigg ( \max(\widehat{\siggen}_{\mathsf{ms}} \covarhat_i + \widehat{\sigint}_{\mathsf{ms}}) - \max(\sigstargen \covarhat_i + \sigstarint) \bigg )^2 \leq  C \bigg (\frac{k d (\log k)(\log n)}{n} + \frac{\norms{\msdiff}^2}{\numobs} \bigg )
\end{align*}
holds with probability at least $1-cn^{-10}$.
\end{lemma}
The proof of this lemma follows standard chaining tools, and is postponed to Appendix~\ref{app:mis-sp}.

We now exploit the relationship between the prediction error and estimation error (Lemma~\ref{lem:pred_est_gen}) to obtain  
\begin{align*}
& \frac{1}{n} \sum_{i =  1}^n \bigg ( \max (\widehat{\siggen}_{\mathsf{ms}} \covarhat_i + \widehat{\sigint}_{\mathsf{ms}}) - \max (\sigstargen \covarhat_i +\sigstarint) \bigg )^2 \\
 &\geq C_1 \log^{-1}(k/\pi_{\min}(\sigstaraug)) \left (\frac{\pi_{\min}(\sigstaraug)}{k} \right)^5 \dist \left( \{\nu_j^* \}_{j=1}^k, \{\nu_j \}_{j=1}^k \right),
\end{align*}
with probability exceeding $1- c k \exp(-c_1 n)$ provided $\numobs \geq C  d\frac{k}{\pi_{\min}}$.

 Hence Proposition~\ref{prop:mis_ls} follows immediately via combining Lemma~\ref{lem:mis_ls_upper} with the above-mentioned lower bound.
 \qed


\subsection{Proof of Lemma~\ref{lem:mis_ls_upper}} \label{app:mis-sp}

Recall the setup of Section~\ref{subsec:mis_ls} along with the notation $\left( \{\covarhat_i \}_{i=1}^\numobs, \widehat{\siggen}_{\mathsf{ms}}, \widehat{\sigint}_{\mathsf{ms}},\sigstargen, \sigstarint, \varrho \right)$. Let us define the class of $d$-variate functions
\begin{align*}
\mathcal{F}= \left\lbrace f_{(\siggen,\sigint)} \bigg | f_{(\siggen,\sigint)}(x) = \max (\siggen x + \sigint) \,\,\, \text{for some} \,\, \siggen \in \real^d \,\,\, \text{and}\,\, \sigint \in \real \right \rbrace.
\end{align*}
We study the least squares estimator, given by
\begin{align*}
\widehat{f}_n \in \argmin_{f_{(\siggen, \sigint)} \in \mathcal{F}} \; \frac{1}{n} \sum_{i=1}^{n} \bigg ( \resp_i - f_{(\siggen, \sigint)}(\covarhat_i) \bigg )^2.
\end{align*}
In particular, since the pair $( \widehat{\siggen}_{\mathsf{ms}}, \widehat{\sigint}_{\mathsf{ms}} )$ solves the least squares problem~\eqref{eqn:ls_estimate}, we have $\widehat{f}_n(x) = \max \left( \hat{\siggen}_{\mathsf{ms}} x + \hat{\sigint}_{\mathsf{ms}} \right)$ for each $x \in \real^d$. Also we define the shorthand $f^*(x) = \max (\sigstargen x + \sigstarint)$. Throughout the proof, we use the shorthand notation
\begin{align*}
\| f_{(\siggen, \sigint)} - f^* \|_n^2 = \frac{1}{n} \sum_{i=1}^{n} \bigg ( f^* (\covarhat_i) - f_{(\siggen, \sigint)}(\covarhat_i) \bigg )^2.
\end{align*}
We now use the standard convergence analysis for non-parametric least squares (e.g. see \cite[Chapter 13]{wainwright2019high}) in order to establish a bound on the prediction error of $\widehat{f}_n$. 
Using equation~\eqref{eqn:ls_estimate} and recalling the definition of the vector $\msdiff \in \real^n$, we  write
\begin{align*}
\frac{1}{n} \sum_{i=1}^{n} \bigg ( \resp_i - f_{(\siggen, \sigint)}(\covarhat_i) \bigg )^2 & =  \frac{1}{n} \sum_{i=1}^{n} \bigg ( f^* (\covarhat_i) + \msdiff_i +\epsilon_i - f_{(\siggen, \sigint)}(\covarhat_i) \bigg )^2 \\
& =  \| f_{(\siggen, \sigint)} - f^* \|_n^2 + \frac{1}{n}\sum_{i=1}^{n}(\epsilon_i + \msdiff_i)^2 \\ 
& \qquad \qquad \qquad -\frac{2}{n} \sum_{i=1}^{n} (\epsilon_i + \msdiff_i) \left( f^*(\covarhat_i) - f_{(\siggen, \sigint)}(\covarhat_i) \right).
\end{align*}
Hence, 
\begin{align*}
\widehat{f}_n \in \argmax_{f_{(\siggen, \sigint)} \in \Fspace } \bigg ( \frac{2}{n} \sum_{i=1}^{n} (\epsilon_i + \msdiff_i) ( f^*(\covarhat_i) - f_{(\siggen, \sigint)}(\covarhat_i) ) - \| f_{(\siggen, \sigint)} - f^* \|_n^2 \bigg )
\end{align*}
and note that 
\begin{align*}
f^* \in \argmin_{f_{(\siggen, \sigint)} \in \Fspace } \| f_{(\siggen, \sigint)} - f^* \|_n^2.
\end{align*}
We choose the distance metric
\begin{align*}
\Delta(f_{(\siggen,\sigint)} ,f^*)= \| f_{(\siggen, \sigint)} - f^* \|_n,
\end{align*}
and for shorthand notation, we use $\Delta:=\Delta(f_{(\siggen,\sigint)} ,f^*)$.  To find the rate of convergence, it is sufficient (see, e.g.,~\cite[Chapter 13]{wainwright2019high}) to control the following quantity:
\begin{align*}
  \mathbb{E} & \left [ \sup_{f_{(\siggen, \sigint)} \in \mathcal{F}: \Delta \leq \delta} \frac{1}{n}\sum_{i=1}^{n} \epsilon_i \left( f_{(\siggen, \sigint)}(\covarhat_i)- f^*(\covarhat_i) \right ) \right ]  + \frac{1}{n}\sum_{i=1}^{n} \msdiff_i (f_{(\siggen, \sigint)}(\covarhat_i)- f^*(\covarhat_i)),
\end{align*}
which, upon further simplification, yields
\begin{align}
 \mathbb{E} & \left[ \sup_{f_{(\siggen, \sigint)} \in \mathcal{F}: \Delta \leq \delta} \frac{1}{n}\sum_{i=1}^{n} \epsilon_i \left ( f_{(\siggen, \sigint)}(\covarhat_i)- f^*(\covarhat_i) \right) \right ] + \frac{\norms{\msdiff} \delta}{\sqrt{n}}. \label{eqn:rate_equation}
\end{align}
To control the first term of the equation~\eqref{eqn:rate_equation}, we  use Dudley's entropy integral. Let us define a sub-Gaussian process $X_f$ indexed by the function $f = f_{(\siggen, \sigint)}$ as 
\begin{align*}
X_f := \frac{1}{\sqrt{n}}\sum_{i=1}^{n} \epsilon_i (f (\covarhat_i)- f^*(\covarhat_i)).
\end{align*}
Since $\epsilon_i$ is $\sigma$-sub-Gaussian, we have, for a pair for functions $(g, h)$, the tail bound
\begin{align*}
\Prob \left ( |X_g -X_{h} | \geq u \right ) \leq 2 \exp \left (-\frac{u^2}{2\Delta^2(g,h)} \right ),
\end{align*}
for all $u \geq 0$. Dudley's entropy integral (e.g. see \cite{wainwright2019high}) then yields
\begin{align*}
& \mathbb{E} \bigg [\sup_{f_{(\siggen, \sigint)} \in \mathcal{F}: \Delta \leq \delta} \frac{1}{n}\sum_{i=1}^{n} \epsilon_i (f_{(\siggen, \sigint)}(\covarhat_i)- f^*(\covarhat_i)) \bigg ] \\
& \qquad \qquad \leq \frac{1}{\sqrt{n}}\mathbb{E} \left [ \sup_{f_{(\siggen, \sigint)} \in \mathcal{F}: \Delta \leq \delta}|X_{f_{(\siggen, \sigint)}} - X_{f^*}| \right ] \\
& \qquad \qquad \leq \frac{C}{\sqrt{n}}\int_{0}^{\delta} \sqrt{\log N(\varepsilon, \{f_{(\siggen, \sigint)} \in \mathcal{F}:\Delta \leq \delta \},\Delta(\cdot,\cdot))} d \varepsilon
\end{align*}
where $N(.)$ is the local covering number of $\mathcal{F}$ in the distance metric $\Delta(\cdot,\cdot)$. Using Lemma B.1 of Guntuboyina \cite{guntuboyina2012optimal}, we obtain
\begin{align*}
N(\varepsilon, \{f\in \mathcal{F}:\Delta \leq \delta \},\Delta(.,.)) \leq C \bigg (1 + \frac{\sqrt{n} \delta }{\varepsilon} \bigg )^{k d \log k}.
\end{align*}
Substituting, we have
\begin{align*}
\mathbb{E} \bigg [ \sup_{f_{(\siggen, \sigint)} \in \mathcal{F}: \Delta \leq \delta} \frac{1}{n}\sum_{i=1}^{n} \epsilon_i (f_{(\siggen, \sigint)}(\covarhat_i)- f^*(\covarhat_i)) \bigg ] & \leq C \sqrt{\frac{k d \log k}{n}} \int_{0}^{\delta} \sqrt{\log \bigg (1+\frac{\sqrt{n} \delta }{\varepsilon} \bigg )} d \varepsilon \\
& \leq C_1 \delta \sqrt{\frac{k d \log k}{n}} \bigg ( \sqrt{\log n} + \sqrt{\log \delta} \bigg) \\
& \leq C_2 \delta \sqrt{\frac{k d (\log k)(\log n)}{n}}.
\end{align*}
provided $n \geq \delta$. Hence, the critical inequality takes the form
\begin{align*}
\delta \sqrt{\frac{k d (\log k)(\log n)}{n}} + \frac{\norms{\msdiff} \delta}{\sqrt{n}} \lesssim \delta^2,
\end{align*}
and applying \cite[Theorem 13.2]{wainwright2019high} with the choice 
\begin{align*}
\delta = c \sqrt{\frac{k d(\log k)(\log n)}{n}} + c \frac{\norms{\msdiff}}{\sqrt{n}},
\end{align*}
immediately implies that
\begin{align*}
\| f_{(\siggen, \sigint)} - f^* \|_n^2 \leq C \bigg (\frac{k d (\log k)(\log n)}{n} + \frac{\norms{\msdiff}^2}{n} \bigg ),
\end{align*}
with probability exceeding $1 - c n^{-10}$. Substituting the explicit forms of $\widehat{f}_n$ and $f^*$, we obtain
\begin{align*}
\frac{1}{n}\sum_{i=1}^{n} \bigg ( \max(\hat{\siggen}_{\mathsf{ms}} \covarhat_i + \hat{\sigint}_{\mathsf{ms}}) - \max(\sigstargen \covarhat_i + \sigstarint) \bigg )^2 \leq C \bigg (\frac{k d (\log k)(\log n)}{n} + \frac{\norms{\msdiff}^2}{n} \bigg ),
\end{align*}
which proves Lemma~\ref{lem:mis_ls_upper}.
\qed

\end{document}